\documentclass{article}

\usepackage{arxiv}

\usepackage[utf8]{inputenc} 
\usepackage[T1]{fontenc}    
\usepackage{hyperref}       
\usepackage{url}            
\usepackage{booktabs}       
\usepackage{amsfonts}       
\usepackage{nicefrac}       
\usepackage{microtype}      
\usepackage{lipsum}		
\usepackage{graphicx}
\usepackage{natbib}
\usepackage{doi}

\usepackage{xcolor}

\usepackage{amsfonts,amsmath,amssymb,wasysym,amsthm}
\usepackage{graphicx}
\usepackage{subcaption} 
\usepackage{enumitem}
\usepackage{booktabs}
\usepackage{tabularx}
\usepackage{algorithm}
\usepackage{algorithmic}

\usepackage{makecell} 
\usepackage{enumitem}
\usepackage{float}
\usepackage{appendix}

\newtheorem{theorem}{Theorem}
\newtheorem{lemma}{Lemma}
\newtheorem{defin}{Definition}

\title{Quantifying Data Similarity Using Cross Learning}

\author{Shudong Sun \\
	Statistics and Data Science GIDP\\
	University of Arizona\\
	Tucson, AZ 85721 \\
	\texttt{shudongsun@arizona.edu} \\
	\And
	Hao Helen Zhang \\
	Statistics and Data Science GIDP\\
	University of Arizona\\
	Tucson, AZ 85721 \\
	\texttt{haozhang@arizona.edu} \\
    \And
	Joseph C Watkins \\
	Statistics and Data Science GIDP\\
	University of Arizona\\
	Tucson, AZ 85721 \\
	\texttt{jwatkins@arizona.edu} \\
}

\renewcommand{\shorttitle}{\textit{arXiv} Template}

\newcommand{\jw}[1]{\textcolor{black}{#1}}

\begin{document}
\maketitle

\begin{abstract}
Measuring dataset similarity is fundamental in machine learning, particularly for transfer learning and domain adaptation. In the context of supervised learning, most existing approaches quantify similarity of two data sets based on their input feature distributions, neglecting label information and feature–response alignment. To address this, we propose the Cross-Learning Score (CLS), which measures dataset similarity through bidirectional generalization performance of decision rules. We establish its theoretical foundation by linking CLS to cosine similarity between decision boundaries under canonical linear models, providing a geometric interpretation. A robust ensemble-based estimator is developed that is easy to implement and bypasses high-dimensional density estimation entirely. For transfer learning applications, we introduce a "transferable zones" framework that categorizes source datasets into positive, ambiguous, and negative transfer regions. To accommodate deep learning, we extend CLS to encoder–head architectures, aligning with modern representation-based pipelines. Extensive experiments on synthetic and real-world datasets validate the effectiveness of CLS for similarity measurement and transfer assessment.
\end{abstract}


\section{Introduction}
Measuring dataset similarity is a fundamental problem in machine learning and AI with broad applications. The core question is whether knowledge from one dataset generalizes usefully to another. This arises in several key contexts. Distribution shift  detection \citep{polo2023unified} requires monitoring whether incoming data has drifted from the training distribution to maintain model reliability in production. Few-shot \citep{snell2017prototypical} and meta-learning \citep{finn2017model} depend on task similarity to identify the most relevant prior experience when learning from limited examples, such as fine-tuning a speech recognition model pre-trained on standard speech for a specific accent \citep{do2024improving}. Multi-task learning \citep{standley2020tasks, fifty2021efficiently} leverages similarity assessment to determine which tasks benefit from joint training and shared representations, for example, combining datasets from multiple hospitals to build generalizable predictive models \citep{reps2022learning}. Federated learning \citep{tan2023pfedsim} requires evaluating similarity across distributed client datasets to guide aggregation strategies. Benchmarking and reproducibility \citep{koh2021wilds} also benefit from similarity metrics to assess how representative benchmark datasets are relative to real-world deployment data.

How to evaluate the similarity between different datasets is also essential for transfer learning, as understanding source–target alignment helps to predict whether transfer will improve or degrade performance. For example, \citet{yosinski2014transferable} demonstrated positive transfer within ImageNet, while \citet{wang2019characterizing} observed negative transfer across domains in Office-31. Similarly, dataset selection and curation \citep{cui2018large} rely on similarity measures to identify the most relevant candidates for training or augmentation \citep{yao2022improving}, avoiding negative transfer and unnecessary computation. Since the effectiveness of knowledge transfer depends critically on source–target similarity, reliable measures are fundamental for data adaptation, fine-tuning, and integration in practice.

Despite its intuitive appeal, rigorously defining and characterizing dataset similarity remains challenging. In this paper, we introduce a new metric for measuring dataset similarity, establish its theoretical properties for certain learning tasks, and evaluate its empirical performance across a variety of settings. We focus on supervised learning, where a dataset
$\mathcal{D}=\{(\mathbf{x}_i, y_i), i=1, \cdots, n\}$
is drawn from a joint distribution $\mathcal{P}(\mathbf{X}, Y)$ over $\mathcal{X}\times\mathcal{Y}$, and
$\mathcal{X}\subseteq\mathbb{R}^p$ is the input space, 
$\mathcal{Y}$ is the output space, $n$ is the sample size. In transfer learning, we deal with one target dataset $\mathcal{D}^{(t)}=\{(\mathbf{x}_i^{(t)}, y_i^{(t)}), i=1, \cdots, n_t\}\sim \mathcal{P}^{(t)}$
and one source dataset 
$\mathcal{D}^{(s)}=\{(\mathbf{x}_i^{(s)}, y_i^{(s)}), i=1,\cdots, n_s\}\sim \mathcal{P}^{(s)}$ defined over $\mathcal{X}\times\mathcal{Y}$ but with potentially different distributions. The target dataset is typically smaller and aims to improve performance by learning from source datasets, which provide auxiliary knowledge. In multi-source transfer learning, given $m$ source datasets \(\mathcal{D}^{(s)}_1,\ldots,\mathcal{D}^{(s)}_m\), the goal is to rank them by their similarity to the target dataset and integrate them via a weighting scheme that promotes positive transfer.

In the literature, various types of approaches have been proposed to measure dataset similarity, including kernel-based, divergence-based, and classification-based metrics; \citet{stolte2024methods} provides a comprehensive survey. The kernel-based \textit{Maximum Mean Discrepancy} (MMD) \citep{gretton2012kernel} quantifies the distance between \(\mathcal{P}^{(t)}\) and \(\mathcal{P}^{(s)}\) in a Reproducing Kernel Hilbert Space.
The f-divergence methods are based on the principle that two identical distributions yield identical likelihoods at every point, and quantify the deviation of the likelihood ratio from one \citep{zhao2022comparing}. Given any convex continuous function \(f:\mathbb{R}_+\!\to\!\mathbb{R}\) satisfying \(f(1)=0\), f-divergence is defined as
\[
D_f(\mathcal{P}^{(t)}\|\mathcal{P}^{(s)})=\mathbb{E}_{\mathbf{X}\sim\mathcal{P}^{(s)}}\!\left[f\!\left(\frac{p^{(t)}(\mathbf{X})}{p^{(s)}(\mathbf{X})}\right)\right],
\]
where \(p^{(t)}(\mathbf{x})\) and \(p^{(s)}(\mathbf{x})\) denote the probability density functions of \(\mathcal{P}^{(t)}\) and \(\mathcal{P}^{(s)}\), respectively. Common examples include the Kullback–Leibler divergence with \(f(t)=t\log t\), and the Jensen–Shannon divergence with \(f(t)=(t+1)\log(2/(t+1))+t\log t\).

Most existing methods consider only feature similarity between $\mathbf{X}^{(t)}$ and $\mathbf{X}^{(s)}$, while ignoring label information in $Y^{(t)}, Y^{(s)}$ and the feature-response relationships. This can seriously hinder transferability assessment, since datasets with similar features may represent fundamentally different prediction tasks. While label-aware methods like Optimal Transport Dataset Distance (OTDD) \citep{alvarez2020geometric} address this limitation, they are computationally expensive and sensitive to dimensionality.

Motivated by these limitations, we propose a new metric, \textbf{Cross-Learning Score (CLS)}, to directly quantify similarity of feature–response relationships between target and source datasets. 
In supervised learning, predictive performance depends on the relationship between $\mathbf{X}$ and $Y$. 
When these relationships are similar across $\mathcal{D}^{(t)}$ and $\mathcal{D}^{(s)}$, the two tasks share similar underlying functions or decision boundaries, making transfer beneficial; otherwise, transfer may be ineffective or harmful.
Based on this idea, CLS measures the similarity between \(Y^{(t)}|\mathbf{X}^{(t)}\) and \(Y^{(s)}|\mathbf{X}^{(s)}\) through bidirectional cross-domain generalization performance. 
We establish theoretical support by showing that CLS is closely related to the cosine similarity between decision boundaries in several canonical settings. 
Compared with distance-based approaches, CLS is computationally efficient and scalable to high-dimensional data because it avoids explicit density estimation. We further improve its robustness through adaptive ensemble estimation.

We further apply CLS to transfer learning, which plays an essential role in deep learning \citep{yang2024deep}, using a principled framework that categorizes source datasets into three transferability zones—positive, ambiguous, and negative—corresponding to beneficial, uncertain, and detrimental transfer effects. Finally, we extend CLS to encoder–head architectures, enabling representation-level transferability assessment aligned with modern deep learning practice. Finally, the effectiveness of CLS is validated through simulated and real-world examples. 

\section{Proposed Research}

\subsection{Notation}

In supervised learning, for any learner $f:\mathcal{X}\to\mathcal{Y}$, the loss function $\ell(f(\mathbf{X}), Y)$ measures the discrepancy between the predicted response given by $f(\mathbf{X})$ and the true response $Y$. The choice of $\ell(\cdot,\cdot)$ depends on the learning task type. For classification, the 0-1 loss $\ell(f(\mathbf{X}),Y)=\mathbb{I}[f(\mathbf{X})\neq Y]$ measures label misclassification; for regression, the squared error loss $\ell(f(\mathbf{X}),Y)=[f(\mathbf{X})-Y]^2$ is commonly used. 
The Bayes rule $f^*: \mathcal{X} \to \mathcal{Y}$ is the optimal learner that minimizes the expected loss, or risk,  $R(f)=\mathbb{E}_{\mathbf{X},Y}l(f(\mathbf{X}),Y)$. For regression with squared error loss, $f^*(\mathbf{X})=\mathbb{E}[Y|\mathbf{X}]$; for classification with 0-1 loss, $f^*(\mathbf{X}) = \arg\max_{y} P(Y=y|\mathbf{X})$. In transfer learning, we assume the target dataset $\mathcal{D}^{(t)}\sim \mathcal{P}^{(t)}$
and source dataset 
$\mathcal{D}^{(s)}\sim \mathcal{P}^{(s)}$, both defined over $\mathcal{X}\times\mathcal{Y}$. Let $f^{*(t)}$ and $f^{*(s)}$ denote the Bayes rules for the target and source learning problems, respectively.

\subsection{Cross-Learning Score(CLS)}

\begin{figure}[t]
    \centering
    \begin{subfigure}{0.49\textwidth}
        \centering
        \includegraphics[width=\textwidth]{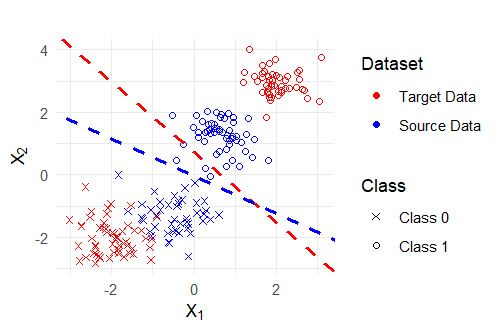}
        \caption{Linear example: similar Bayes boundaries between the target and the source data.}
        \label{figCLSLinear:subfig1}
    \end{subfigure}
    \begin{subfigure}{0.49\textwidth}
        \centering
        \includegraphics[width=\textwidth]{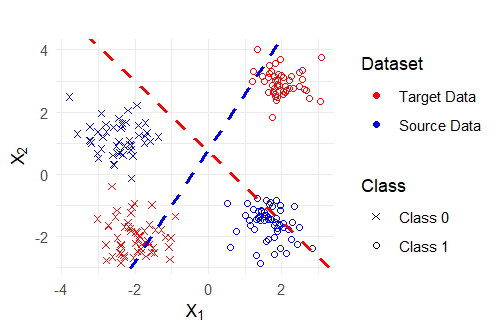}
        \caption{Linear example: dissimilar Bayes boundaries between the target and the source data.}
        \label{figCLSLinear:subfig2}
    \end{subfigure}
    \caption{Different linear feature-response relationships between two datasets.}
    \label{figCLSLinear}
\end{figure}

\begin{figure}[t]
    \centering
    \begin{subfigure}{0.49\textwidth}
        \centering
        \includegraphics[width=\textwidth]{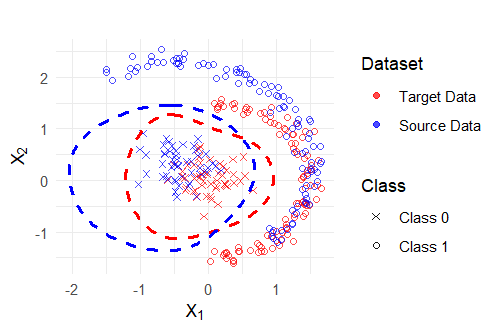}
        \caption{Nonlinear example: similar Bayes boundaries between the target and the source data.}
        \label{figCLSNonLinear:subfig1}
    \end{subfigure}
    \begin{subfigure}{0.49\textwidth}
        \centering
        \includegraphics[width=\textwidth]{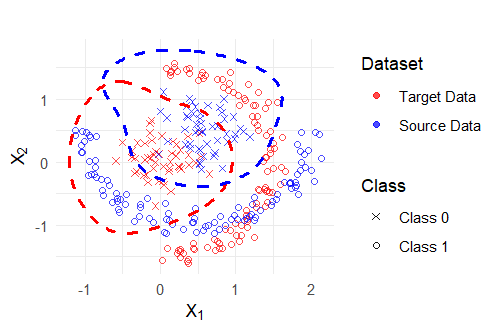}
        \caption{Nonlinear example: dissimilar Bayes boundaries between the target and the source data.}
        \label{figCLSNonLinear:subfig2}
    \end{subfigure}
    \caption{Different nonlinear feature-response relationships between two datasets.}
    \label{figCLSNonLinear}
\end{figure}

To assess similarity between target and source tasks, we consider their feature-response relationships, $Y^{(t)}|\mathbf{X}^{(t)}$ and $Y^{(s)}|\mathbf{X}^{(s)}$. Intuitively, when these relationships are similar, $f^{*(t)}$ should predict well on source data, and $f^{*(s)}$ should also predict well on target data. Figure~\ref{figCLSLinear}  illustrates this concept with a binary classification example. Red markers indicate samples in $\mathcal{D}^{(t)}$ (“$\times$” for class 0, “$\circ$” for class 1) with the red dashed line showing the target Bayes boundary. Blue markers indicate samples in $\mathcal{D}^{(s)}$ with the blue dashed line showing the source Bayes boundary. It is observed that, in Figure~\ref{figCLSLinear:subfig1}, when the two Bayes boundaries are similar, both Bayes classifiers generalize well on the other dataset. 
In Figure~\ref{figCLSLinear:subfig2}, when the boundaries are differ substantially, each classifier trained on one task performs poorly on the other task. Figure~\ref{figCLSNonLinear} illustrates this for nonlinear classification. When the two datasets share similar decision boundaries (Figure~\ref{figCLSNonLinear:subfig1}), cross-task errors are low, indicating strong similarity. When boundaries diverge (Figure~\ref{figCLSNonLinear:subfig2}), high cross-task errors indicate weak similarity.

Motivated by the above, we introduce a new metric, the \textbf{Cross-Learning Score} (CLS), to quantify the similarity between two learning tasks through their predictive structures. Intuitively, CLS measures how differently the two domain-optimal predictors behave when evaluated on the same data distribution. If the optimal predictor from the source domain performs similarly to the optimal predictor from the target domain on target data, and vice versa, then the two tasks share similar decision rules. Large performance gaps, on the other hand, indicate structural differences in the feature–response relationships.

Let $\ell(\cdot,\cdot)$ denote a task-specific loss function. 
We define the directional discrepancy from source to target as
\[
\Delta_{s \to t}
=
\mathbb{E}_{(\mathbf{X},Y)\sim\mathcal{P}^{(t)}}
\!\left[\ell(f^{*(s)}(\mathbf{X}),Y)\right]
-
\mathbb{E}_{(\mathbf{X},Y)\sim\mathcal{P}^{(t)}}
\!\left[\ell(f^{*(t)}(\mathbf{X}),Y)\right],
\]
which measures how much worse the source-optimal predictor performs compared to the target-optimal predictor under the target distribution. Similarly, define
\[
\Delta_{t \to s}
=
\mathbb{E}_{(\mathbf{X},Y)\sim\mathcal{P}^{(s)}}
\!\left[\ell(f^{*(t)}(\mathbf{X}),Y)\right]
-
\mathbb{E}_{(\mathbf{X},Y)\sim\mathcal{P}^{(s)}}
\!\left[\ell(f^{*(s)}(\mathbf{X}),Y)\right],
\]
which quantifies the corresponding discrepancy under the source distribution.

The Cross-Learning Score is then defined as the symmetric average of these two directional discrepancies:
\[
\mathrm{CLS}
=
\frac{1}{2}
\left(
\Delta_{s \to t}
+
\Delta_{t \to s}
\right).
\]

Under the 0-1 loss for classification, each expectation corresponds to a misclassification rate in $[0,1]$. 
Therefore, $\Delta_{s\to t}, \Delta_{t\to s} \in [-1,1]$, and $\mathrm{CLS} \in [-1,1]$ in classification problems. For other loss functions, such as the squared error loss in regression, CLS is not necessarily bounded within $[-1,1]$, as the loss values themselves are unbounded. In general, a small CLS indicates that the two optimal predictors behave similarly across both domains, implying high data similarity and strong cross-domain compatibility. A large CLS indicates substantial predictive discrepancy between domains, reflecting lower data similarity.

In the above formulation, both tasks are treated equally, making CLS a symmetric measure. If one task is more important than the other, a weighted average 
$\mathrm{CLS}_w =
w ~\Delta_{s \to t}  + (1-w) ~  \Delta_{t \to s}
$ can be used, where $0\leq w\leq 1$ is a pre-specified weight parameter to indicate the difference.  

The CLS is a theoretical concept that can not computed without knowledge of the Bayes rule $f^*$. We refer to this as the \textit{oracle} CLS. In practice, we must estimate it from data. Section~\ref{comp_alg} presents its empirical approximation, $\widehat{\mathrm{CLS}}$, computed from the observed datasets $\mathcal{D}^{(t)}$ and $\mathcal{D}^{(s)}$ using several algorithms.

\subsubsection{Theoretical Validity of CLS}
We establish the theoretical foundation of CLS under two canonical linear settings: the probit regression model and linear discriminant analysis (LDA). In these settings, CLS is shown to be closely connected to cosine similarity between decision boundaries of the target and source tasks. This result provides a geometric interpretation of CLS in terms of boundary alignment. Although closed-form expressions are available only in linear cases, the theoretical analysis offers intuition for more general scenarios. Our numerical experiments suggest that the monotonic relationship between CLS and underlying task similarity extends to nonlinear decision boundaries as well.
\begin{defin}
\jw{For any symmetric positive semidefinite matrix $\Sigma$ and nonnegative scalars $\tau_x,\tau_y$, define the noise-adjusted generalized cosine similarity
\[
\cos_{\Sigma}^{(\tau_x,\tau_y)}(u,v)
=
\frac{u^\top \Sigma v}
{\sqrt{u^\top \Sigma u+\tau_x}\,
 \sqrt{v^\top \Sigma v+\tau_y}},
\]
whenever the denominator is nonzero.}
\end{defin}

\begin{theorem}
\label{theorem1}
Consider two binary classification tasks with the inputs 
$\mathbf{X}^{(t)} \sim \mathcal{N}_p(0,\Sigma^{(t)})$ and 
$\mathbf{X}^{(s)} \sim \mathcal{N}_p(0,\Sigma^{(s)})$, 
where $\Sigma^{(t)}$ and $\Sigma^{(s)}$ are positive semidefinite matrices, and 
the label follows the probit models
\[
Y^{(t)}=\mathbb{I}(Z^{(t)}\ge 0),\qquad
Y^{(s)}=\mathbb{I}(Z^{(s)}\ge 0),
\]
with latent scores
\[
Z^{(t)}=\beta^{(t)\top}\mathbf{X}^{(t)}+\xi^{(t)},\qquad
Z^{(s)}=\beta^{(s)\top}\mathbf{X}^{(s)}+\xi^{(s)},
\]
where $\xi^{(t)}\sim\mathcal{N}(0,\sigma_t^2)$ and 
$\xi^{(s)}\sim\mathcal{N}(0,\sigma_s^2)$ are independent of \jw{both $\mathbf{X}^{(s)}$ and $\mathbf{X}^{(t)}$}. 

Then the Cross-Learning Score admits the bidirectional angular decomposition
\[
\mathrm{CLS}
=
\frac{\theta_{t\to s}-\theta_{s\to s}}{2\pi}
+
\frac{\theta_{s\to t}-\theta_{t\to t}}{2\pi},
\]
where $\theta_{a\to b}=\arccos(\rho_{a\to b})$ and
\[
\rho_{t\to s}
=
\cos_{\Sigma^{(s)}}^{(0,\sigma_s^2)}
\big(\beta^{(t)},\beta^{(s)}\big),
\qquad
\rho_{s\to t}
=
\cos_{\Sigma^{(t)}}^{(\sigma_t^2,0)}
\big(\beta^{(t)},\beta^{(s)}\big),
\]
\[
\rho_{t\to t}
=
\cos_{\Sigma^{(t)}}^{(0,\sigma_t^2)}
\big(\beta^{(t)},\beta^{(t)}\big),
\qquad
\rho_{s\to s}
=
\cos_{\Sigma^{(s)}}^{(0,\sigma_s^2)}
\big(\beta^{(s)},\beta^{(s)}\big).
\]

In particular,
\[
\rho_{a\to a}
=
\sqrt{
\frac{\beta^{(a)\top}\Sigma^{(a)}\beta^{(a)}}
{\beta^{(a)\top}\Sigma^{(a)}\beta^{(a)}+\sigma_a^2}
},
\]
which characterizes the intrinsic signal-to-noise alignment within task $a$.
\end{theorem}

\jw{{\it Remark}.} When $\sigma_t^2=\sigma_s^2=0$, the quantity 
$\cos_{\Sigma}^{(0,0)}(u,v)$ reduces to the standard generalized cosine similarity \citep{qamar2009online} induced by $\Sigma$,
\[
\cos_{\Sigma}(u,v)
=
\frac{u^\top \Sigma v}
{\sqrt{u^\top \Sigma u}\sqrt{v^\top \Sigma v}},
\]
so that $\rho_{t\to s}$ and $\rho_{s\to t}$ become covariance-weighted cosine similarities between $\beta^{(t)}$ and $\beta^{(s)}$ under the respective task geometries.

The quantity $\theta_{a\to b}$ measures the angular misalignment between the predictor learned from task $a$ and the decision boundary of task $b$ under the distribution of task $b$. 
In particular, $\theta_{t\to s}-\theta_{s\to s}$ quantifies the additional misalignment incurred when applying the target predictor to the source task relative to the optimal source predictor, while $\theta_{s\to t}-\theta_{t\to t}$ captures the analogous effect when transferring from the source task to the target task. 
Therefore, CLS represents the total bidirectional excess angular discrepancy between the two tasks. 
A larger CLS indicates greater geometric misalignment between the decision boundaries, whereas CLS close to zero suggests that the two tasks share nearly identical latent decision structures.

\begin{lemma}\label{lemma1}
Assume that \(\Sigma^{(t)}=\Sigma^{(s)}=\mathbf{I}_p\). Let
\[
\theta_\beta
:=
\arccos\!\left(
\frac{\beta^{(t)\top}\beta^{(s)}}{\|\beta^{(t)}\|\,\|\beta^{(s)}\|}
\right)
\in [0,\pi]
\]
denote the angle between \(\beta^{(t)}\) and \(\beta^{(s)}\). Then, as
\(\sigma_s \to 0\) and \(\sigma_t \to 0\), the Cross-Learning Score satisfies
\[
\mathrm{CLS}
=
\begin{cases}
\dfrac{\theta_\beta}{\pi}
+ O\!\big(\sigma_s + \sigma_t\big),
& \text{if } 0<\theta_\beta<\pi, \\[6pt]
0,
& \text{if } \theta_\beta = 0, \\[6pt]
1 + O\!\big(\sigma_s + \sigma_t\big),
& \text{if } \theta_\beta = \pi.
\end{cases}
\]
\end{lemma}

\begin{theorem}\label{theorem2}
Consider a Linear Discriminant Analysis (LDA) setting with balanced priors for both the target and the source data. 
Furthermore, assume for the target data
\(\mathbf{X}^{(t)}|Y^{(t)}=1\sim\mathcal{N}_p(\mu^{(t)},I)\) and 
\(\mathbf{X}^{(t)}|Y^{(t)}=0\sim\mathcal{N}_p(-\mu^{(t)},I)\); 
for the source data, 
\(\mathbf{X}^{(s)}|Y^{(s)}=1\sim\mathcal{N}_p(\mu^{(s)},I)\) and 
\(\mathbf{X}^{(s)}|Y^{(s)}=0\sim\mathcal{N}_p(-\mu^{(s)},I)\), 
where $\mu^{(t)}, \mu^{(s)}\in\mathbb{R}^p$. Then
\[
\mathrm{CLS}=\frac{1}{2}\Big[
\Phi\!\big(-\|\mu^{(s)}\|\cos\theta_{\mu}\big)
- \Phi\!\big(-\|\mu^{(s)}\|\big)
\Big] + \frac{1}{2}\Big[
\Phi\!\big(-\|\mu^{(t)}\|\cos\theta_{\mu}\big)
- \Phi\!\big(-\|\mu^{(t)}\|\big)
\Big],
\]
where $\Phi(\cdot)$ is the standard Gaussian CDF and
$\cos\theta_\mu=\frac{{\mu^{(t)}}^\top\mu^{(s)}}{\|\mu^{(t)}\|\cdot\|\mu^{(s)}\|}$.
\end{theorem}

{\it Remark}. In both scenarios, CLS is theoretically linked to the cosine similarity between the target and source decision boundaries. The result aligns with the angle-based similarity metric introduced
by \citet{gu2024robust}.

\subsubsection{Computational Algorithms} \label{comp_alg}
Since the true Bayes rules and conditional distributions $Y^{(t)}|\mathbf{X}^{(t)}$ and $Y^{(s)}|\mathbf{X}^{(s)}$, are unknown in practice, CLS is not directly computable. We therefore develop two computational algorithms to obtain the empirical estimate $\widehat{\mathrm{CLS}}$ from 
available datasets.

Algorithm 1 employs a single model $M$ to estimate both Bayes rules: train $\hat{f}_M^{(t)}$  on $\mathcal{D}^{(t)}$ to approximate $f^{*(t)}$, and train $\hat{f}_M^{(s)}$ on $\mathcal{D}^{(s)}$ to approximate $f^{*(s)}$. The learned models are then cross-evaluated to compute average prediction losses. Details are given below.

\begin{algorithm}[t]
\caption{Estimate CLS using a single model $M$}
\label{alg:main}
\begin{algorithmic}[1]
\STATE \textbf{Input:} target dataset $\mathcal{D}^{(t)}$, source dataset $\mathcal{D}^{(s)}$, number of folds $K$
\STATE Train $M$ on $\mathcal{D}^{(t)}$ to obtain $\hat{f}^{(t)}_M$, then evaluate it on $\mathcal{D}^{(s)}$ to compute
\[
e^{(t \to s)}_M
=
\frac{1}{n_s}\sum_{i=1}^{n_s}
\ell\!\left(\hat{f}^{(t)}_M(\mathbf{x}^{(s)}_i), y^{(s)}_i\right).
\]
\STATE Train $M$ on $\mathcal{D}^{(s)}$ to obtain $\hat{f}^{(s)}_M$, then evaluate it on $\mathcal{D}^{(t)}$ to compute
\[
e^{(s \to t)}_M
=
\frac{1}{n_t}\sum_{i=1}^{n_t}
\ell\!\left(\hat{f}^{(s)}_M(\mathbf{x}^{(t)}_i), y^{(t)}_i\right).
\]
\STATE Partition $\mathcal{D}^{(t)}$ into $K$ disjoint folds $\mathcal{D}^{(t)}_1,\ldots,\mathcal{D}^{(t)}_K$.
For each fold $k$, train $M$ on $\mathcal{D}^{(t)}\setminus \mathcal{D}^{(t)}_k$ to obtain $\hat{f}^{(t,-k)}_M$.
Compute the $K$-fold cross-validation error on $\mathcal{D}^{(t)}$ as
\[
e^{(t)}_M
=
\frac{1}{K}
\sum_{k=1}^K
\frac{1}{|\mathcal{D}^{(t)}_k|}
\sum_{(\mathbf{x},y)\in\mathcal{D}^{(t)}_k}
\ell\!\left(\hat{f}^{(t,-k)}_M(\mathbf{x}),y\right).
\]
\STATE Partition $\mathcal{D}^{(s)}$ into $K$ disjoint folds $\mathcal{D}^{(s)}_1,\ldots,\mathcal{D}^{(s)}_K$.
For each fold $k$, train $M$ on $\mathcal{D}^{(s)}\setminus \mathcal{D}^{(s)}_k$ to obtain $\hat{f}^{(s,-k)}_M$.
Compute the $K$-fold cross-validation error on $\mathcal{D}^{(s)}$ as
\[
e^{(s)}_M
=
\frac{1}{K}
\sum_{k=1}^K
\frac{1}{|\mathcal{D}^{(s)}_k|}
\sum_{(\mathbf{x},y)\in\mathcal{D}^{(s)}_k}
\ell\!\left(\hat{f}^{(s,-k)}_M(\mathbf{x}),y\right).
\]
\STATE \textbf{Output:}
\[
\widehat{\mathrm{CLS}}_M
=
\tfrac{1}{2}\left(e^{(s \to t)}_M - e^{(t)}_M \right)
+
\tfrac{1}{2}\left(e^{(t \to s)}_M - e^{(s)}_M \right).
\]
\end{algorithmic}
\end{algorithm}


The choice of model $M$ in Algorithm 1 is important, as a poor choice may yield unreliable CLS estimates. A more robust alternative employs multiple models $\{M_1,\ldots,M_l\}$, estimate the CLS value based on each model, and then aggregate their values using a weighting scheme. We discuss two such schemes below.

\noindent \textit{Scheme 1 (Weighted Average).}  
Compute individual scores \( \widehat{\mathrm{CLS}}_{M_i} \) and take their convex combination:
\[
\widehat{\mathrm{CLS}}_{A} = \sum_{i=1}^l w_{M_i}\,\widehat{\mathrm{CLS}}_{M_i},
\]
where \( w_{M_i} \ge 0 \) and \( \sum_i w_{M_i} = 1 \). 
We evaluate each model's performance using cross-validation error and assign higher weights to better models. Specifically,
we use \(w_{M_i} = (w_{M_i}^{(t)} + w_{M_i}^{(s)})/2\), where the weights \( w_{M_i}^{(t)} \) and \( w_{M_i}^{(s)} \) given in Steps~2 and~6 of Algorithm~\ref{alg:ensemble}. 
\medskip

\noindent \textit{Scheme 2 (Weighted ensemble).}  
Instead of aggregating multiple CLS, we can ensemble predictions with dataset-specific weights obtained via cross-validation. Algorithm~\ref{alg:ensemble} shows details. Here $\lambda>0$ sharpens the softmax weighting; larger $\lambda$ concentrates weight on lower-CV-error models. We experimented with a range of $\lambda$ values and found $\lambda = 500$ a reasonable choice. 

\begin{algorithm}[!ht]
\caption{Weighted Ensemble CLS Estimation}
\label{alg:ensemble}
\begin{algorithmic}[1]
\STATE \textbf{Input:} datasets $\mathcal{D}^{(t)}$, $\mathcal{D}^{(s)}$, tuning parameter $\lambda$
\STATE Compute the cross-validation error $E_i^{(t)} \equiv E_{M_i}^{(t),cv}$ for each model $M_i$, $i=1,\ldots,l$, on $\mathcal{D}^{(t)}$.  
Define the sample mean and standard deviation across models as
\[
\bar E^{(t)} = \frac{1}{l}\sum_{i=1}^l E_i^{(t)}, 
\qquad
s_E^{(t)} = \sqrt{\frac{1}{l-1}\sum_{i=1}^l \left(E_i^{(t)} - \bar E^{(t)}\right)^2 }.
\]
Define the ensemble weights
\[
w_{M_i}^{(t)} =
\frac{\exp\!\left(-\lambda \, s_E^{(t)} \, E_i^{(t)}\right)}
{\sum_{j=1}^l \exp\!\left(-\lambda \, s_E^{(t)} \, E_j^{(t)}\right)}.
\]
\STATE Train each $M_i$ on $\mathcal{D}^{(t)}$ and form the ensemble predictor
\[
\hat{f}^{(t)}(x) = \sum_{i=1}^l w_{M_i}^{(t)} \hat{f}^{(t)}_{M_i}(x).
\]
\STATE Evaluate it on $\mathcal{D}^{(s)}$ to obtain
\[
e^{(t \to s)}_E = \frac{1}{n_s}\sum_{r=1}^{n_s}
\ell\!\left(\hat{f}^{(t)}(\mathbf{x}^{(s)}_r), y^{(s)}_r\right).
\]
\STATE Compute the $K$-fold cross-validation error of the ensemble on $\mathcal{D}^{(t)}$:
\[
e^{(t)}_E
=
\frac{1}{K}
\sum_{k=1}^K
\frac{1}{|\mathcal{D}^{(t)}_k|}
\sum_{(\mathbf{x},y)\in\mathcal{D}^{(t)}_k}
\ell\!\left(\hat{f}^{(t,-k)}(\mathbf{x}),y\right),
\]
where $\hat{f}^{(t,-k)}$ is constructed by repeating Steps~2--3 using
$\mathcal{D}^{(t)}\setminus \mathcal{D}^{(t)}_k$.
\STATE Repeat Steps 2--5 with the roles of $\mathcal{D}^{(t)}$ and $\mathcal{D}^{(s)}$ swapped to obtain $w_{M_i}^{(s)}$, $e^{(s \to t)}_E$ and $e^{(s)}_E$.
\STATE \textbf{Output:} $\widehat{\mathrm{CLS}}_E = \tfrac{1}{2}\left(e^{(s \to t)}_E - e^{(t)}_E\right) + \tfrac{1}{2}\left(e^{(t \to s)}_E - e^{(s)}_E\right)$.
\end{algorithmic}
\end{algorithm}

\subsection{CLS for Transferability Assessment}

For any pair of target dataset $\mathcal{D}^{(s)}$ and source dataset $\mathcal{D}^{(s)}$,
after computing the CLS, we 
define a decision function $G$ to categorize its transferability into three zones
\[
G =
\begin{cases}
\text{PT (positive transfer)}, & \text{if} ~ \mathrm{CLS}\leq\tau_1, \\
\text{AZ (ambiguous zone)}, & \text{if} ~ \tau_1 < \mathrm{CLS}<\tau_2, \\[4pt]
\text{NT (negative transfer)}, & \text{if} ~ \mathrm{CLS}\geq\tau_2.
\end{cases}
\]

\begin{figure}[t]
    \centering
    \includegraphics[width=0.7\textwidth]{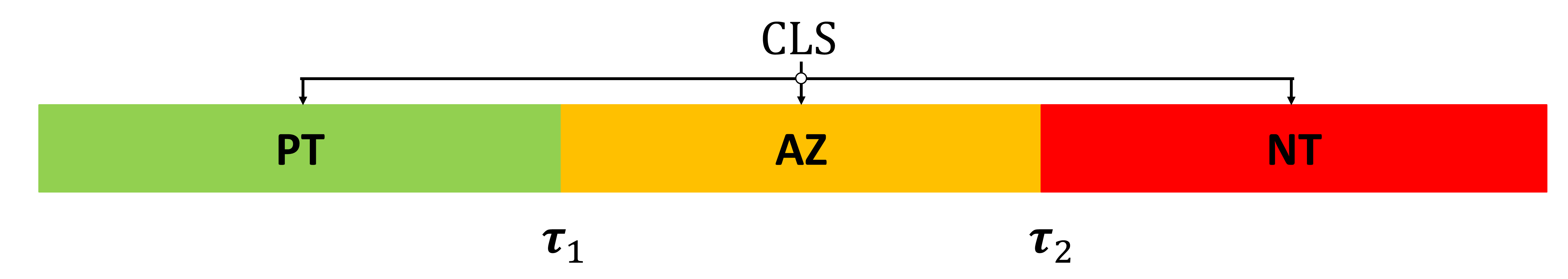}
    \caption{illustration for transferable zones}
    \label{fig:TZ}
\end{figure}

Here positive transfer (PT) indicates high similarity and likely performance gain from transfer, the ambiguous zone (AZ) reflects uncertain benefit, and negative transfer (NT) suggests potential performance degradation \citep{wang2023introduction}. 
To determine the thresholds $\tau_1$ and $\tau_2$, we propose a systematic data-driven procedure.

Let \(e^{(t)}_0\) and \(e^{(s)}_0\) denote the \(k\)-fold cross-validation errors of models trained only on the target dataset \(\mathcal{D}^{(t)}\) and the source dataset \(\mathcal{D}^{(s)}\), respectively. 
We define the baseline error
\[
e_0=\frac{e^{(t)}_0+e^{(s)}_0}{2}.
\]

Let \(\mathrm{SE}(e_0)\) denote the standard error of this quantity. 
Under suitable algorithmic stability conditions, the distribution of \(k\)-fold cross-validation errors can be approximated by a normal distribution \citep{bayle2020cross}. 
Consequently, \(e_0\) is also approximately normal, allowing uncertainty to be characterized using standard errors and normal quantiles.

We define the thresholds as
\[
\hat{\tau}_1=\gamma_1\,\mathrm{SE}(e_0), \qquad
\hat{\tau}_2=\gamma_2\,\mathrm{SE}(e_0),
\]
where \(\gamma_1\) and \(\gamma_2\) control the sensitivity for declaring positive or negative transfer. 
In this work we set \( \gamma_1=0, \gamma_2=\Phi^{-1}(0.9999),\)
where \(\Phi^{-1}(0.9999)\) is the 99.99\% quantile of the standard normal distribution. 
Thus,\(\hat{\tau}_1=0, \hat{\tau}_2=\Phi^{-1}(0.9999)\,\mathrm{SE}(e_0).\)

This choice defines PT relative to the no-transfer baseline \(e_0\) and declares NT only when the CLS score exceeds a conservative upper confidence bound. 
Under the normal approximation, the probability that cross-validation variability alone produces such a deviation is approximately \(10^{-4}\). 
The AZ therefore captures intermediate cases where deviations may arise from sampling variability rather than structural mismatch.

When algorithmic stability conditions are not guaranteed, we approximate
\[
\mathrm{SE}(e_0)\approx
\frac{\sqrt{\mathrm{Var}(e^{(t)}_0)+\mathrm{Var}(e^{(s)}_0)}}{2}
\approx
\frac{\sqrt{s_t^2/k+s_s^2/k}}{2},
\]
where \(s_t\) and \(s_s\) are the empirical standard deviations of the fold errors for the target-only and source-only models.

\subsection{Extensions of CLS to Deep Learning}

The proposed CLS performs well for shallow models (e.g., linear classifiers and SVMs). 
However, it encounters challenges in deep learning settings. 
Modern deep transfer learning typically follows an \textit{encoder--head} paradigm: 
a shared encoder learns feature embedding representations, 
while lightweight task-specific heads adapt to domain-specific label semantics.

A key issue arises because the standard CLS is computed via direct cross-domain testing without adaptation. 
In deep models, the resulting error consists of two components:
(i) a \emph{feature embedding error} caused by representation mismatch, and 
(ii) a \emph{learning task error} arising from the head-level prediction function.
Even when transfer is beneficial, the standard CLS may substantially exceed the 0, 
sometimes surpassing the threshold \(\gamma_2 \cdot \mathrm{SE}(e_0)\), thereby incorrectly classifying the transfer as negative.

This discrepancy occurs because fine-tuning---the most common transfer learning strategy in deep learning---updates not only the task-specific head but also part (or all) of the encoder. 
Such adaptation can significantly reduce the feature embedding error. 
In contrast, the standard CLS does not account for encoder adaptation and therefore overestimates cross-domain error. 
Consequently, it underestimates transferability in deep learning settings.

To address this limitation, we extend CLS to the encoder--head architecture (Algorithm~\ref{alg:enc-head-cls}). 
The procedure consists of two stages:

(1) Stage 1: Train a shared encoder jointly on both datasets to obtain aligned feature representations.

\begin{figure}[t]
\centering
\begin{subfigure}{0.48\textwidth}
    \centering
    \includegraphics[width=\linewidth]{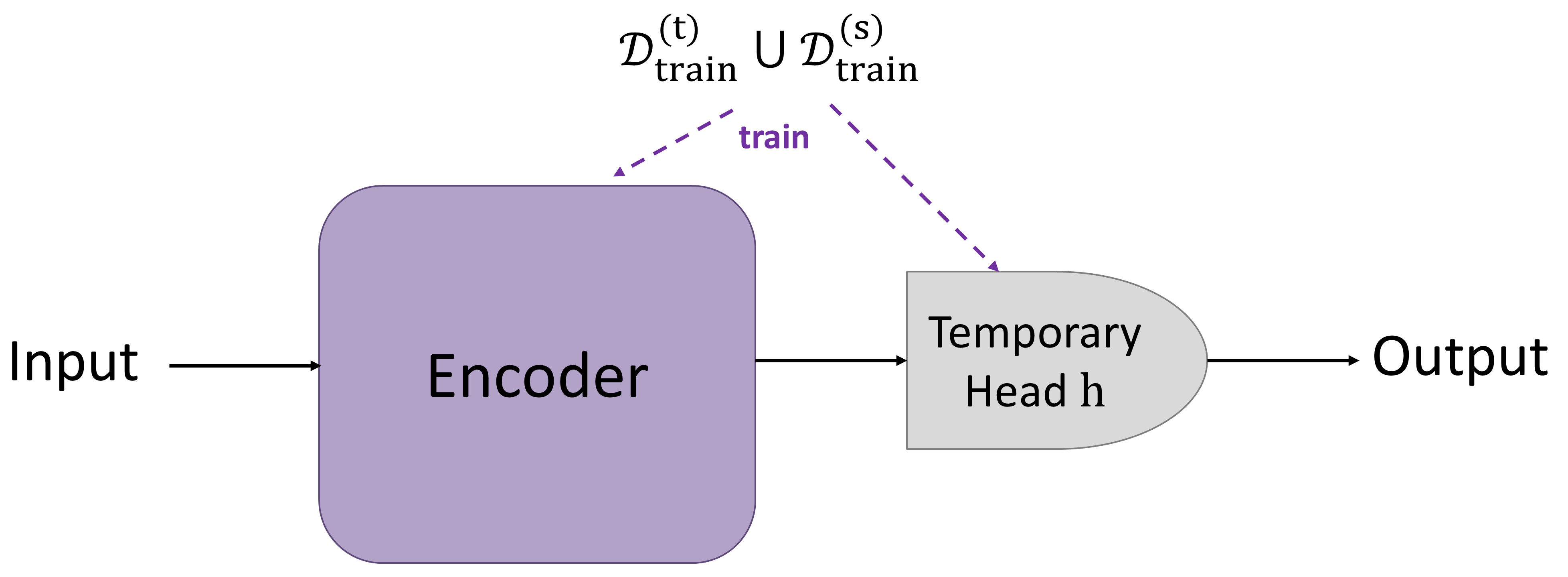}
    \caption{Stage 1}
    \label{fig:DL1}
\end{subfigure}
\hfill
\begin{subfigure}{0.48\textwidth}
    \centering
    \includegraphics[width=\linewidth]{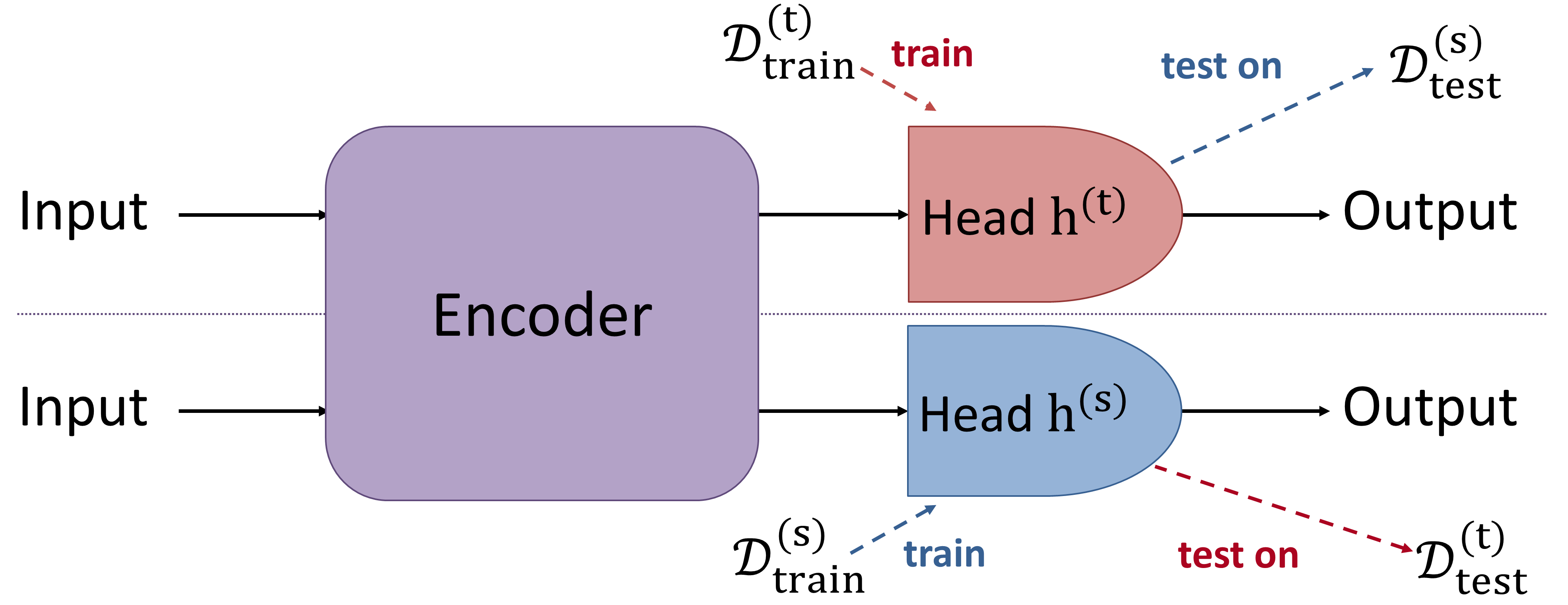}
    \caption{Stage 2}
    \label{fig:DL2}
\end{subfigure}

\caption{Illustration of the encoder--head CLS procedure. 
Stage 1 learns a shared encoder using both datasets, while Stage 2 trains task-specific heads for cross-domain evaluation.}
\label{fig:DL}
\end{figure}

(2) Stage 2: Fix the encoder and train domain-specific heads for cross-domain evaluation, analogous to the standard CLS computation.

By isolating representation learning from task adaptation, 
this extended CLS better reflects the transferability achievable under realistic deep transfer learning protocols.

\begin{algorithm}[t]
\caption{Estimating Encoder--Head CLS}
\label{alg:enc-head-cls}
\begin{algorithmic}[1]
\STATE \textbf{Input:} 
Target dataset $\mathcal{D}^{(t)}=\{\mathcal{D}^{(t)}_{\mathrm{train}},\mathcal{D}^{(t)}_{\mathrm{test}}\}$, 
source dataset $\mathcal{D}^{(s)}=\{\mathcal{D}^{(s)}_{\mathrm{train}},\mathcal{D}^{(s)}_{\mathrm{test}}\}$.

\STATE \textbf{Stage 1: Representation Learning}
\STATE Jointly train encoder $f_\theta$ on 
$\mathcal{D}^{(t)}_{\mathrm{train}} \cup \mathcal{D}^{(s)}_{\mathrm{train}}$ 
using a temporary head (discarded after training).

\STATE \textbf{Stage 2: Cross-Domain Head Evaluation}
\STATE Freeze encoder $f_\theta$.

\STATE Train source head $h^{(s)}$ on $\mathcal{D}^{(s)}_{\mathrm{train}}$.
\STATE Compute
\[
e^{(s \to t)}
= \frac{1}{|\mathcal{D}^{(t)}_{\mathrm{test}}|}
\sum_{(\mathbf{x},y)\in\mathcal{D}^{(t)}_{\mathrm{test}}}
\ell\!\left(h^{(s)}(f_\theta(\mathbf{x})), y\right),
\]
\[
e^{(s)}_0
= \frac{1}{|\mathcal{D}^{(s)}_{\mathrm{test}}|}
\sum_{(\mathbf{x},y)\in\mathcal{D}^{(s)}_{\mathrm{test}}}
\ell\!\left(h^{(s)}(f_\theta(\mathbf{x})), y\right).
\]

\STATE Train target head $h^{(t)}$ on $\mathcal{D}^{(t)}_{\mathrm{train}}$.
\STATE Compute
\[
e^{(t \to s)}
= \frac{1}{|\mathcal{D}^{(s)}_{\mathrm{test}}|}
\sum_{(\mathbf{x},y)\in\mathcal{D}^{(s)}_{\mathrm{test}}}
\ell\!\left(h^{(t)}(f_\theta(\mathbf{x})), y\right),
\]
\[
e^{(t)}_0
= \frac{1}{|\mathcal{D}^{(t)}_{\mathrm{test}}|}
\sum_{(\mathbf{x},y)\in\mathcal{D}^{(t)}_{\mathrm{test}}}
\ell\!\left(h^{(t)}(f_\theta(\mathbf{x})), y\right).
\]

\STATE Compute
\[
\widehat{\mathrm{CLS}}_{\mathrm{Enc\text{--}Head}}
=
\frac{1}{2}\big(e^{(t \to s)} + e^{(s \to t)}\big)
-
\frac{1}{2}\big(e^{(t)}_0 + e^{(s)}_0\big).
\]

\STATE \textbf{Output:} 
$\widehat{\mathrm{CLS}}_{\mathrm{Enc\text{--}Head}}$.
\end{algorithmic}
\end{algorithm}

$\mathrm{CLS}_{\text{Enc--Head}}$ better captures transferability in deep learning,  as the encoder learns shared representations while the heads adapt to domain-specific labels. This structure aligns with practical transfer pipelines such as pretraining on 
$\mathcal{D}^{(s)}$ followed by fine-tuning on $\mathcal{D}^{(t)}$.

\section{Numerical Experiments}
We conduct synthetic experiments to evaluate the effectiveness of CLS in quantifying dataset similarity for supervised learning. 

For binary classification, we consider four settings: (i) logistic regression, (ii) probit regression, (iii) linear discriminant analysis (LDA), and (iv) a mixture Gaussian model that induces distribution shift through a mixture parameter $\alpha$. 
For multi-class classification, we construct a four-class problem in which samples are projected onto two orthogonal directions and assigned to quadrants according to the projection signs with Gaussian noise perturbation. 
For regression, we consider linear regression and nonlinear regression models with sinusoidal, polynomial, interaction, and exponential transformations with Gaussian noise. 
Detailed data-generation procedures are provided in the Supplemental Material.

To estimate CLS, we employ several commonly used models, including logistic regression, multinomial logistic regression, linear SVM, radial SVM, XGBoost, and linear regression, depending on the learning task. 
In all experiments, we set $n_t = n_s = 200$, perform 50 replicates, and report average results. 
Hyperparameter tuning yields negligible improvements but substantially increases computational cost; therefore, we use default model configurations.

\subsection{Relationships of CLS and Data Similarity}

We investigate the relationship between CLS and dataset similarity. For binary classification settings (i)–(iii), the multi-class example, and regression settings (i)–(iii), we control the level of similarity by adjusting the \emph{cosine similarity} between the model parameters of the target and source distributions. A larger cosine similarity indicates that the target and source data are more similar. For binary classification setting (iv), we control the similarity using the mixing coefficient $\alpha \in [0,1]$. A smaller value of $\alpha$ corresponds to higher similarity between the target and source data. Detailed experimental settings are provided in the Supplemental Material.

We report three variants of CLS: the weighted average version (\textit{Score~Weighted~Avg.}), the weighted ensemble version (\textit{Score~Ensemble}), and the oracle version (\textit{Score~Oracle}), which assumes access to the true feature--label mapping. We compare CLS with several existing dataset similarity measures, including KL Divergence, Wasserstein Distance, and Optimal Transport Dataset Distance (OTDD). KL Divergence and Wasserstein Distance measure similarity at the feature distribution level, while OTDD additionally incorporates label structure information.

\begin{figure}[t]
    \begin{center}
    \includegraphics[width=0.7\textwidth]{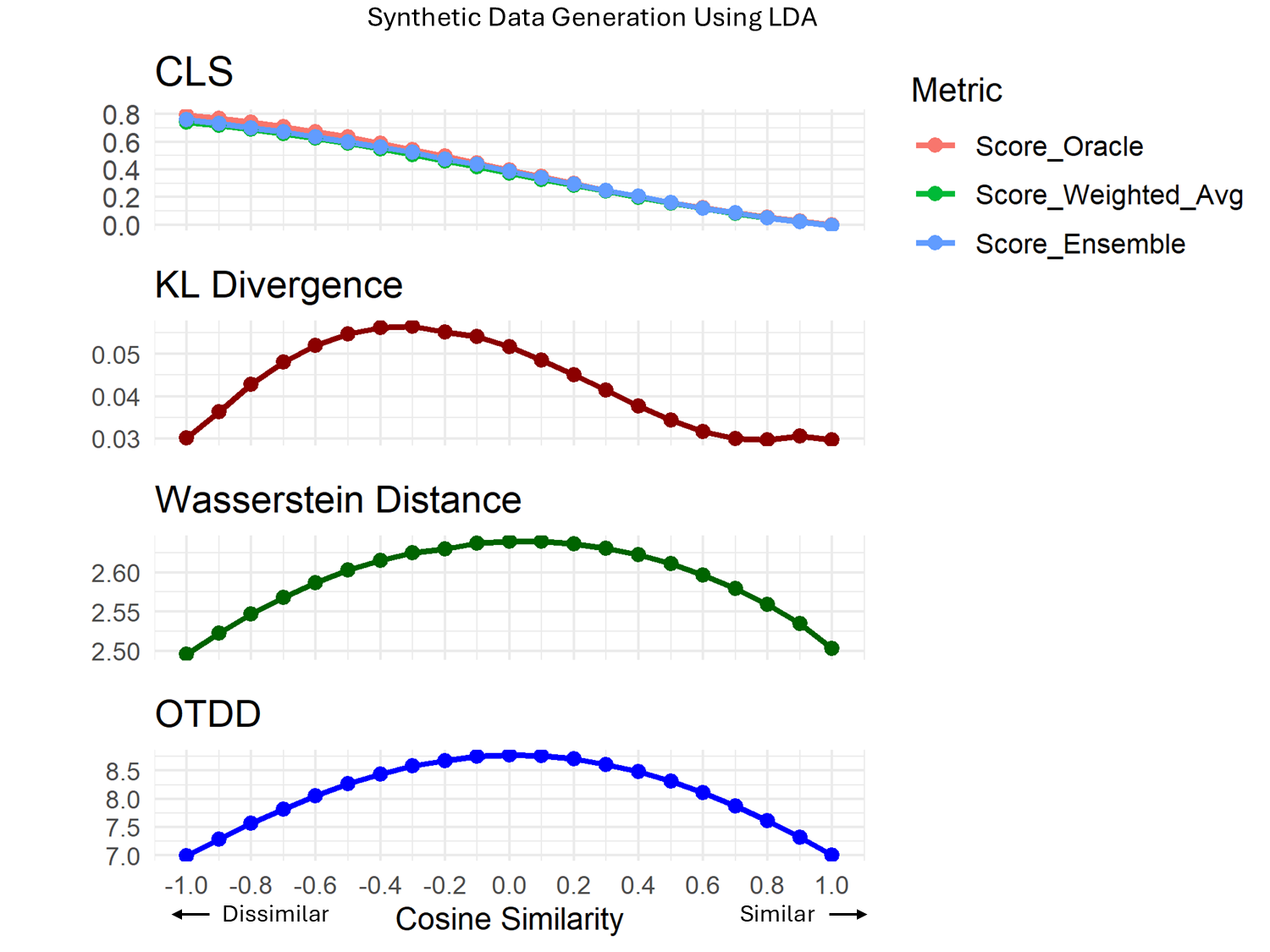}
    \caption{ Comparison of CLS vs other similarity metrics under the LDA setting. }
    \label{fig:num_LDA}
    \end{center}
\end{figure}

Figure~\ref{fig:num_LDA} illustrates the relationship between CLS and cosine similarity under the binary classification setting (iii) (LDA). 
The x-axis represents the cosine similarity between the target and source parameters, and the y-axis shows the corresponding similarity scores.
In the top panel, the curves correspond to the oracle CLS (red), the weighted average estimate (green), and the ensemble estimate (blue). 
CLS decreases monotonically as cosine similarity increases, and both estimated CLS curves closely follow the oracle curve, indicating that the proposed estimation methods provide accurate approximations. The remaining panels compare CLS with other similarity measures, including KL Divergence, Wasserstein Distance, and Optimal Transport Dataset Distance (OTDD). 
Under this LDA setting, none of these alternative metrics shows a clear monotonic relationship with cosine similarity. The results and figures for other experimental settings are provided in the Supplemental Material.

To further quantify these relationships, we compute the absolute value of the Spearman rank correlation coefficient between each similarity metric and the similarity level within each setting. The results are summarized in Table~\ref{tab:cls_results_abs}. 
Values closer to 1 indicate stronger monotonic relationships, while values closer to 0 indicate little or no association. Across all the settings, three CLS variants consistently achieve a value of 1, indicating a perfect monotonic relationship with the similarity level. This observation is consistent with Theorems 1 and 2, even under more complex data-generating processes. 

For KL Divergence and Wasserstein Distance, the correlation values are relatively high (but still less than 1) under the Mixture Gaussian setting. 
This suggests that these metrics partially capture similarity changes in that setting. 
However, in several other settings, the feature distributions remain the same while only the label distribution changes as similarity varies. 
In those cases, KL Divergence and Wasserstein Distance remain constant, leading to a Spearman correlation of 0. For OTDD, the performance is strong in the Mixture Gaussian and 4-class classification settings, where it successfully reflects changes in similarity. However, its performance is less stable in other classification settings. In addition, the original OTDD formulation cannot be directly applied to regression problems.

\begin{table}[t]
\centering
\begin{tabular}{l||c|c|c||c|c|c|c}
\toprule
Setting 
& Oracle~CLS
& Wtd.~Avg.~CLS
& Ens.~CLS
& KL Div. 
& Wass. Dist. 
& OTDD \\
\midrule
Logistic Reg. 
& 1.000 & 1.000 & 1.000 & 0.000 & 0.000 & 0.156 \\

Probit Model 
& 1.000 & 1.000 & 1.000 & 0.000 & 0.000 & 0.035 \\

LDA 
& 1.000 & 1.000 & 1.000 & 0.530 & 0.078 & 0.071 \\

Mixture Gaus. 
& 1.000 & 1.000 & 1.000 & 0.791 & 0.973 & 0.990 \\

Linear Regr. 
& 1.000 & 1.000 & 1.000 & 0.000 & 0.000 & --- \\

Nonlinear Regr. 
& 1.000 & 1.000 & 1.000 & 0.000 & 0.000 & --- \\

4-Class Class. 
& 1.000 & 1.000 & 1.000 & 0.000 & 0.000 & 1.000 \\
\bottomrule
\end{tabular}
\caption{Absolute values of Spearman correlation coefficients between CLS and data similarity signals ($\alpha$ for the Mixture Gaussian setting and cosine similarity for other settings).}
\label{tab:cls_results_abs}
\end{table}

\subsection{Comparison of CLS Estimation}

This experiment compares different CLS estimation methods. Since the true data-generating mechanism is known, we compute the oracle CLS as the gold standard. Table~\ref{tab:diff-common} reports \textit{Diff}, defined as the average absolute deviation across all similarity levels:
\[
\textit{Diff} = \frac1h\sum_{i=1}^h \big| \widehat{\mathrm{CLS}}_i - \text{oracle CLS}_i \big|,
\]
where index $i$ ranges over similarity levels from low to high. A smaller \textit{Diff} indicates a more accurate CLS estimate. The first four columns present single-model estimates based on generalized linear models, linear SVM (\textit{SVM-L}), radial kernel SVM (\textit{SVM-R}), and XGBoost. 
The last three columns report multi-model estimates obtained using three aggregation schemes: unweighted average (\textit{Unw.\ Avg.}), weighted average (\textit{Wtd.\ Avg.}), and ensemble estimate (\textit{Ens.}).  To investigate nonlinear scenarios, we also include QDA, PLR, and Radial settings in this experiment. The implementation details are provided in the supplementary materials.

\begin{table*}[t]
\centering
\setlength{\tabcolsep}{6pt}
\renewcommand{\arraystretch}{0.8}
\small
\caption{$\text{Diff}$ between $\widehat{\mathrm{CLS}}$ and oracle CLS across all settings (Binary Classification, Multi-Class Classification, and Regression). Lower values indicate smaller deviation from Oracle CLS.}
\label{tab:diff-common}
\resizebox{0.9\linewidth}{!}{
\begin{tabular}{l||cccc|ccc}
\toprule
\multicolumn{8}{c}{\textbf{Binary Classification}} \\
\midrule
\textbf{Setting} & \textbf{Log.Regr.} & \textbf{SVM-L} & \textbf{SVM-R} & \textbf{Xgb} & \textbf{Unw.Avg} & \textbf{Wtd.Avg} & \textbf{Ens.} \\
\midrule
Logistic Regr.   
& 0.0256 & 0.0302 & 0.0351 & 0.0680 & 0.0397 & 0.0365 & \textbf{0.0291} \\

Probit           
& 0.0164 & 0.0217 & 0.0388 & 0.0792 & 0.0390 & 0.0324 & \textbf{0.0239} \\

LDA              
& 0.0192 & 0.0185 & 0.0312 & 0.0486 & 0.0274 & 0.0257 & \textbf{0.0162} \\

Mixture Gaussian 
& 0.0118 & 0.0125 & 0.0152 & 0.0377 & 0.0192 & 0.0167 & \textbf{0.0131} \\

QDA
& 0.0754 & 0.0693 & 0.0216 & 0.0018 & 0.0303 & 0.0279 & \textbf{0.0255} \\

PLR
& 0.0383 & 0.0393 & 0.0288 & 0.0238 & 0.0326 & 0.0309 & \textbf{0.0167} \\

Radial
& 0.0816 & 0.0884 & 0.0537 & 0.085 & 0.0772 & 0.0537 & \textbf{0.0536} \\

\midrule\midrule
\multicolumn{8}{c}{\textbf{Multi-Class Classification}} \\
\midrule
\textbf{Setting} & \textbf{MLR} & \textbf{SVM-L} & \textbf{SVM-R} & \textbf{Xgb} & \textbf{Unw.Avg} & \textbf{Wtd.Avg} & \textbf{Ens.} \\
\midrule
4-Class Class.   
& 0.0488 & 0.0564 & 0.1008 & 0.1560 & 0.0905 & 0.0628 & \textbf{0.0486} \\

\midrule\midrule
\multicolumn{8}{c}{\textbf{Regression}} \\
\midrule
\textbf{Setting} & \textbf{Lin.Regr.} & \textbf{SVM-L} & \textbf{SVM-R} & \textbf{Xgb} & \textbf{Unw.Avg} & \textbf{Wtd.Avg} & \textbf{Ens.} \\
\midrule
Linear Regr.     
& 0.0562 & 0.0651 & 1.1909 & 1.0826 & 0.5987 & \textbf{0.0607} & 0.0661 \\

Nonlinear Regr.  
& 25.1714 & 34.1827 & 20.2443 & 13.3547 & 23.2383 & 14.5025 & \textbf{14.1004} \\

\bottomrule
\end{tabular}
}
\end{table*}

\begin{figure}[H]
    \centering
    \includegraphics[width=0.8\textwidth]{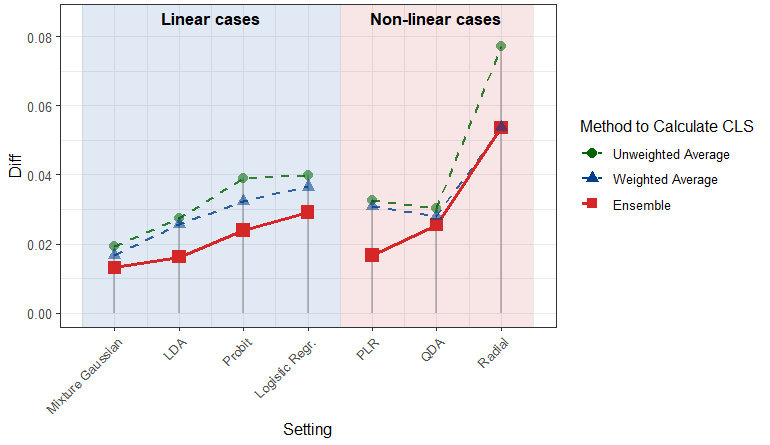}
    \caption{Deviation from the oracle CLS for aggregation methods. Diff is defined as in Table~\ref{tab:diff-common}.}
    \label{fig:diff1}
\end{figure}

\begin{figure}[t]
    \centering
    \includegraphics[width=0.8\textwidth]{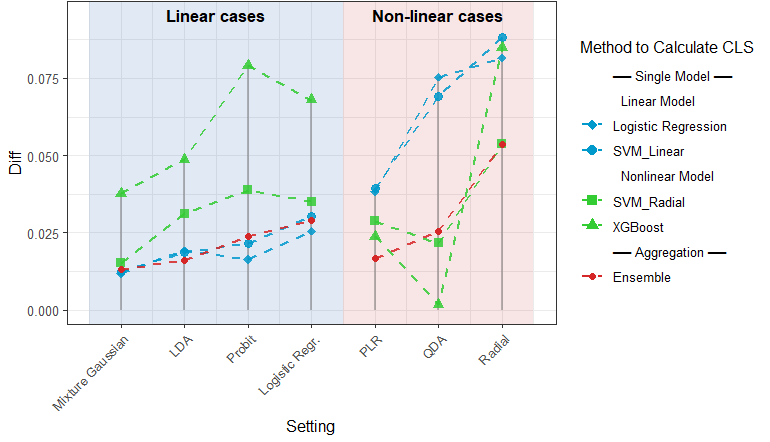}
    \caption{Deviation from oracle CLS across different settings. Diff (reported in Table~\ref{tab:diff-common}) is defined as the average absolute deviation from the oracle CLS across similarity levels.}
    \label{fig:diff2}
\end{figure}

To make the comparison clearer, we plot the binary classification results in Table~\ref{tab:diff-common} as Figure~\ref{fig:diff1} and Figure~\ref{fig:diff2}. Several observations can be drawn from Table~\ref{tab:diff-common},
Figure~\ref{fig:diff1}, and Figure~\ref{fig:diff2}.
First, among the three aggregation methods, the ensemble estimator
(\textit{Ens.}) consistently achieves the smallest \textit{Diff},
followed by the weighted average, while the unweighted average
performs the worst (Figure~\ref{fig:diff1}). This pattern is also
confirmed by Table~\ref{tab:diff-common}. Although the weighted
average slightly outperforms the ensemble method in the linear
regression setting, the difference is small.

Second, the performance of single-model estimators depends strongly
on whether the model matches the underlying data structure
(Figure~\ref{fig:diff2} and Table~\ref{tab:diff-common}). For linear
problems (logistic regression, probit model, LDA, 4-class
classification, and linear regression), linear models such as
logistic regression and SVM-L achieve smaller \textit{Diff} values
than nonlinear models. In contrast, for nonlinear problems (e.g.,
nonlinear regression and other nonlinear classification settings),
nonlinear models such as SVM-R and XGBoost perform better.

In practice, the true data structure is usually unknown.
Consequently, the ensemble method provides a reliable and stable
solution, as it combines multiple models and mitigates the impact
of poorly fitted ones (Figure~\ref{fig:diff2}).

\subsection{Relationships of CLS and Relative Error Reduction(RER)} \label{RER}

In this section, we examine the relationship between CLS and dataset transferability. Here, 
we measure transferability by the gain in performance when applying transfer learning methods that use both target and source data, compared to training on a single dataset only.
To make the results comparable across different dataset pairs, we report the relative error reduction (RER) in classification error. 
For the target dataset, we define $\textit{Target RER} = \frac{ \mathrm{error}(D^{(t)}) - \mathrm{error}(D^{(s)} \to D^{(t)}) }{ \mathrm{error}(D^{(t)}) }$, where $\mathrm{error}(D^{(t)})$ is the classification error obtained by training only on the target data, and $\mathrm{error}(D^{(s)} \to D^{(t)})$ is the classification error on the target test set after applying transfer learning algorithm using $D^{(s)}$ to assist $D^{(t)}$. Similarly, for the source dataset, we define $\textit{Source RER} = \frac{ \mathrm{error}(D^{(s)}) - \mathrm{error}(D^{(t)} \to D^{(s)}) } { \mathrm{error}(D^{(s)}) }$, where the roles of target and source are reversed. We further define the overall transferability measure as $ \textit{Average RER} = \frac{\textit{Target RER} + \textit{Source RER}}{2}$.

\begin{figure}[H]
    \centering
    \includegraphics[width=0.8\textwidth]{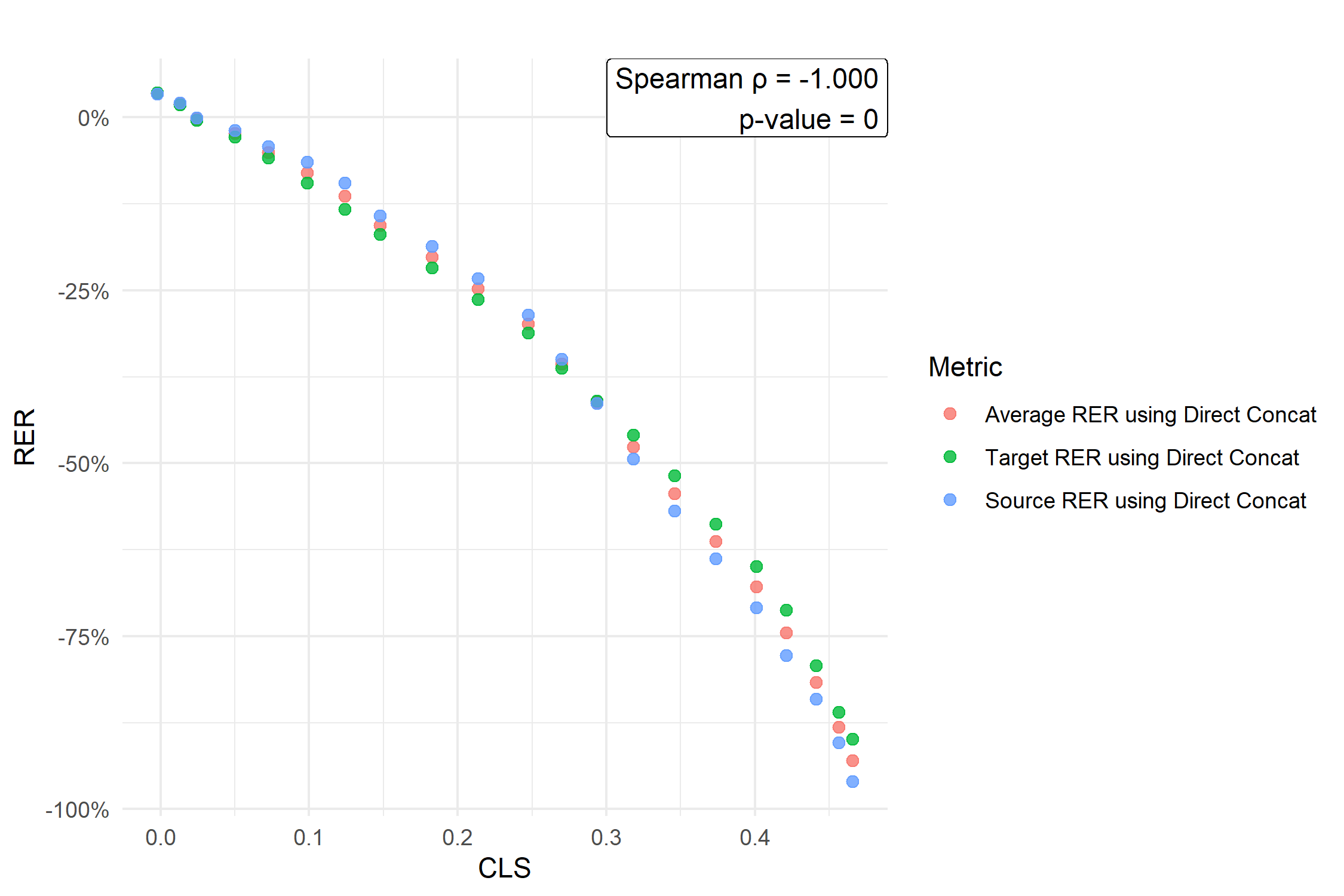}
    \caption{Relative Error Reduction(RER) v.s. CLS under the logit model. The Spearman rank correlation coefficient and the p-value of corresponding test is also shown.}
    \label{fig:CLS_logit}
\end{figure}

We first consider a simple and direct transfer learning strategy: direct concatenation, where the target and source datasets are combined and used to train a single model. Figure~\ref{fig:CLS_logit} shows the relationship between CLS and the three RER measures under the logistic regression setting. 
As CLS increases, all three RER values decrease monotonically. 
This indicates that CLS effectively captures dataset transferability: higher CLS corresponds to lower performance gain from transfer learning. The 
results and corresponding figures for other experimental settings are provided in the Supplemental Material.  In all settings, RER decreases monotonically as CLS increases.

To evaluate the proposed transferability zones, we fix the target dataset and generate source datasets with varying levels of similarity, controlled by cosine similarity or the mismatch parameter $\alpha$. 
Each source dataset is assigned to one of the three zones based on the proposed CLS thresholds.
At each similarity level, in addition to direct concatenation, we apply three additional transfer learning methods: TrAdaBoost~\citep{dai2007boosting}, Dynamic TrAdaBoost~\citep{al2011adaptive}, and Adaptive Robust Transfer Learning (ART)~\citep{wang2023art}. 
For each method, we compute the Average RER. 
All experiments are repeated 50 times.

\begin{figure}[t]
    \centering
    \includegraphics[width=0.6\textwidth]{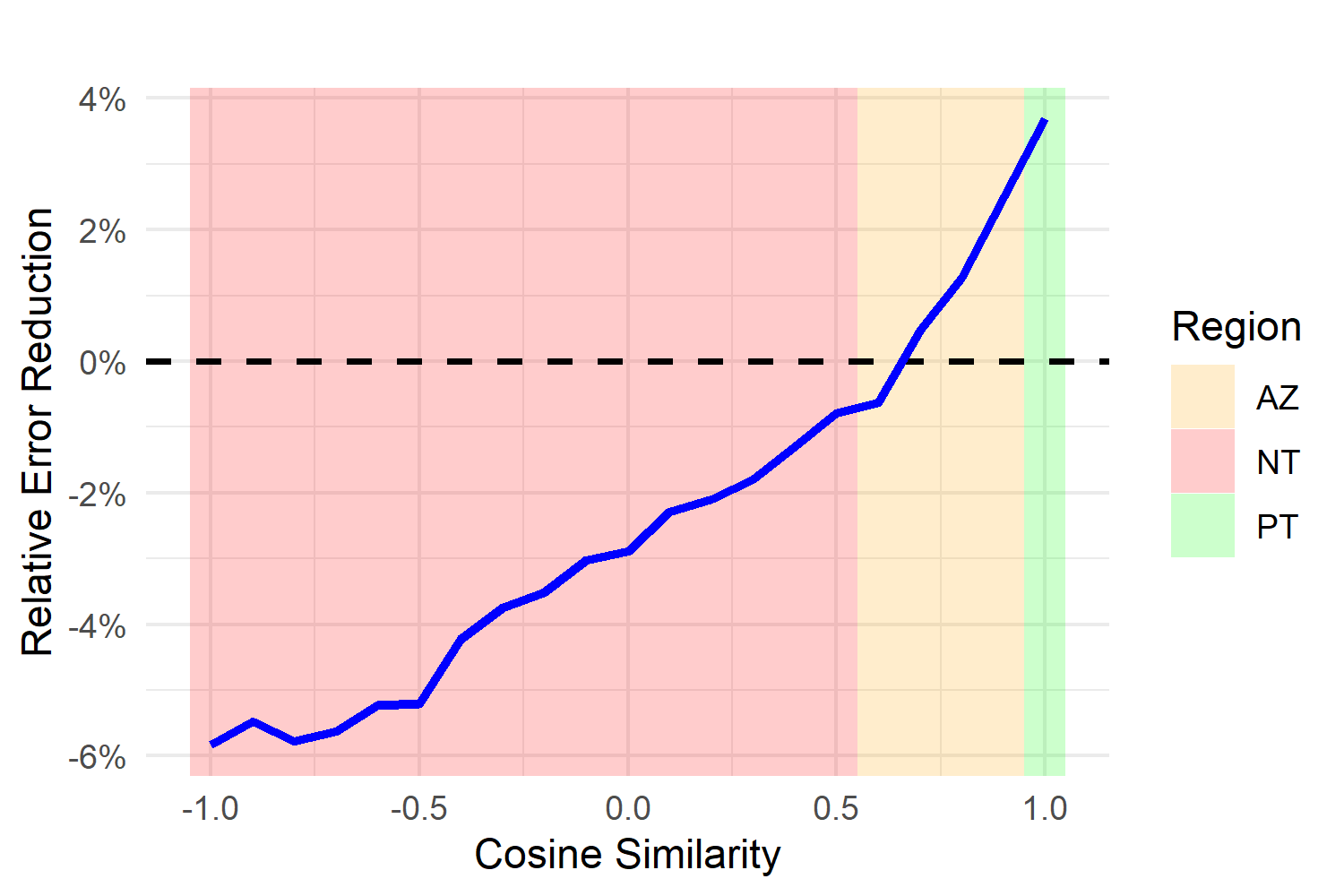}
    \caption{Transferability zones (positive, ambiguous, negative) identified under the logistic model.}
    \label{fig:TL_logit}
\end{figure}

Figure~\ref{fig:TL_logit} shows the average RER of the best-performing transfer learning method (among the four methods) at each similarity level under logistic regression. 
The background colors indicate the predicted transferability zone (PT, AZ, or NT).
In the NT region, the average RER is consistently below 0, indicating that transfer learning does not provide performance improvement. 
In the AZ region, the average RER gradually increases, transitioning from negative to positive values. This suggests that transfer learning begins to provide benefit as similarity increases. 
In the PT region, the average RER reaches its largest positive values, indicating strong transferability. The results and figures for other experimental settings are provided in the Supplemental Material. Across all settings, we observe the same pattern, confirming that the CLS-based zones consistently align with empirical transferability.

\section{Applications to Real-world Data}

\subsection{eICU Mortality Prediction}

ICUs provide critical care for life-threatening conditions, but at high cost, which motivates the need for accurate post-ICU survival prediction 
\citep{halpern2015critical,stacy2011progressive}. 
Using the \emph{eICU Collaborative Research Database} \citep{pollard2018eicu}, we construct features from demographic information, hospital characteristics, and admission physiological variables after data cleaning and class balancing. 
The hospital with the highest post-ICU mortality rate (9.47\%) is selected as the target dataset, and 10 other hospitals are treated as source datasets.

We evaluate dataset similarity using the \emph{ensemble CLS}, based on logistic regression, linear and radial SVMs, and XGBoost, and map each source dataset $\mathcal{D}^{(s)}$ to a transferability zone. 
Among the 10 source datasets, 6 fall into the \emph{positive transfer zone} (PT), suggesting strong potential transfer benefit, while the remaining 4 fall into the \emph{ambiguous zone} (AZ), indicating uncertain transferability.

To validate these predictions, we compute the Average Relative Error Reduction (RER) defined in Section~\ref{RER}. 
For each dataset, we randomly split $\mathcal{D}^{(t)}$ or $\mathcal{D}^{(s)}$ into 70\% training data and 30\% test data, and repeat the evaluation 50 times. Following \citet{wang2023art}, we adopt Random Forest as the base classifier for all transfer learning methods. 
We compare four transfer learning approaches: Direct Concatenation, TrAdaBoost, Dynamic TrAdaBoost (DTrAdaBoost), and ART, across all 10 source datasets.

Table~\ref{tab:eicu_results_relred} reports the Average RER for each source dataset under the four transfer learning methods. 
A positive Average RER indicates that transfer learning improves performance (i.e., reduces classification error), while a negative value indicates no improvement or performance degradation. For all source datasets classified into the PT zone, at least two transfer learning methods improve performance, and Direct Concatenation consistently yields performance gains.  For the datasets in the AZ zone, datasets 3 and 7 show performance improvement under all four methods, while for datasets 5 and 8, only one of the four methods achieves improvement.

Overall, the empirical transfer learning results are consistent with the ensemble CLS predictions. 
The PT datasets consistently demonstrate transfer benefits, while the AZ datasets exhibit mixed outcomes, reflecting their intermediate transferability.

\begin{table}[t]
\centering
\renewcommand{\arraystretch}{0.8}
\caption{Average relative error reduction (RER) on the target hospital across 10 source hospitals.}
\label{tab:eicu_results_relred}
\resizebox{0.7\linewidth}{!}{
\begin{tabular}{l|cccc|c}
\toprule
Src. & Direct Concat & ART & TrAdaboost & DTrAdaboost & Zone \\
\midrule
1  & \textbf{+4.1\%} & \textbf{+1.1\%} & \textbf{+7.5\%} & -5.4\% & PT \\
2  & \textbf{+7.3\%} & \textbf{+1.0\%} & -1.9\% & -4.5\% & PT \\
3  & \textbf{+4.0\%} & \textbf{+1.2\%} & \textbf{+1.1\%} & \textbf{+10.1\%} & AZ \\
4  & \textbf{+5.3\%} & \textbf{+7.7\%} & -0.3\% & \textbf{+0.5\%} & PT \\
5  & \textbf{+4.1\%} & -1.5\% & -11.1\% & -5.0\% & AZ \\
6  & \textbf{+4.8\%} & \textbf{+6.5\%} & -14.4\% & \textbf{+4.8\%} & PT \\
7  & \textbf{+4.9\%} & \textbf{+7.6\%} & \textbf{+0.2\%} & \textbf{+6.1\%} & AZ \\
8  & -0.9\% & \textbf{+3.4\%} & -2.3\% & -7.5\% & AZ \\
9  & \textbf{+7.3\%} & \textbf{+1.0\%} & -6.4\% & \textbf{+2.0\%} & PT \\
10 & \textbf{+6.6\%} & \textbf{+6.4\%} & \textbf{+0.8\%} & \textbf{+5.8\%} & PT \\
\bottomrule
\end{tabular}
}
\end{table}

\subsection{Canine Image Classification}

Canine image classification, particularly distinguishing dogs from wolves, provides a challenging testbed for evaluating transferability in vision tasks. 
Subtle inter-class differences and background confounding effects (e.g., snow biasing predictions toward wolves) make this task more difficult than standard binary classification. 
We use the \emph{Roboflow Dogs vs.\ Wolves} dataset as the target dataset $\mathcal{D}^{(t)}$, due to its small sample size, class imbalance, and visual ambiguity. 

We consider three source domains with varying semantic similarity:
(i) \emph{Kaggle Dogs vs.\ Wolves} (1{,}000 samples per class),
(ii) \emph{Kaggle Cats vs.\ Dogs} (sampled to 1{,}000 per class), and
(iii) \emph{Kaggle Horses vs.\ Camels} (200 per class). We employ the \emph{Encoder--Head CLS} (Algorithm~\ref{alg:enc-head-cls}) using ResNet-18 as the encoder, initialized with ImageNet weights for stable convergence. A shared encoder is jointly trained on $\mathcal{D}^{(t)}$ and $\mathcal{D}^{(s)}$, followed by domain-specific heads for cross-domain evaluation. 
Each experiment is repeated 10 times, and we report the average $\widehat{\mathrm{CLS}}_{\text{Enc--Head}}$.
Based on the predicted transferability zones, \emph{Kaggle Dogs vs.\ Wolves} falls into the ambiguous zone (AZ), whereas \emph{Kaggle Cats vs.\ Dogs} and \emph{Kaggle Horses vs.\ Camels} are classified into the negative transfer zone (NT).

To evaluate whether the source data improves performance, we adopt a standard deep transfer learning setup: a ResNet-18 model is first pretrained on each $\mathcal{D}^{(s)}$ (initialized from ImageNet weights) and then fine-tuned on $\mathcal{D}^{(t)}$. 
We compute the Target RER defined in Section~\ref{RER}, as well as the Source RER and the Average RER. Table~\ref{tab:dogs_results_relred} reports the relative error reduction, computed in the same manner as in Section~\ref{RER}. 

\begin{table}[H]
\centering
\renewcommand{\arraystretch}{0.8}
\caption{Relative error reduction (RER) on Dogs vs Wolves when using transfer learning}
\label{tab:dogs_results_relred}
\resizebox{0.7\linewidth}{!}{
\begin{tabular}{l|ccc|c}
\toprule
Source Dataset & Target RER & Source RER & Average RER & Zone \\
\midrule
Dogs vs Wolves    & \textbf{+5.0\%}   & -11.0\%    & -3.0\%   & AZ \\
Cats vs Dogs      & -7.9\%   & -4.7\%    & -6.3\%   & NT \\
Horses vs Camels  & -9.0\%   & -18.2\%    & -13.6\%   & NT \\
\bottomrule
\end{tabular}
}
\end{table}

We make the following observations. For \emph{Kaggle Dogs vs.\ Wolves}, the Target RER is positive (+5.0\%), while the Source RER is negative ($-11.1\%$). This indicates that using this source dataset improves the target performance, but the reverse transfer does not yield improvement. Its classification into the ambiguous zone (AZ) is therefore consistent with the empirical results. For \emph{Kaggle Cats vs.\ Dogs} and \emph{Kaggle Horses vs.\ Camels}, all RER values are negative, indicating that transfer learning does not improve performance in either direction. Both datasets are classified into the negative transfer zone (NT), which aligns with the observed transfer outcomes.

Overall, these results demonstrate that the proposed CLS framework can effectively predict both beneficial and detrimental transfer in real-world vision tasks.

\section{Conclusion}

In this work, we propose the Cross-Learning Score (CLS), a metric for quantifying dataset similarity through bidirectional generalization performance. 
Unlike traditional distribution-based measures that focus primarily on feature alignment, CLS directly evaluates the similarity of feature–response relationships that determine predictive transferability.

We establish theoretical foundations of CLS under canonical linear settings, showing that it is closely related to cosine similarity between decision boundaries---a geometric interpretation that explains its ability to capture structural alignment between tasks. Extensive synthetic experiments further demonstrate that CLS maintains a monotonic relationship with controlled similarity levels across diverse settings, outperforming KL divergence, Wasserstein distance, and OTDD especially when feature distributions are similar but label structures differ.

To improve practical reliability, we develop stable estimation procedures including weighted averaging and ensemble estimators, with the ensemble variant performing particularly well across diverse data-generating mechanisms. We also introduce a transferability zoning framework based on statistical uncertainty calibration, and extend CLS to encoder--head architectures to accommodate modern deep learning pipelines.

Overall, CLS provides a principled and computationally efficient framework for assessing dataset similarity. Beyond transfer learning, it may also support dataset selection, multi-source weighting, domain adaptation diagnostics, and pretraining data evaluation.

\paragraph{Code Availability.} Our code and data will be released publicly upon acceptance.

\bibliography{bibliography}

@ARTICLE{article,
    author = {{Author}, First. and {Author}, Second. and {Author}, Third},
    title = {Random article about some crap},
    journal = {Random Journal},
    year = 2002,
    month = jan, 
    volume = 666,
    pages = {1-20},
}

@inproceedings{do2024improving,
  title={Improving Accented Speech Recognition using Data Augmentation based on Unsupervised Text-to-Speech Synthesis},
  author={Do, Cong-Thanh and Imai, Shuhei and Doddipatla, Rama and Hain, Thomas},
  booktitle={2024 32nd European Signal Processing Conference (EUSIPCO)},
  pages={136--140},
  year={2024},
  organization={IEEE}
}

@article{reps2022learning,
  title={Learning patient-level prediction models across multiple healthcare databases: evaluation of ensembles for increasing model transportability},
  author={Reps, Jenna Marie and Williams, Ross D and Schuemie, Martijn J and Ryan, Patrick B and Rijnbeek, Peter R},
  journal={BMC medical informatics and decision making},
  volume={22},
  number={1},
  pages={142},
  year={2022},
  publisher={Springer}
}

@article{alvarez2020geometric,
  title={Geometric dataset distances via optimal transport},
  author={Alvarez-Melis, David and Fusi, Nicolo},
  journal={Advances in Neural Information Processing Systems},
  volume={33},
  pages={21428--21439},
  year={2020}
}

@article{gretton2012kernel,
  title={A kernel two-sample test},
  author={Gretton, Arthur and Borgwardt, Karsten M and Rasch, Malte J and Sch{\"o}lkopf, Bernhard and Smola, Alexander},
  journal={The Journal of Machine Learning Research},
  volume={13},
  number={1},
  pages={723--773},
  year={2012},
  publisher={JMLR. org}
}

@inproceedings{zhao2022comparing,
  title={Comparing distributions by measuring differences that affect decision making},
  author={Zhao, Shengjia and Sinha, Abhishek and He, Yutong and Perreault, Aidan and Song, Jiaming and Ermon, Stefano},
  booktitle={International Conference on Learning Representations},
  year={2022}
}

@article{gu2024robust,
  title={Robust angle-based transfer learning in high dimensions},
  author={Gu, Tian and Han, Yi and Duan, Rui},
  journal={Journal of the Royal Statistical Society Series B: Statistical Methodology},
  pages={qkae111},
  year={2024},
  publisher={Oxford University Press UK}
}

@book{wang2023introduction,
  title={Introduction to transfer learning: algorithms and practice},
  author={Wang, Jindong and Chen, Yiqiang},
  year={2023},
  publisher={Springer Nature}
}

@inproceedings{al2011adaptive,
  title={Adaptive boosting for transfer learning using dynamic updates},
  author={Al-Stouhi, Samir and Reddy, Chandan K},
  booktitle={Joint European Conference on Machine Learning and Knowledge Discovery in Databases},
  pages={60--75},
  year={2011},
  organization={Springer}
}

@inproceedings{dai2007boosting,
  title={Boosting for transfer learning International Conference on Machine Learning},
  author={Dai, Wenyuan and Yang, Qiang and Xue, Gui-Rong and Yu, Yong},
  booktitle={International Conference on Machine Learning},
  pages={193--200},
  year={2007}
}

@article{wang2023art,
  title={The art of transfer learning: An adaptive and robust pipeline},
  author={Wang, Boxiang and Wu, Yunan and Ye, Chenglong},
  journal={Stat},
  volume={12},
  number={1},
  pages={e582},
  year={2023},
  publisher={Wiley Online Library}
}

@article{halpern2015critical,
  title={Critical care medicine beds, use, occupancy, and costs in the United States: a methodological review},
  author={Halpern, Neil A and Pastores, Stephen M},
  journal={Critical care medicine},
  volume={43},
  number={11},
  pages={2452--2459},
  year={2015},
  publisher={LWW}
}

@article{stacy2011progressive,
  title={Progressive care units: different but the same},
  author={Stacy, Kathleen M},
  journal={Critical Care Nurse},
  volume={31},
  number={3},
  pages={77--83},
  year={2011},
  publisher={American Association of Critical Care Nurses}
}

@article{pollard2018eicu,
  title={The eICU Collaborative Research Database, a freely available multi-center database for critical care research},
  author={Pollard, Tom J and Johnson, Alistair EW and Raffa, Jesse D and Celi, Leo A and Mark, Roger G and Badawi, Omar},
  journal={Scientific data},
  volume={5},
  number={1},
  pages={1--13},
  year={2018},
  publisher={Nature Publishing Group}
}

@inproceedings{wang2019characterizing,
  title={Characterizing and avoiding negative transfer},
  author={Wang, Zirui and Dai, Zihang and P{\'o}czos, Barnab{\'a}s and Carbonell, Jaime},
  booktitle={Proceedings of the IEEE/CVF conference on computer vision and pattern recognition},
  pages={11293--11302},
  year={2019}
}

@article{yosinski2014transferable,
  title={How transferable are features in deep neural networks?},
  author={Yosinski, Jason and Clune, Jeff and Bengio, Yoshua and Lipson, Hod},
  journal={Advances in neural information processing systems},
  volume={27},
  year={2014}
}

@article{yang2024deep,
  title={Deep learning approaches for similarity computation: A survey},
  author={Yang, Peilun and Wang, Hanchen and Yang, Jianye and Qian, Zhengping and Zhang, Ying and Lin, Xuemin},
  journal={IEEE Transactions on Knowledge and Data Engineering},
  volume={36},
  number={12},
  pages={7893--7912},
  year={2024},
  publisher={IEEE}
}

@inproceedings{pardoe2010boosting,
  title={Boosting for regression transfer},
  author={Pardoe, David and Stone, Peter},
  booktitle={Proceedings of the 27th International Conference on International Conference on Machine Learning},
  pages={863--870},
  year={2010}
}

@inproceedings{zhong2023learning,
  title={Learning fair classifiers via min-max f-divergence regularization},
  author={Zhong, Meiyu and Tandon, Ravi},
  booktitle={2023 59th Annual Allerton Conference on Communication, Control, and Computing (Allerton)},
  pages={1--8},
  year={2023},
  organization={IEEE}
}

@inproceedings{zhong2024intrinsic,
  title={Intrinsic fairness-accuracy tradeoffs under equalized odds},
  author={Zhong, Meiyu and Tandon, Ravi},
  booktitle={2024 IEEE International Symposium on Information Theory (ISIT)},
  pages={220--225},
  year={2024},
  organization={IEEE}
}

@article{zhong2025splitz,
  title={Splitz: Certifiable robustness via split lipschitz randomized smoothing},
  author={Zhong, Meiyu and Tandon, Ravi},
  journal={IEEE Transactions on Information Forensics and Security},
  year={2025},
  publisher={IEEE}
}

@inproceedings{zhong2024learning,
  title={Learning Fair Robustness via Domain Mixup},
  author={Zhong, Meiyu and Tandon, Ravi},
  booktitle={2024 58th Asilomar Conference on Signals, Systems, and Computers},
  pages={196--202},
  year={2024},
  organization={IEEE}
}

@inproceedings{zhang2024filtered,
  title={Filtered randomized smoothing: A new defense for robust modulation classification},
  author={Zhang, Wenhan and Zhong, Meiyu and Tandon, Ravi and Krunz, Marwan},
  booktitle={MILCOM 2024-2024 IEEE Military Communications Conference (MILCOM)},
  pages={789--794},
  year={2024},
  organization={IEEE}
}

@article{polo2023unified,
  title={A unified framework for dataset shift diagnostics},
  author={Polo, Felipe Maia and Izbicki, Rafael and Lacerda Jr, Evanildo Gomes and Ibieta-Jimenez, Juan Pablo and Vicente, Renato},
  journal={Information Sciences},
  volume={649},
  pages={119612},
  year={2023},
  publisher={Elsevier}
}

@article{snell2017prototypical,
  title={Prototypical networks for few-shot learning},
  author={Snell, Jake and Swersky, Kevin and Zemel, Richard},
  journal={Advances in neural information processing systems},
  volume={30},
  year={2017}
}

@inproceedings{finn2017model,
  title={Model-agnostic meta-learning for fast adaptation of deep networks},
  author={Finn, Chelsea and Abbeel, Pieter and Levine, Sergey},
  booktitle={International conference on machine learning},
  pages={1126--1135},
  year={2017},
  organization={PMLR}
}

@inproceedings{standley2020tasks,
  title={Which tasks should be learned together in multi-task learning?},
  author={Standley, Trevor and Zamir, Amir and Chen, Dawn and Guibas, Leonidas and Malik, Jitendra and Savarese, Silvio},
  booktitle={International conference on machine learning},
  pages={9120--9132},
  year={2020},
  organization={PMLR}
}

@article{fifty2021efficiently,
  title={Efficiently identifying task groupings for multi-task learning},
  author={Fifty, Chris and Amid, Ehsan and Zhao, Zhe and Yu, Tianhe and Anil, Rohan and Finn, Chelsea},
  journal={Advances in Neural Information Processing Systems},
  volume={34},
  pages={27503--27516},
  year={2021}
}

@article{tan2023pfedsim,
  title={pfedsim: Similarity-aware model aggregation towards personalized federated learning},
  author={Tan, Jiahao and Zhou, Yipeng and Liu, Gang and Wang, Jessie Hui and Yu, Shui},
  journal={arXiv preprint arXiv:2305.15706},
  year={2023}
}

@inproceedings{koh2021wilds,
  title={Wilds: A benchmark of in-the-wild distribution shifts},
  author={Koh, Pang Wei and Sagawa, Shiori and Marklund, Henrik and Xie, Sang Michael and Zhang, Marvin and Balsubramani, Akshay and Hu, Weihua and Yasunaga, Michihiro and Phillips, Richard Lanas and Gao, Irena and others},
  booktitle={International conference on machine learning},
  pages={5637--5664},
  year={2021},
  organization={PMLR}
}

@inproceedings{yao2022improving,
  title={Improving out-of-distribution robustness via selective augmentation},
  author={Yao, Huaxiu and Wang, Yu and Li, Sai and Zhang, Linjun and Liang, Weixin and Zou, James and Finn, Chelsea},
  booktitle={International Conference on Machine Learning},
  pages={25407--25437},
  year={2022},
  organization={PMLR}
}

@inproceedings{cui2018large,
  title={Large scale fine-grained categorization and domain-specific transfer learning},
  author={Cui, Yin and Song, Yang and Sun, Chen and Howard, Andrew and Belongie, Serge},
  booktitle={Proceedings of the IEEE conference on computer vision and pattern recognition},
  pages={4109--4118},
  year={2018}
}

@article{stolte2024methods,
  title={Methods for quantifying dataset similarity: a review, taxonomy and comparison},
  author={Stolte, Marieke and Kappenberg, Franziska and Rahnenf{\"u}hrer, J{\"o}rg and Bommert, Andrea},
  journal={Statistic Surveys},
  volume={18},
  pages={163--298},
  year={2024},
  publisher={The American Statistical Association, the Bernoulli Society, the Institute~…}
}

@inproceedings{qamar2009online,
  title={Online and batch learning of generalized cosine similarities},
  author={Qamar, Ali Mustafa and Gaussier, Eric},
  booktitle={2009 Ninth IEEE International Conference on Data Mining},
  pages={926--931},
  year={2009},
  organization={IEEE}
}

@article{bayle2020cross,
  title={Cross-validation confidence intervals for test error},
  author={Bayle, Pierre and Bayle, Alexandre and Janson, Lucas and Mackey, Lester},
  journal={Advances in Neural Information Processing Systems},
  volume={33},
  pages={16339--16350},
  year={2020}
}

\appendix

\section{Proof of the theorems and lemma}

This section contains the proofs of the results presented in Section 2.

\subsection{Proof of Theorem 1}

\begin{proof}

For the target learning problem, the Bayes rule is given by:
\[
f^{*(t)}(\mathbf{X}) =
\begin{cases} 
1, & \text{if } (\beta^{(t)\top} \mathbf{X}) \geq 0, \\
0, & \text{if } (\beta^{(t)\top} \mathbf{X}) < 0.
\end{cases}
\]

The classification error when applying the target classifier to the source data, denoted as \( e^{(t\to s)} \), represents the first half of the CLS in the first part:
\begin{align*}
e^{(t\to s)}
&=\mathbb{E}_{(\mathbf{X}^{(s)},Y^{(s)})\sim\mathcal{P}^{(s)}}\!\left[\ell(f^{*(t)}(\mathbf{X}^{(s)}),Y^{(s)})\right]\\
&=\mathbb{E}\left[\mathbb{I}(f^{*(t)}(\mathbf{X}^{(s)}) \neq Y^{(s)})\right]\\
&= \mathbb{P}(f^{*(t)}(\mathbf{X}^{(s)}) \neq Y^{(s)}) \\
&= \mathbb{P}(f^{*(t)}(\mathbf{X}^{(s)}) = 1, Y^{(s)} = 0) + \mathbb{P}(f^{*(t)}(\mathbf{X}^{(s)}) = 0, Y^{(s)} = 1) \\
&= \mathbb{P}[\beta^{(t)\top} \mathbf{X}^{(s)} \geq 0, \beta^{(s)\top} \mathbf{X}^{(s)}+\xi^{(s)} < 0] + \mathbb{P}[\beta^{(t)\top} \mathbf{X}^{(s)} < 0, \beta^{(s)\top} \mathbf{X}^{(s)}+\xi^{(s)} \geq 0].
\end{align*}

Since \( \mathbf{X}^{(s)} \) and \(\xi^{(s)}\) follows an i.i.d. normal distribution:
\[
\mathbf{X}^{(s)} \sim \mathcal{N}_p(0, \Sigma^{(s)}), \quad \xi^{(s)} \sim \mathcal{N}_p(0, \sigma_s^2)
\]
the linear projections \( \beta^{(s)\top} \mathbf{X}^{(s)} \) and \( \beta^{(t)\top} \mathbf{X}^{(s)} +\xi^{(s)} \) are also normally distributed:
\[
\beta^{(t)\top} \mathbf{X}^{(s)}\sim \mathcal{N}(0, \beta^{(t)\top} \Sigma^{(s)} \beta^{(t)}), \quad \beta^{(s)\top} \mathbf{X}^{(s)}+\xi^{(s)} \sim \mathcal{N}(0, \beta^{(s)\top} \Sigma^{(s)} \beta^{(s)}+\sigma_s^2).
\]

The covariance between these projections is given by:
\[
\text{cov}(\beta^{(t)\top} \mathbf{X}^{(s)}, \beta^{(s)\top} \mathbf{X}^{(s)}+\xi^{(s)}) = \text{cov}(\beta^{(t)\top} \mathbf{X}^{(s)}, \beta^{(s)\top} \mathbf{X}^{(s)}) = \beta^{(t)\top} \Sigma^{(s)} \beta^{(s)}.
\]

Thus, the joint covariance matrix is:
\[
\Sigma_{\text{joint}} = 
\begin{pmatrix}
    \beta^{(t)\top} \Sigma^{(s)} \beta^{(t)} & \beta^{(t)\top} \Sigma^{(s)} \beta^{(s)} \\
    \beta^{(t)\top} \Sigma^{(s)} \beta^{(s)} & \beta^{(s)\top} \Sigma^{(s)} \beta^{(s)} + \sigma_s^2
\end{pmatrix}.
\]

To obtain standard normal variables, we define:
\[
Z_t = \frac{\beta^{(t)\top} \mathbf{X}^{(s)}}{\sqrt{\beta^{(t)\top} \Sigma^{(s)} \beta^{(t)}}} \sim \mathcal{N}(0,1), \quad
Z_s = \frac{\beta^{(s)\top} \mathbf{X}^{(s)}+\xi^{(s)}}{\sqrt{\beta^{(s)\top} \Sigma^{(s)} \beta^{(s)} + \sigma_s^2}} \sim \mathcal{N}(0,1).
\]

The joint distribution of \( (Z_t, Z_s) \) is a bivariate normal with zero mean and covariance matrix:
\[
\Sigma_Z = 
\begin{pmatrix}
    1 & \rho_1 \\
    \rho_1 & 1
\end{pmatrix}.
\]
where 
\[
\rho_1 = \frac{\beta^{(t)\top} \Sigma^{(s)} \beta^{(s)}}{\sqrt{\beta^{(t)\top} \Sigma^{(s)} \beta^{(t)}} \sqrt{\beta^{(s)\top} \Sigma^{(s)} \beta^{(s)} + \sigma_s^2}}=\jw{\cos_{\Sigma^{(s)}}^{(0,\sigma_s^2)}
\big(\beta^{(t)},\beta^{(s)}\big)}.
\]
Its density is given by:
\[
f(z_t, z_s) = \frac{1}{2\pi \sqrt{1 - \rho_1^2}} \exp\left( -\frac{z_t^2 - 2 \rho_1 z_t z_s + z_s^2}{2(1 - \rho_1^2)} \right).
\]

The classification error corresponds to the probability that \( Z_t \geq 0 \) and \( Z_s < 0 \) (or vice versa):
\[
\mathbb{P}(Z_t \geq 0, Z_s < 0) = \int_{0}^{\infty} \int_{-\infty}^{0} f(z_t, z_s) \, dz_s \, dz_t.
\]

A known result from bivariate normal probability computations states:
\[
\mathbb{P}(Z_t \geq 0, Z_s < 0) = \frac{\arccos(\rho_1)}{2\pi}.
\]

Since \( \mathbb{P}(Z_t < 0, Z_s \geq 0) \) is symmetric, we obtain:
\[
e^{(t \to s)} = \mathbb{P}(Z_t \geq 0, Z_s < 0) + \mathbb{P}(Z_t < 0, Z_s \geq 0) = \frac{\arccos(\rho_1)}{\pi}.
\]

Thus, the classification error when applying the target classifier to the source data is:
\[
e^{(t \to s)} = \frac{\arccos(\rho_1)}{\pi}.
\]

Similarly, the classification error when applying the source classifier to the target data is:
\[
e^{(s \to t)} = \frac{\arccos(\rho_2)}{\pi}.
\]
where 
\[
\rho_2 = \frac{\beta^{(t)\top} \Sigma^{(t)} \beta^{(s)}}{\sqrt{\beta^{(t)\top} \Sigma^{(t)} \beta^{(t)} + \sigma_t^2} \sqrt{\beta^{(s)\top} \Sigma^{(t)} \beta^{(s)}}}=\jw{\cos_{\Sigma^{(t)}}^{(\sigma_t^2,0)}
\big(\beta^{(t)},\beta^{(s)}\big)}.
\]

And the target-domain error and the source-domain error are:
\[
e^{(t)}_0 = \frac{\arccos(\rho_3)}{\pi} , \quad e^{(s)}_0 = \frac{\arccos(\rho_4)}{\pi}
\]
where 
\[
\rho_3 = \sqrt{\frac{\beta^{(t)\top} \Sigma^{(t)} \beta^{(t)}}{\beta^{(t)\top} \Sigma^{(t)} \beta^{(t)} + \sigma_t^2 }}  =\jw{\cos_{\Sigma^{(t)}}^{(0,\sigma_t^2)}
\big(\beta^{(t)},\beta^{(t)}\big)}
\]
and
\[\rho_4 = \sqrt{\frac{\beta^{(s)\top} \Sigma^{(s)} \beta^{(s)}}{\beta^{(s)\top} \Sigma^{(s)} \beta^{(s)} + \sigma_s^2 }} =\jw{\cos_{\Sigma^{(s)}}^{(0,\sigma_s^2)}
\big(\beta^{(s)},\beta^{(s)}\big)}
\]

Then 
\[
\mathrm{CLS} = \frac{e^{(t \to s)} + e^{(s \to t)}}{2} - \frac{e^{(t)}_0 + e^{(s)}_0}{2} = \frac{\arccos(\rho_1)+\arccos(\rho_2)}{2\pi} - \frac{\arccos(\rho_3)+\arccos(\rho_4)}{2\pi}.
\]
or 
\[
\jw{\mathrm{CLS} =
\frac{\arccos(\rho_1)-\arccos(\rho_4)}{2\pi} + \frac{\arccos(\rho_2)-\arccos(\rho_3)}{2\pi}
=
\frac{\theta_{t\to s}-\theta_{s\to s}}{2\pi}
+
\frac{\theta_{s\to t}-\theta_{t\to t}}{2\pi},}
\]

\end{proof}

\subsection{Proof of Lemma 1}

\begin{proof}
We recall that
\[
\mathrm{CLS}
=
\frac{\arccos(\rho_1)-\arccos(\rho_4)}{2\pi}
+
\frac{\arccos(\rho_2)-\arccos(\rho_3)}{2\pi},
\]
where
\[
\rho_1
=
\frac{\beta^{(t)\top} \Sigma^{(s)} \beta^{(s)}}
{\sqrt{\beta^{(t)\top} \Sigma^{(s)} \beta^{(t)}}\sqrt{\beta^{(s)\top} \Sigma^{(s)} \beta^{(s)}+\sigma_s^2}},
\qquad
\rho_4
=
\sqrt{
\frac{\beta^{(s)\top} \Sigma^{(s)} \beta^{(s)}}
{\beta^{(s)\top} \Sigma^{(s)} \beta^{(s)}+\sigma_s^2}
},
\]
and
\[
\rho_2
=
\frac{\beta^{(t)\top} \Sigma^{(t)} \beta^{(s)}}
{\sqrt{\beta^{(t)\top} \Sigma^{(t)} \beta^{(t)}+\sigma_t^2}\sqrt{\beta^{(s)\top} \Sigma^{(t)} \beta^{(s)}}},
\qquad
\rho_3
=
\sqrt{
\frac{\beta^{(t)\top} \Sigma^{(t)} \beta^{(t)}}
{\beta^{(t)\top} \Sigma^{(t)} \beta^{(t)}+\sigma_t^2}
}.
\]

For notational convenience, define
\[
a^{(s)} := \beta^{(s)\top}\Sigma^{(s)}\beta^{(s)},
\qquad
c^{(s)} := \frac{\beta^{(t)\top}\Sigma^{(s)}\beta^{(s)}}{\sqrt{\beta^{(t)\top}\Sigma^{(s)}\beta^{(t)}}},
\]
and
\[
a^{(t)} := \beta^{(t)\top}\Sigma^{(t)}\beta^{(t)},
\qquad
c^{(t)} := \frac{\beta^{(t)\top}\Sigma^{(t)}\beta^{(s)}}{\sqrt{\beta^{(s)\top}\Sigma^{(t)}\beta^{(s)}}}.
\]

Then
\[
\rho_1 = \frac{c^{(s)}}{\sqrt{a^{(s)}+\sigma_s^2}},
\qquad
\rho_4 = \sqrt{\frac{a^{(s)}}{a^{(s)}+\sigma_s^2}},
\]
and
\[
\rho_2 = \frac{c^{(t)}}{\sqrt{a^{(t)}+\sigma_t^2}},
\qquad
\rho_3 = \sqrt{\frac{a^{(t)}}{a^{(t)}+\sigma_t^2}}.
\]

By the Cauchy--Schwarz inequality under the semi-inner product induced by
\(\Sigma^{(s)}\),
\[
\left|
\beta^{(t)\top}\Sigma^{(s)}\beta^{(s)}
\right|^2
\le
\bigl(\beta^{(t)\top}\Sigma^{(s)}\beta^{(t)}\bigr)
\bigl(\beta^{(s)\top}\Sigma^{(s)}\beta^{(s)}\bigr),
\]
we have
\[
(c^{(s)})^2 \le a^{(s)}.
\]
Hence \(\rho_1 \le \rho_4\), and therefore
\[
\arccos(\rho_1)-\arccos(\rho_4)
=
\arccos\!\left(
\rho_1\rho_4+\sqrt{1-\rho_1^2}\sqrt{1-\rho_4^2}
\right).
\]

Now,
\[
\rho_1\rho_4
=
\frac{c^{(s)}}{\sqrt{a^{(s)}+\sigma_s^2}}
\sqrt{\frac{a^{(s)}}{a^{(s)}+\sigma_s^2}}
=
\frac{c^{(s)}\sqrt{a^{(s)}}}{a^{(s)}+\sigma_s^2},
\]
and
\[
1-\rho_1^2
=
1-\frac{(c^{(s)})^2}{a^{(s)}+\sigma_s^2}
=
\frac{a^{(s)}+\sigma_s^2-(c^{(s)})^2}{a^{(s)}+\sigma_s^2},
\]
so
\[
\sqrt{1-\rho_1^2}
=
\frac{\sqrt{a^{(s)}+\sigma_s^2-(c^{(s)})^2}}{\sqrt{a^{(s)}+\sigma_s^2}}.
\]
Also,
\[
1-\rho_4^2
=
1-\frac{a^{(s)}}{a^{(s)}+\sigma_s^2}
=
\frac{\sigma_s^2}{a^{(s)}+\sigma_s^2},
\]
and thus
\[
\sqrt{1-\rho_4^2}
=
\frac{|\sigma_s|}{\sqrt{a^{(s)}+\sigma_s^2}}.
\]
Therefore,
\[
\arccos(\rho_1)-\arccos(\rho_4)
=
\arccos\!\left(
\frac{
c^{(s)}\sqrt{a^{(s)}}
+
|\sigma_s|\sqrt{a^{(s)}+\sigma_s^2-(c^{(s)})^2}
}{
a^{(s)}+\sigma_s^2
}
\right).
\]
Since \(\sigma_s \ge 0\), this becomes
\[
\arccos(\rho_1)-\arccos(\rho_4)
=
\arccos\!\left(
\frac{
c^{(s)}\sqrt{a^{(s)}}
+
\sigma_s\sqrt{a^{(s)}+\sigma_s^2-(c^{(s)})^2}
}{
a^{(s)}+\sigma_s^2
}
\right).
\]

Assume first that
\[
(c^{(s)})^2 < a^{(s)}.
\]
Let
\[
\Delta^{(s)} := a^{(s)}-(c^{(s)})^2 > 0.
\]
Using Taylor expansions around $\sigma_s=0$,
\[
\sqrt{\Delta^{(s)}+\sigma_s^2}
=
\sqrt{\Delta^{(s)}}+\frac{\sigma_s^2}{2\sqrt{\Delta^{(s)}}}+O(\sigma_s^4),
\qquad
\frac{1}{a^{(s)}+\sigma_s^2}
=
\frac{1}{a^{(s)}}\left(1-\frac{\sigma_s^2}{a^{(s)}}+O(\sigma_s^4)\right),
\]
we obtain
\[
\frac{
c^{(s)}\sqrt{a^{(s)}}
+
\sigma_s\sqrt{a^{(s)}+\sigma_s^2-(c^{(s)})^2}
}{
a^{(s)}+\sigma_s^2
}
=
\frac{c^{(s)}}{\sqrt{a^{(s)}}}
+
\frac{\sqrt{\Delta^{(s)}}}{a^{(s)}}\sigma_s
-
\frac{c^{(s)}}{(a^{(s)})^{3/2}}\sigma_s^2
+
O(\sigma_s^3).
\]
Since
\[
1-\left(\frac{c^{(s)}}{\sqrt{a^{(s)}}}\right)^2
=
\frac{\Delta^{(s)}}{a^{(s)}},
\]
a Taylor expansion of \(\arccos(\cdot)\) around $\sigma_s=0$ gives
\[
\arccos(\rho_1)-\arccos(\rho_4)
=
\arccos\!\left(\frac{c^{(s)}}{\sqrt{a^{(s)}}}\right)
-\frac{\sigma_s}{\sqrt{a^{(s)}}}
+
\frac{c^{(s)}}{2a^{(s)}\sqrt{a^{(s)}-(c^{(s)})^2}}\sigma_s^2
+
O(\sigma_s^3).
\]

In the boundary case
\[
(c^{(s)})^2=a^{(s)},
\]
equality holds in the Cauchy--Schwarz inequality. In particular, if
\[
c^{(s)}=\sqrt{a^{(s)}},
\]
then
\[
\arccos(\rho_1)-\arccos(\rho_4)=0
\qquad\text{for all }\sigma_s\ge 0.
\]

Similarly, by applying the same argument to the pair \((\rho_2,\rho_3)\), we obtain
\[
\arccos(\rho_2)-\arccos(\rho_3)
=
\arccos\!\left(
\frac{
c^{(t)}\sqrt{a^{(t)}}
+
\sigma_t\sqrt{a^{(t)}+\sigma_t^2-(c^{(t)})^2}
}{
a^{(t)}+\sigma_t^2
}
\right),
\]
and, for \(\sigma_t\ge 0\) and \((c^{(t)})^2<a^{(t)}\),
\[
\arccos(\rho_2)-\arccos(\rho_3)
=
\arccos\!\left(\frac{c^{(t)}}{\sqrt{a^{(t)}}}\right)
-\frac{\sigma_t}{\sqrt{a^{(t)}}}
+
\frac{c^{(t)}}{2a^{(t)}\sqrt{a^{(t)}-(c^{(t)})^2}}\sigma_t^2
+
O(\sigma_t^3).
\]
Moreover, if \(c^{(t)}=\sqrt{a^{(t)}}\), then
\[
\arccos(\rho_2)-\arccos(\rho_3)=0
\qquad\text{for all }\sigma_t\ge 0.
\]

Combining the above two expansions yields
\[
\mathrm{CLS}
=
\frac{1}{2\pi}
\left[
\arccos\!\left(\frac{c^{(s)}}{\sqrt{a^{(s)}}}\right)
+
\arccos\!\left(\frac{c^{(t)}}{\sqrt{a^{(t)}}}\right)
\right]
-\frac{1}{2\pi}
\left[
\frac{\sigma_s}{\sqrt{a^{(s)}}}
+
\frac{\sigma_t}{\sqrt{a^{(t)}}}
\right]
\]
\[
\qquad
+
\frac{1}{4\pi}
\left[
\frac{c^{(s)}}{a^{(s)}\sqrt{a^{(s)}-(c^{(s)})^2}}\sigma_s^2
+
\frac{c^{(t)}}{a^{(t)}\sqrt{a^{(t)}-(c^{(t)})^2}}\sigma_t^2
\right]
+
O(\sigma_s^3+\sigma_t^3),
\]
as \(\sigma_s\to 0\) and \(\sigma_t\to 0\), provided that
\[
(c^{(s)})^2<a^{(s)},
\qquad
(c^{(t)})^2<a^{(t)}.
\]

Under the special case \(\Sigma^{(s)}=\Sigma^{(t)}=I\), we have
\[
a^{(s)}=\|\beta^{(s)}\|^2,
\qquad
a^{(t)}=\|\beta^{(t)}\|^2,
\]
and
\[
c^{(s)}=\frac{\beta^{(t)\top}\beta^{(s)}}{\|\beta^{(t)}\|},
\qquad
c^{(t)}=\frac{\beta^{(t)\top}\beta^{(s)}}{\|\beta^{(s)}\|}.
\]
Define
\[
\theta_\beta
:=
\arccos\!\left(
\frac{\beta^{(t)\top}\beta^{(s)}}{\|\beta^{(t)}\|\,\|\beta^{(s)}\|}
\right).
\]
Then
\[
\frac{c^{(s)}}{\sqrt{a^{(s)}}}
=
\frac{c^{(t)}}{\sqrt{a^{(t)}}}
=
\cos\theta_\beta.
\]
Moreover,
\[
c^{(s)}=\|\beta^{(s)}\|\cos\theta_\beta,
\qquad
c^{(t)}=\|\beta^{(t)}\|\cos\theta_\beta,
\]
so that
\[
\sqrt{a^{(s)}-(c^{(s)})^2}
=
\|\beta^{(s)}\|\sin\theta_\beta,
\qquad
\sqrt{a^{(t)}-(c^{(t)})^2}
=
\|\beta^{(t)}\|\sin\theta_\beta.
\]
Substituting these identities into the general expansion yields
\[
\mathrm{CLS}
=
\frac{\theta_\beta}{\pi}
-\frac{1}{2\pi}
\left[
\frac{\sigma_s}{\|\beta^{(s)}\|}
+
\frac{\sigma_t}{\|\beta^{(t)}\|}
\right]
+
\frac{\cot\theta_\beta}{4\pi}
\left[
\frac{\sigma_s^2}{\|\beta^{(s)}\|^2}
+
\frac{\sigma_t^2}{\|\beta^{(t)}\|^2}
\right]
+
O(\sigma_s^3+\sigma_t^3),
\]
provided that \(0<\theta_\beta<\pi\).

The boundary cases correspond to collinearity of \(\beta^{(s)}\) and \(\beta^{(t)}\).

If \(\theta_\beta=0\), equivalently \(\beta^{(t)}=\lambda\beta^{(s)}\) for some \(\lambda>0\), then
\[
\mathrm{CLS}=0.
\]

If \(\theta_\beta=\pi\), equivalently \(\beta^{(t)}=\lambda\beta^{(s)}\) for some \(\lambda<0\), then
\[
\mathrm{CLS}
=
1-\frac{\arccos(\rho_3)+\arccos(\rho_4)}{\pi},
\]
and in particular,
\[
\mathrm{CLS}\to 1
\qquad\text{as }\sigma_s\to 0,\ \sigma_t\to 0.
\]

\end{proof}

\subsection{Proof of Theorem 2}

\begin{proof}

Consider a simple Linear Discriminant Analysis (LDA) setting. For $k =0,1$, we assume that the target data follows
\[
\mathbf{X}^{(t)} \mid Y^{(t)} = k \sim \mathcal{N}_p((-1)^{k+1}\mu^{(t)}, I), 
\]
and the source data follows
\[
\mathbf{X}^{(s)} \mid Y^{(s)} = k \sim \mathcal{N}_p((-1)^{k+1}\mu^{(s)}, I), 
\]
where $\mu^{(t)},\mu^{(s)} \in \mathbb{R}^p$. The class priors are balanced, i.e., 
\[
P(Y^{(t)} = k)=\frac{1}{2} , 
\qquad 
P(Y^{(s)} = k)=\frac{1}{2}
\]

The probability density functions (PDFs) for the target data are
\[
p(\mathbf{X}^{(t)} \mid Y^{(t)} = k) 
= \frac{1}{(2\pi)^{p/2} }
\exp \left( -\frac{1}{2} \|\mathbf{X}^{(t)} - (-1)^{k+1}\mu^{(t)}\|^2 \right),
\quad k\in\{0,1\}.
\]

With a balanced prior, the Bayesian classification rule assigns $\mathbf{X}^{(t)}$ to class $Y^{(t)} = 1$ if
\[
 p(\mathbf{X}^{(t)} \mid Y^{(t)} = 1) > p(\mathbf{X}^{(t)} \mid Y^{(t)} = 0).
\]
Taking logarithms and substituting the Gaussian PDFs, we obtain
\[
-\frac{1}{2} \|\mathbf{X}^{(t)} - \mu^{(t)}\|^2  > -\frac{1}{2} \|\mathbf{X}^{(t)} + \mu^{(t)}\|^2 
\]
Using the identity for the difference of squares, we obtain, 
\[
\mu^{(t)\top}\mathbf{x} > 0.
\]
Since the decision boundary has probability zero under the Gaussian model, we may equivalently define the Bayes classifier as
\[
f^{*(t)}(\mathbf{x}) =
\begin{cases}
1, & \text{if } \mu^{(t)\top}\mathbf{x} \ge 0,\\
0, & \text{if } \mu^{(t)\top}\mathbf{x} < 0.
\end{cases}
\]

\medskip
The classification error when applying the target classifier to the source data, denoted as $ e^{(t \to s)} $, represents the first half of the first part in the CLS:
\begin{align*}
e^{(t \to s)}
&=\mathbb{E}_{(\mathbf{X}^{(s)},Y^{(s)})\sim\mathcal{P}^{(s)}}\!\left[\ell(f^{*(t)}(\mathbf{X}^{(s)}),Y^{(s)})\right]\\
&=\mathbb{E}\left[\mathbb{I}\big(f^{*(t)}(\mathbf{X}^{(s)}) \neq Y^{(s)}\big)\right]\\
&= \mathbb{P}\big(f^{*(t)}(\mathbf{X}^{(s)}) \neq Y^{(s)}\big) \\
&= \mathbb{P}\big(f^{*(t)}(\mathbf{X}^{(s)}) = 1, Y^{(s)} = 0\big) 
  + \mathbb{P}\big(f^{*(t)}(\mathbf{X}^{(s)}) = 0, Y^{(s)} = 1\big) \\
&= \mathbb{P}\big(f^{*(t)}(\mathbf{X}^{(s)}) = 1 \mid Y^{(s)} = 0\big)\, \mathbb{P}(Y^{(s)} = 0) 
  + \mathbb{P}\big(f^{*(t)}(\mathbf{X}^{(s)}) = 0 \mid Y^{(s)} = 1\big)\, \mathbb{P}(Y^{(s)} = 1).
\end{align*}
Let
\[
Z^{(s)} = \mu^{(t)\top} \mathbf{X}^{(s)} .
\]
By the properties of linear transformations of Gaussian random variables,
\[
Z^{(s)} \mid Y^{(s)} = k 
\sim \mathcal{N}\big((-1)^{k+1}\mu^{(t)^\top} \mu^{(s)},\ \|\mu^{(t)}\|^2\big),
\quad k\in\{0,1\}.
\]

Then
\begin{align*}
\mathbb{P}\big(f^{*(t)}(\mathbf{X}^{(s)}) = 1 \mid Y^{(s)} = 0\big)
&= \mathbb{P}\big(Z^{(s)} \ge 0 \mid Y^{(s)} = 0\big)
= 1 - \Phi\!\left(\frac{ \mu^{(t)^\top} \mu^{(s)}}{\|\mu^{(t)}\|}\right),\\
\mathbb{P}\big(f^{*(t)}(\mathbf{X}^{(s)}) = 0 \mid Y^{(s)} = 1\big)
&= \mathbb{P}\big(Z^{(s)} < 0 \mid Y^{(s)} = 1\big)
= \Phi\!\left(\frac{-\mu^{(t)^\top} \mu^{(s)}}{\|\mu^{(t)}\|}\right),
\end{align*}
where $\Phi(\cdot)$ is the standard normal CDF. Hence,
\begin{align*}
e^{(t \to s)} 
&= \frac12
\left[
1 - \Phi\!\left(\frac{\mu^{(t)^\top} \mu^{(s)}}{\|\mu^{(t)}\|}\right)
\right]
+ \frac12
\Phi\!\left(\frac{-\mu^{(t)^\top} \mu^{(s)}}{\|\mu^{(t)}\|}\right)\\
&= \Phi\!\left(-\frac{\mu^{(t)^\top} \mu^{(s)}}{\|\mu^{(t)}\|}\right) =  \Phi\!\Big(-\|\mu^{(s)}\|\cos\theta_{\mu}\Big)
\end{align*}

where
\[
\cos\theta_{\mu} 
= \frac{\mu^{(t)\top} \mu^{(s)}}{\|\mu^{(t)}\|\,\|\mu^{(s)}\|}.
\]

\medskip
Similarly,  for the source learning problem
\[
e^{(s\to t)} =\Phi\!\Big(-\|\mu^{(t)}\|\cos\theta_{\mu}\Big)
\]

The target-domain error and the source-domain error are
\begin{align*}
e^{(t)}_0
&= \mathbb{P}\big(f^{*(t)}(\mathbf{X}^{(t)}) \neq Y^{(t)}\big) \\
&= \mathbb{P}\big(f^{*(t)}(\mathbf{X}^{(t)}) = 1 \mid Y^{(t)} = 0\big)\, \mathbb{P}(Y^{(t)} = 0) 
  + \mathbb{P}\big(f^{*(t)}(\mathbf{X}^{(t)}) = 0 \mid Y^{(t)} = 1\big)\, \mathbb{P}(Y^{(t)} = 1) \\
&= \frac12
\left[
1 - \Phi(\|\mu^{(t)}\|)
\right]
+ \frac12
\Phi(-\|\mu^{(t)}\|)=  \Phi(-\|\mu^{(t)}\|)
\end{align*}
and similarly
\[
e^{(s)}_0
= \mathbb{P}\big(f^{*(s)}(\mathbf{X}^{(s)}) \neq Y^{(s)}\big)
= \Phi\!(-\|\mu^{(s)}\|).
\]

Therefore, the Cross-Learning Score reduces to the simple closed form
\begin{align*}
\mathrm{CLS}
&= \frac{e^{(t \to s)} + e^{(s \to t)}}{2} - \frac{e^{(t)}_0 + e^{(s)}_0}{2} \\
&= \frac{1}{2}\Big[
\Phi\!\big(-\|\mu^{(s)}\|\cos\theta_{\mu}\big)
+ \Phi\!\big(-\|\mu^{(t)}\|\cos\theta_{\mu}\big)
\Big] - \frac{1}{2}\Big[
\Phi\!\big(-\|\mu^{(s)}\|\big)
+ \Phi\!\big(-\|\mu^{(t)}\|\big)
\Big] \\
&= \frac{1}{2}\Big[
\Phi\!\big(-\|\mu^{(s)}\|\cos\theta_{\mu}\big) -\Phi\!\big(-\|\mu^{(s)}\|\big)\Big] 
+ \frac{1}{2}\Big[\Phi\!\big(-\|\mu^{(t)}\|\cos\theta_{\mu}\big)-
 \Phi\!\big(-\|\mu^{(t)}\|\big)
\Big]. 
\end{align*}

\end{proof}

\section{More Details of Numerical Experienments}
In this section, we present the detailed setup of the synthetic experiments designed to evaluate the effectiveness of CLS in quantifying dataset similarity for supervised learning.

Our code and data will be released publicly upon acceptance.
Our code will be available at \href{https://github.com/ShudongSun/CLS}{https://github.com/ShudongSun/CLS}.


Let $\nu^{(t)}$ denote the parameter vector for the target data and $\nu^{(s)}$ denote that for the source data. Given $\nu^{(t)}$, Section~\ref{subsec:generate} introduces a method to generate $\nu^{(s)}$ with controlled cosine similarity to $\nu^{(t)}$. 

We design seven distinct experimental settings to characterize the similarity between target and source data, using the generation method described in Section~\ref{subsec:generate}. The details and results for these settings are presented in Sections~\ref{subsec:bc}--\ref{subsec:mc}. For all seven experiments, we set $n_t = n_s = 200$, conduct 50 independent replicates, and report the averaged results. To assess prediction accuracy, an additional 5{,}000 data points are independently generated as the test set.

Each setting above contains three figures and two tables, which are described in the following four paragraphs.


The first figure (Figure~\ref{fig:na_logit},~\ref{fig:na_probit},~\ref{fig:na_LDA},~\ref{fig:na_Mix_Gau},~\ref{fig:na_linr},~\ref{fig:na_nlinr},~\ref{fig:na_multi}) in each setting illustrates the relationship between the estimated similarity metrics and the underlying data similarity, which is controlled either by the cosine similarity between the model parameters of the target and source distributions or by the mixing coefficient \(\alpha \in [0,1]\). In each plot, the x-axis represents data similarity, while the y-axis shows various similarity metrics, including three variants of the CLS: the oracle CLS (red), the weighted-average estimate \(\widehat{\text{CLS}}\) (light green), and the ensemble estimate \(\widehat{\text{CLS}}_E\) (light blue), as well as benchmark measures such as KL Divergence (brown), Wasserstein Distance (dark green), and Optimal Transport Dataset Distance (OTDD) (dark blue).

The second figure (Figure~\ref{fig:trans_logit},~\ref{fig:trans_probit},~\ref{fig:trans_LDA},~\ref{fig:trans_Mix_Gau},~\ref{fig:trans_linr},~\ref{fig:trans_nlinr},~\ref{fig:trans_multi}) in each setting presents the evaluation results for the proposed transferable zones. The top panel shows the average number of transfer methods outperforming the baseline, with background colors indicating the predicted
zone (PT, AZ, or NT). The bottom panel shows the difference between the test error of the baseline and the naive
transfer method.

\begin{figure}[!ht]
    \centering
    \includegraphics[width=\textwidth]{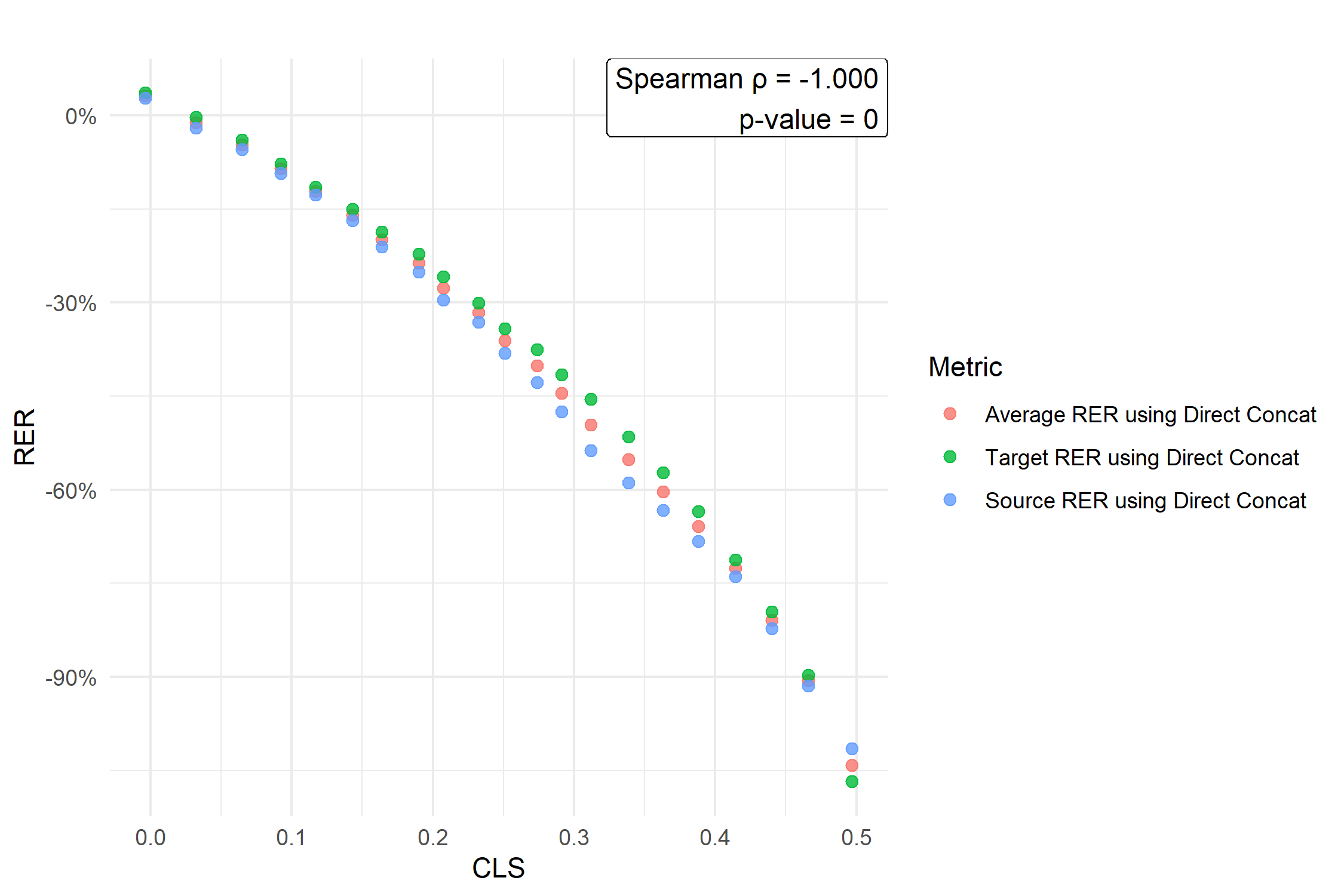}
    \caption{Relative Error Reduction(RER) v.s. CLS under the probit model. The Spearman rank correlation coefficient and the p-value of corresponding test is also shown.}
    \label{fig:CLS_probit}
\end{figure}

\begin{figure}[!ht]
    \centering
    \includegraphics[width=\textwidth]{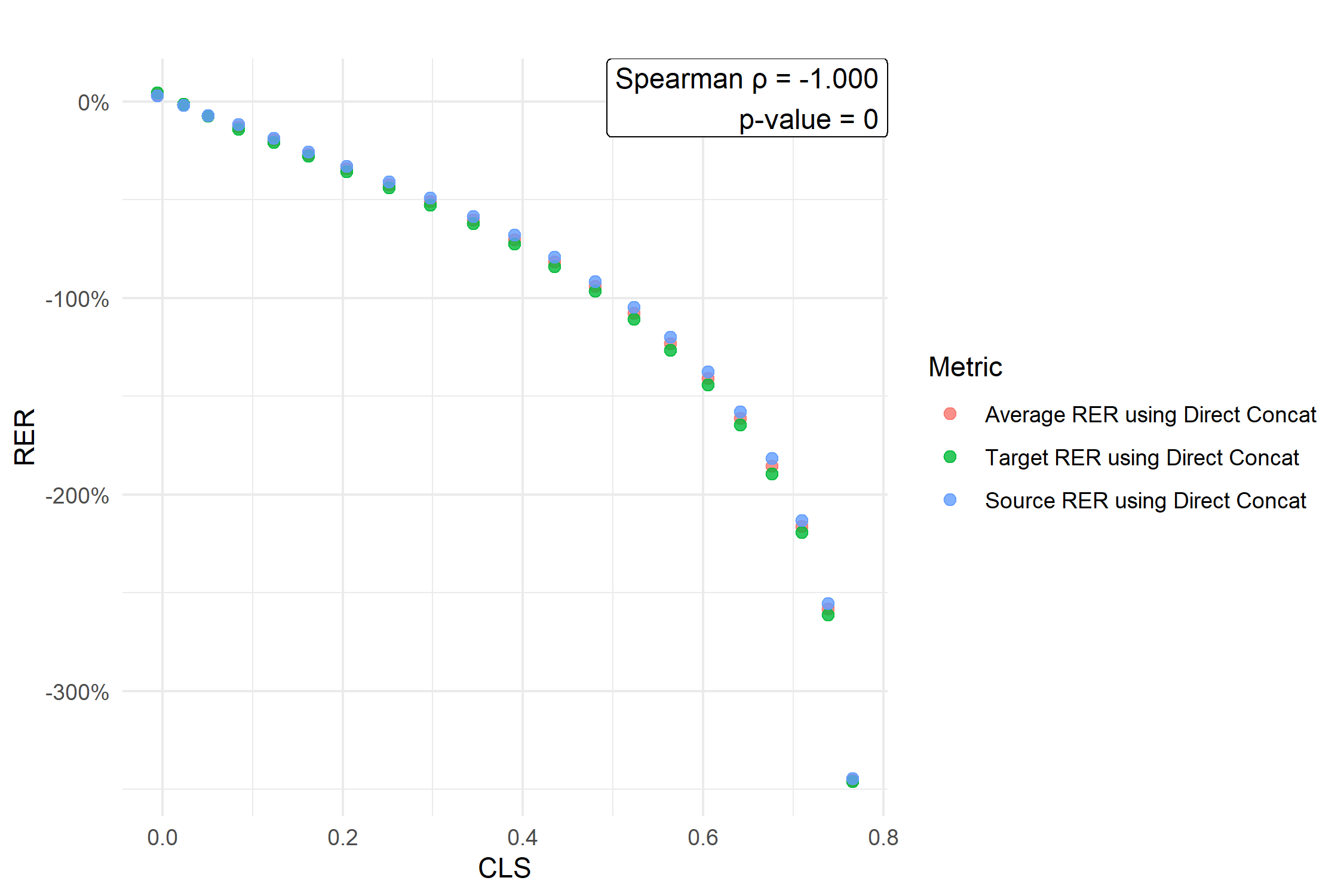}
    \caption{Relative Error Reduction(RER) v.s. CLS under the LDA model. The Spearman rank correlation coefficient and the p-value of corresponding test is also shown.}
    \label{fig:CLS_lda}
\end{figure}

\begin{figure}[!ht]
    \centering
    \includegraphics[width=\textwidth]{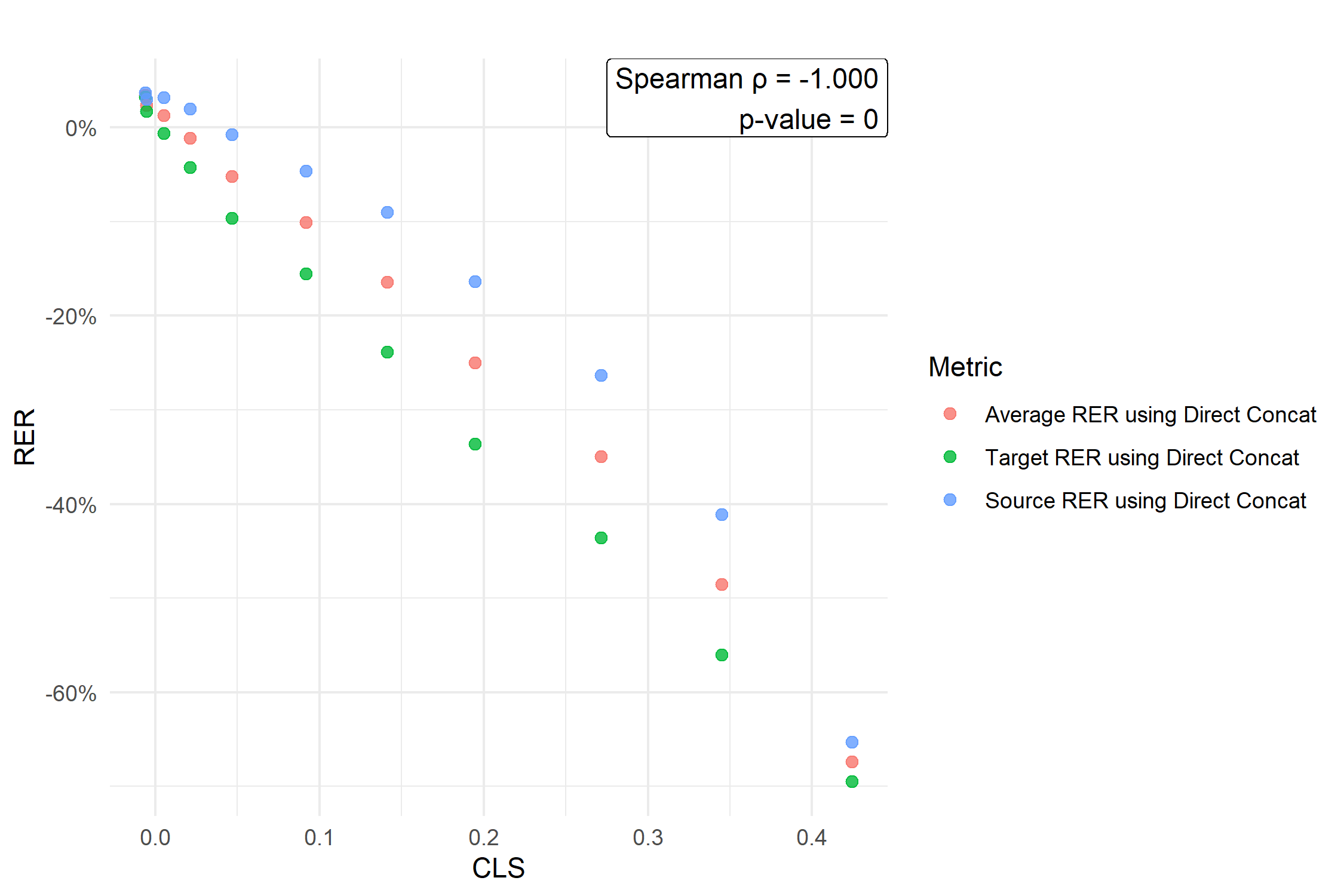}
    \caption{Relative Error Reduction(RER) v.s. CLS under the Mixture Gaussian model. The Spearman rank correlation coefficient and the p-value of corresponding test is also shown.}
    \label{fig:CLS_mix_gau}
\end{figure}

\begin{figure}[!ht]
    \centering
    \includegraphics[width=\textwidth]{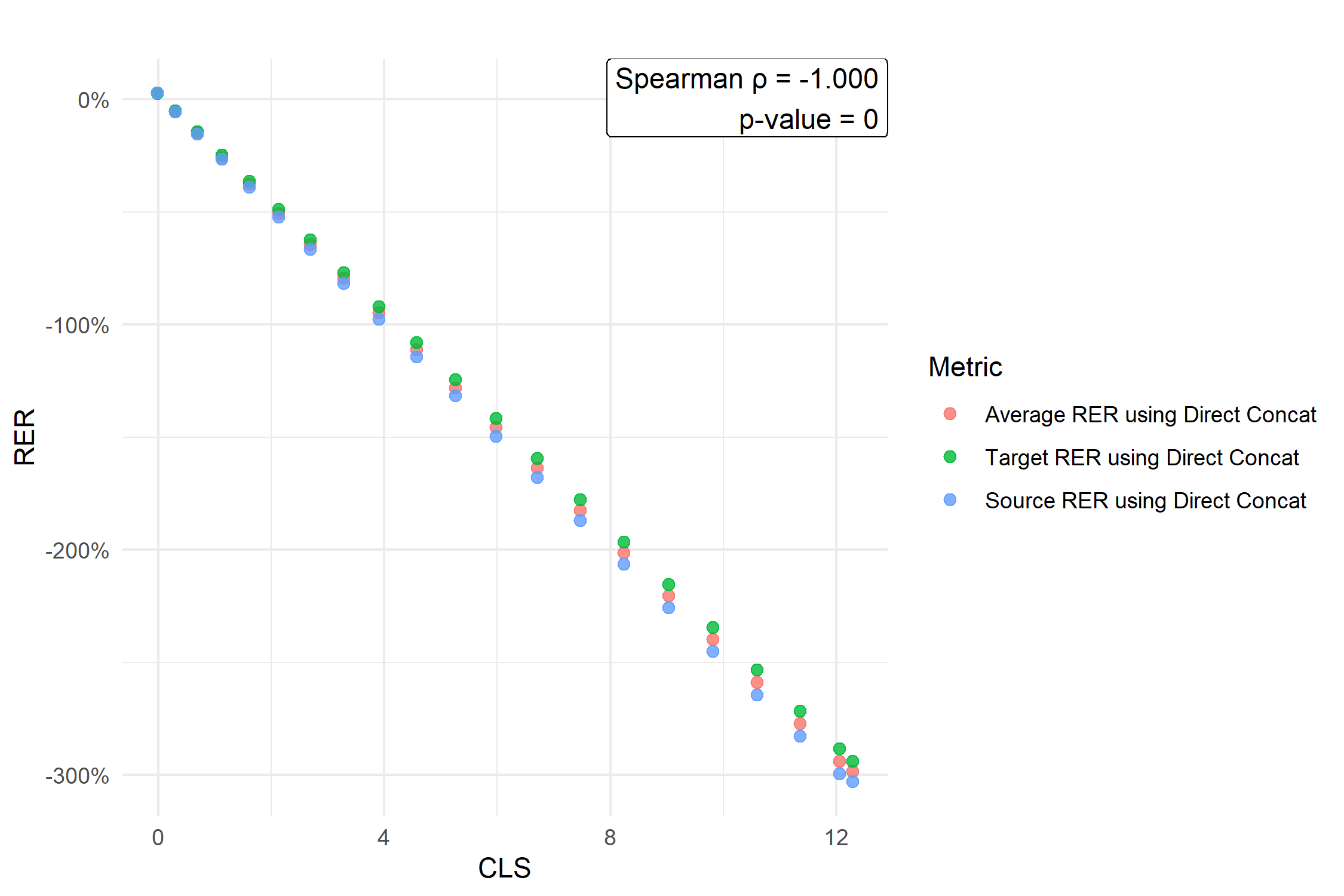}
    \caption{Relative Error Reduction(RER) v.s. CLS under the Linear Regression model. The Spearman rank correlation coefficient and the p-value of corresponding test is also shown.}
    \label{fig:CLS_linr}
\end{figure}

\begin{figure}[!ht]
    \centering
    \includegraphics[width=\textwidth]{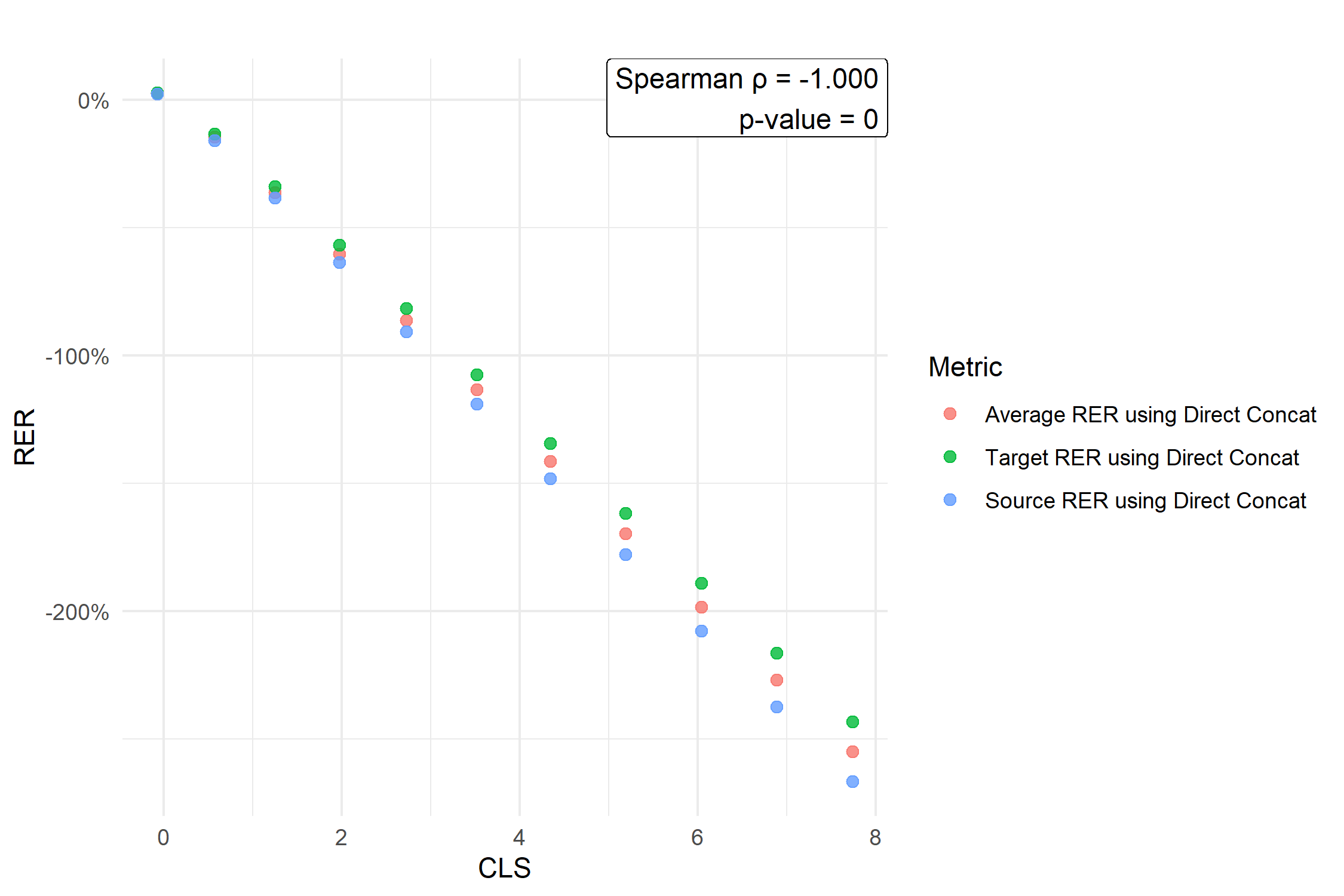}
    \caption{Relative Error Reduction(RER) v.s. CLS under the non-linear regression model. The Spearman rank correlation coefficient and the p-value of corresponding test is also shown.}
    \label{fig:CLS_nlinr}
\end{figure}

\begin{figure}[!ht]
    \centering
    \includegraphics[width=\textwidth]{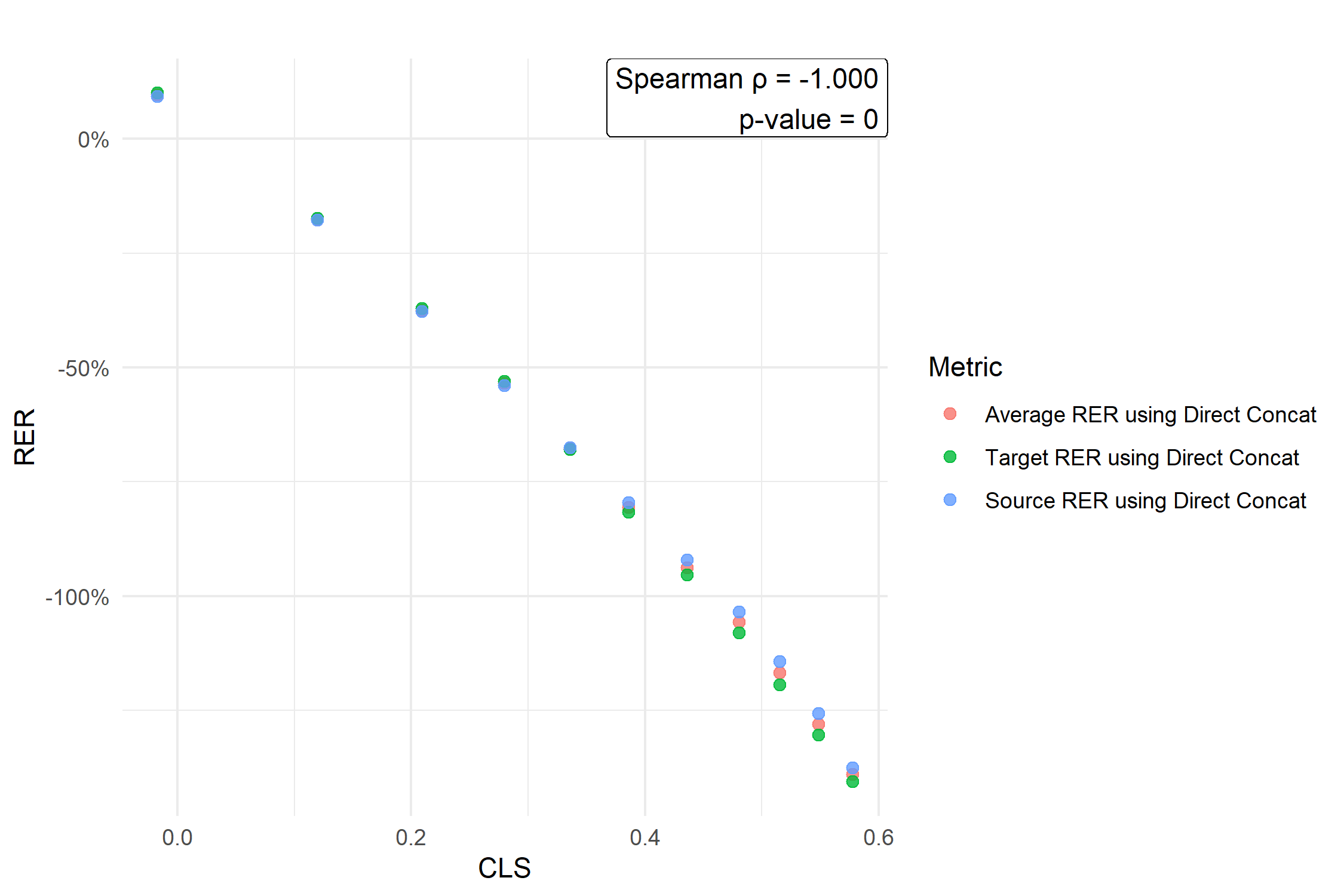}
    \caption{Relative Error Reduction(RER) v.s. CLS under the 4-class classification model. The Spearman rank correlation coefficient and the p-value of corresponding test is also shown.}
    \label{fig:CLS_multi_class}
\end{figure}

The third figure (Figure~\ref{fig:CLS_logit},~\ref{fig:CLS_probit},~\ref{fig:CLS_lda},~\ref{fig:CLS_mix_gau},~\ref{fig:CLS_linr},~\ref{fig:CLS_nlinr},~\ref{fig:CLS_multi_class}) in each setting presents the relationship between CLS and the three RER measures using direct concat transfer learning method. 
As CLS increases, all three RER values decrease monotonically. 
This indicates that CLS effectively captures dataset transferability: higher CLS corresponds to lower performance gain from transfer learning.

The first table (Table~\ref{table1.4.1},~\ref{table1.4.2},~\ref{table1.4.3},~\ref{table1.4.4},~\ref{table1.4.6},~\ref{table1.4.7},~\ref{table1.4.8}) in each setting presents comparisons of various similarity score metrics across different cosine similarity values or mixture parameters $\alpha$ in the mixture Gaussian setting. For each configuration, we additionally report the corresponding KL Divergence, Wasserstein Distance, and Optimal Transport Dataset Distance (OTDD) metrics (noting that OTDD cannot be directly computed for regression tasks). The notation $\left|\rho_{p}\right|$ denotes the absolute value of the Pearson correlation coefficient between each metric and the cosine similarity, while $\left|\rho_{s}\right|$ represents the absolute value of the Spearman rank correlation coefficient. Mathematically, the Pearson and Spearman correlation coefficients between KL Divergence or Wasserstein Distance and cosine similarity are undefined; for convenience, we set them to zero in our analysis.

The second table (Table~\ref{table1.4.1w},~\ref{table1.4.2w},~\ref{table1.4.3w},~\ref{table1.4.4w},~\ref{table1.4.6w},~\ref{table1.4.7w},~\ref{table1.4.8w}) in each setting present comparisons of various estimated CLS values across different cosine similarity levels or mixture parameters $\alpha$ in the mixture Gaussian setting. \textit{Diff.} denotes the mean absolute difference between each estimated score and the oracle CLS, as defined in Section~5.2. Lower values indicate smaller deviation from the Oracle CLS. The first column reports the optimal-model estimates assuming access to the Bayes rule of the target learning problem. The next four columns show single-model estimates obtained from the generalized linear model, linear SVM, radial SVM, and XGBoost. The following three columns present multi-model estimates based on unweighted average (\textit{Unw.Avg}), weighted average (\textit{Wtd.Avg}), and ensemble (\textit{Ens.}) schemes using the above four single models. The final column lists the Oracle CLS as the reference gold standard.

In the \textit{Comparison of CLS Estimation} section, we additionally include three nonlinear experimental settings. 
For these settings, it is difficult to systematically control the similarity level when generating different source datasets. 
Therefore, instead of varying similarity levels, we fix one pair of target and source data-generating parameters and repeat the experiment 50 times to obtain stable results.

These experiments provide an important complementary evaluation of CLS estimation under nonlinear data-generating mechanisms. 
Detailed descriptions of these settings are provided in subsections~\ref{subsubsec:qda} to~\ref{subsubsec:radial}.

\subsection{Generating Source Parameter Vectors with Controlled Cosine Similarity}
\label{subsec:generate}

We begin by considering the parameter vectors \(\nu^{(t)}\) corresponding to the target data. 
The source parameter vectors \(\nu^{(s)}\) are then generated using the procedure described below to control the cosine similarity between \(\nu^{(t)}\) and \(\nu^{(s)}\).

To illustrate the process of generating \(\nu^{(s)}\), we begin with a three-dimensional space (\(p = 3\)) by transforming the Cartesian coordinates \(\nu^{(t)} = (x_t, y_t, z_t)\) into spherical coordinates \((r, \theta_t, \phi)\) using the following equations:

\[
r = \sqrt{x_t^2 + y_t^2 + z_t^2}
\]
\[
\theta_t = \arccos\left(\frac{z_t}{r}\right)
\]
\[
\phi = \arccos\left(\frac{x_t}{r \cdot \sin \theta_t}\right)
\]

In these equations, the radius \( r \) indicates the Euclidean distance from the origin to the point, the polar angle \( \theta_t \) quantifies the angle between the positive \( z \)-axis and the point, and the azimuthal angle \( \phi \) measures the angle between the positive \( x \)-axis and the projection of the point onto the \( xy \)-plane.

\begin{figure}[!htbp]
    \begin{center}
    \includegraphics[width=0.4\textwidth]{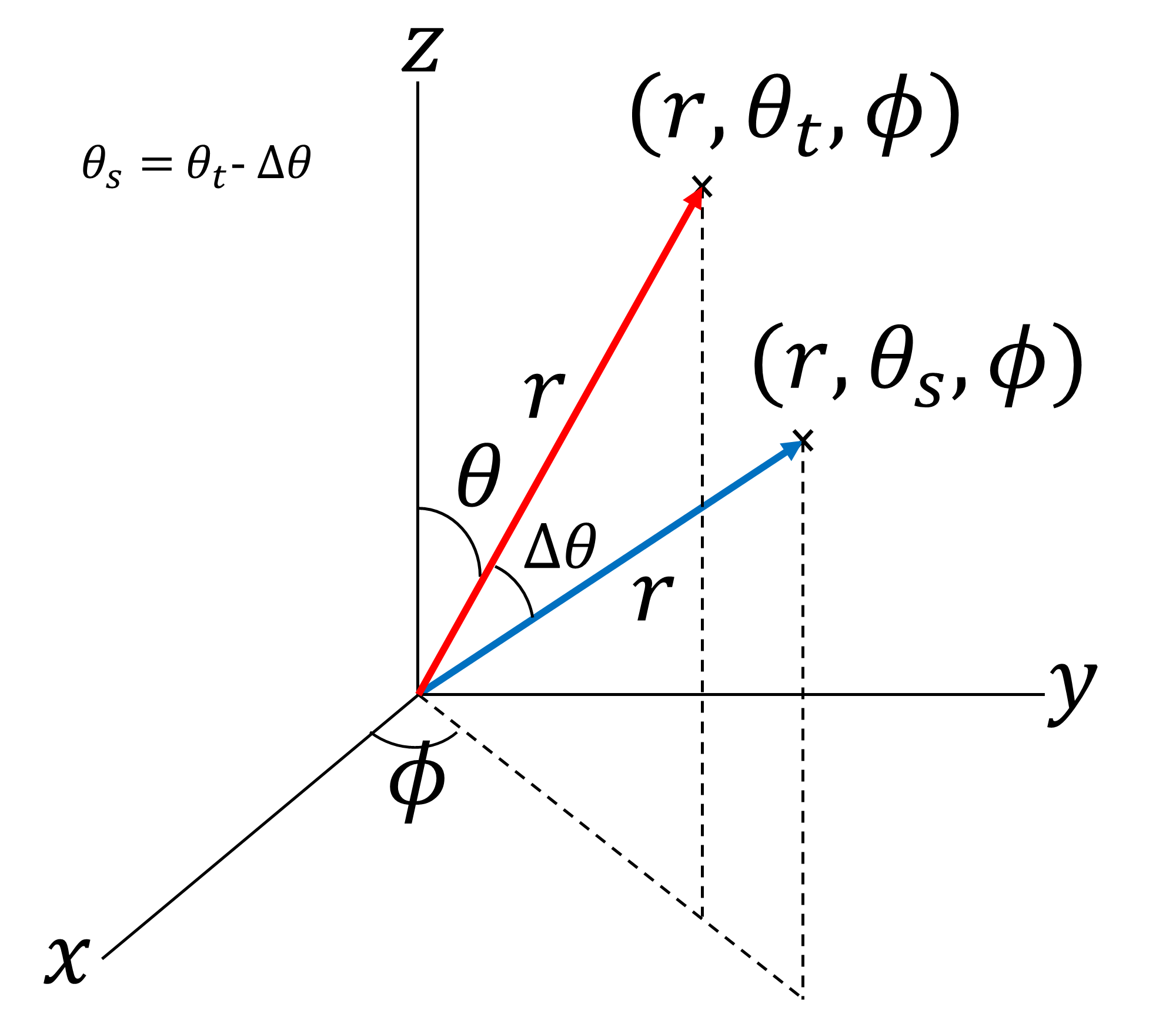}
    \caption[illustration 1]{illustration to generate the new vector $\nu^{(s)}$\label{figurechangetheta}}
    \end{center}
\end{figure}

To generate a new vector \(\nu^{(s)}\) for source data, we maintain a constant azimuthal angle \(\phi\) and modify the polar angle \( \theta_t \) to  \( \theta_s \). The detailed steps are as follows:

\begin{enumerate}
  \item \emph{Defining Cosine Similarity Values}:
    Begin by defining a set of cosine similarity values, denoted by \( t \), which are uniformly distributed between -1 and 1. These values correspond to the desired similarity relative to the original vector's orientation for target data.
  
  \item \emph{Determining Angular Shifts}:
    Using the cosine similarity values, calculate the corresponding changes in the polar angle, \(\Delta \theta\), by taking the inverse cosine (arccos) of the similarity values: \(\Delta \theta = \arccos(t)\). This calculation establishes the angular adjustments necessary to reach the specified similarities. 
  
  \item \emph{Computing New Polar Angles}:
    Derive the new polar angles, \(\theta_s\), by subtracting the angular shifts, \(\Delta \theta\), from the original polar angle, \(\theta_t\): \(\theta_s = \theta_t - \Delta \theta\). This operation results in a series of modified polar angles while retaining the original the radius \( r \) and azimuthal angle \(\phi\).
  
  \item \emph{Reconstructing Cartesian Coordinates}:
    Finally, utilize the updated polar angles  \(\theta_s\) and the unchanged \( r \), \(\phi\) to convert back to Cartesian coordinates, thereby defining the new vector \(\nu^{(s)}=(x_s,y_s,z_s)\). The updated coordinates are given by:
  
    \[
    x_s = r \sin \theta_s \cos \phi
    \]
    \[
    y_s = r \sin \theta_s \sin \phi
    \]
    \[
    z_s = r \cos \theta_s
    \]
\end{enumerate}

This systematic approach facilitates the controlled generation of a new vector \(\nu^{(s)}\) by varying the polar angle \(\theta\) while preserving the orientation in the \( xy \)-plane. This method is especially useful for applications that demand precise directional control.

Let us calculate the cosine similarity between $\nu^{(t)}$ and $\nu^{(s)}$, denoted as \( S_C(\nu^{(t)},\nu^{(s)}) \):
\begin{align*}
S_C(\nu^{(t)},\nu^{(s)}) &= \frac{\nu^{(t)} \cdot \nu^{(s)}}{\|\nu^{(t)}\| \|\nu^{(s)}\|} = \frac{x_t \cdot x_s + y_t \cdot y_s + z_t \cdot z_s}{\sqrt{x_t^2 + y_t^2 + z_t^2} \sqrt{x_s^2 + y_s^2 + z_s^2}} \\
&= \frac{r \cdot \sin \theta_t \cdot \cos \phi \cdot r \cdot \sin \theta_s \cdot \cos \phi + r \cdot \sin \theta_t \cdot \sin \phi \cdot r \cdot \sin \theta_s \cdot \sin \phi + r \cdot \cos \theta_t \cdot r \cdot \cos \theta_s}{r \cdot r} \\
&= \sin \theta_t \cdot \cos \phi \cdot \sin \theta_s \cdot \cos \phi + \sin \theta_t \cdot \sin \phi \cdot \sin \theta_s \cdot \sin \phi + \cos \theta_t \cdot \cos \theta_s\\
&= \sin \theta_t \cdot \sin \theta_s \left( \cos^2 \phi + \sin^2 \phi \right) + \cos \theta_t \cdot \cos \theta_s\\
&= \cos(\theta_t - \theta_s) \\
&= \cos(\Delta \theta)
\end{align*}

The cosine similarity between \( \nu^{(t)} \) and \( \nu^{(s)} \) is, in fact, given by \( \cos(\Delta \theta) \), where \( \Delta \theta \) represents the angular difference between the two vectors. By adjusting \( \Delta \theta \) to be uniformly distributed between -1 and 1, we can effectively control the quality of the source data, ranging from extremely poor to extremely good (i.e., having the same distribution as the target data).

For dimensions \( d \geq 3 \), we can employ a similar method to extend the concept to higher-dimensional spaces.

We define a coordinate system in an \( n \)-dimensional Euclidean space that generalizes the spherical coordinate system used in three-dimensional Euclidean space. In this system, the coordinates are composed of a radial coordinate \( r \) and \( n - 1 \) angular coordinates \( \varphi_1, \varphi_2, \dots, \varphi_{n-1} \). If \( x_i \) are the Cartesian coordinates, i.e. $\nu^{(t)}=(x_1^{(t)},\cdots,x_n^{(t)})$, then we can compute \( r, \varphi_1^{(t)}, \varphi_2, \dots, \varphi_{n-1} \) using the following transformations:

\[
r = \sqrt{(x_1^{(t)})^2 + (x_2^{(t)})^2 + \cdots + (x_{n-1}^{(t)})^2 + (x_{n}^{(t)})^2},
\]
\[
\varphi_1^{(t)} = \arccos\left( \frac{x_1^{(t)}}{r} \right),
\]
\[
\varphi_2 = \arccos\left( \frac{x_2^{(t)}}{r \sin(\varphi_1)} \right),
\]
\[
\vdots
\]
\[
\varphi_{n-2} = \arccos\left( \frac{x_{n-2}^{(t)}}{r \sin(\varphi_1) \cdots \sin(\varphi_{n-3})} \right),
\]
\[
\varphi_{n-1} = \arccos\left( \frac{x_{n-1}^{(t)}}{r \sin(\varphi_1) \cdots \sin(\varphi_{n-2})} \right).
\]

Following the same procedure as outlined above, we modify the angle \( \varphi_1 \) to \( \varphi_1^{(s)} \) by applying an angular shift \( \Delta \varphi_1 \), such that:

\[
\varphi_1^{(s)} = \varphi_1 - \Delta \varphi_1.
\]

and we keep the \( r, \varphi_2, \dots, \varphi_{n-1} \) unchanged.

The Cartesian coordinates can then be reconstructed using the modified angle \( \varphi_1^{(s))} \) for the source data:

\[
x_1^{(s)} = r \cos(\varphi_1^{(s)}),
\]
\[
x_2^{(s)} = r \sin(\varphi_1^{(s)}) \cos(\varphi_2),
\]
\[
x_3^{(s)} = r \sin(\varphi_1^{(s)}) \sin(\varphi_2) \cos(\varphi_3),
\]
\[
\vdots
\]
\[
x_{n-1}^{(s)} = r \sin(\varphi_1^{(s)}) \cdots \sin(\varphi_{n-2}) \cos(\varphi_{n-1}),
\]
\[
x_n^{(s)} = r \sin(\varphi_1^{(s)}) \cdots \sin(\varphi_{n-2}) \sin(\varphi_{n-1}).
\]
and
\[
\nu^{(s)}=(x_1^{(s)},x_2^{(s)},\cdots,x_{n-1}^{(s)},x_n^{(s)})
\]

Consequently, we arrive at a similar conclusion, where the cosine similarity is given by \( S_C(\nu^{(t)},\nu^{(s)}) = \cos(\Delta \varphi_1) \).

\subsection{Synthetic Data Generation for Binary Classification}
\label{subsec:bc}

In the binary classification experiments, we consider five generative settings:
(i) logistic regression;
(ii) probit model regression;
(iii) LDA;
(iv) QDA;
(v) mixture Gaussian model.

For CLS estimation, we consider four models for 
binary classification: logistic regression, linear SVM, radial SVM, XGBoost.

\subsubsection{Logistic Regression Setting}

We consider a commonly utilized linear classifier-logistic regression. Each sample $\mathbf{x}_i^{(t)}$ and $\mathbf{x}_i^{(s)}$ is independently drawn from $\mathcal{N}_p(0, \Sigma)$, 
where $i = 1, \dots, n_t$ for the target data and $i = 1, \dots, n_s$ for the source data. The covariance matrix \( \Sigma \) has an auto-regression correlation structure, specifically \( \Sigma = (0.5^{|r - c|})_{p \times p} \) for feature indices $r, c = 1, \dots, p$ and feature dimension $p=10$. For the target data, each binary outcome is generated according to a Bernoulli distribution, where the probability is given by:
\[
P(y_i = 1) = \frac{1}{1 + \exp(-\pi_i)},
\]
where $\pi_i = \beta^{(t)\top} \mathbf{x}_i$, and coefficient vector $\beta^{(t)}$ is drawn from 
$\mathcal{N}_p\!\left(\tfrac{1}{4}\mathbf{1}, \tfrac{1}{16}\mathbf{I}_p\right)$, 
where $\mathbf{1}$ denotes a $p$-dimensional vector of ones, 
and $\mathbf{I}_p$ denotes the $p \times p$ identity matrix. The binary responses in the source datasets are generated similarly, but with varying coefficients: we generate $\beta^{(s)}$ using the method described in Section~\ref{subsec:generate} in order to control the cosine similarity between $\beta^{(t)}$ and $\beta^{(s)}$. 

The Oracle CLS is calculated using \(\beta^{(t)},\beta^{(s)}\) based on the above data generation mechanism.

\begin{figure}[!htbp]
    \centering
    \includegraphics[width=0.8\textwidth]{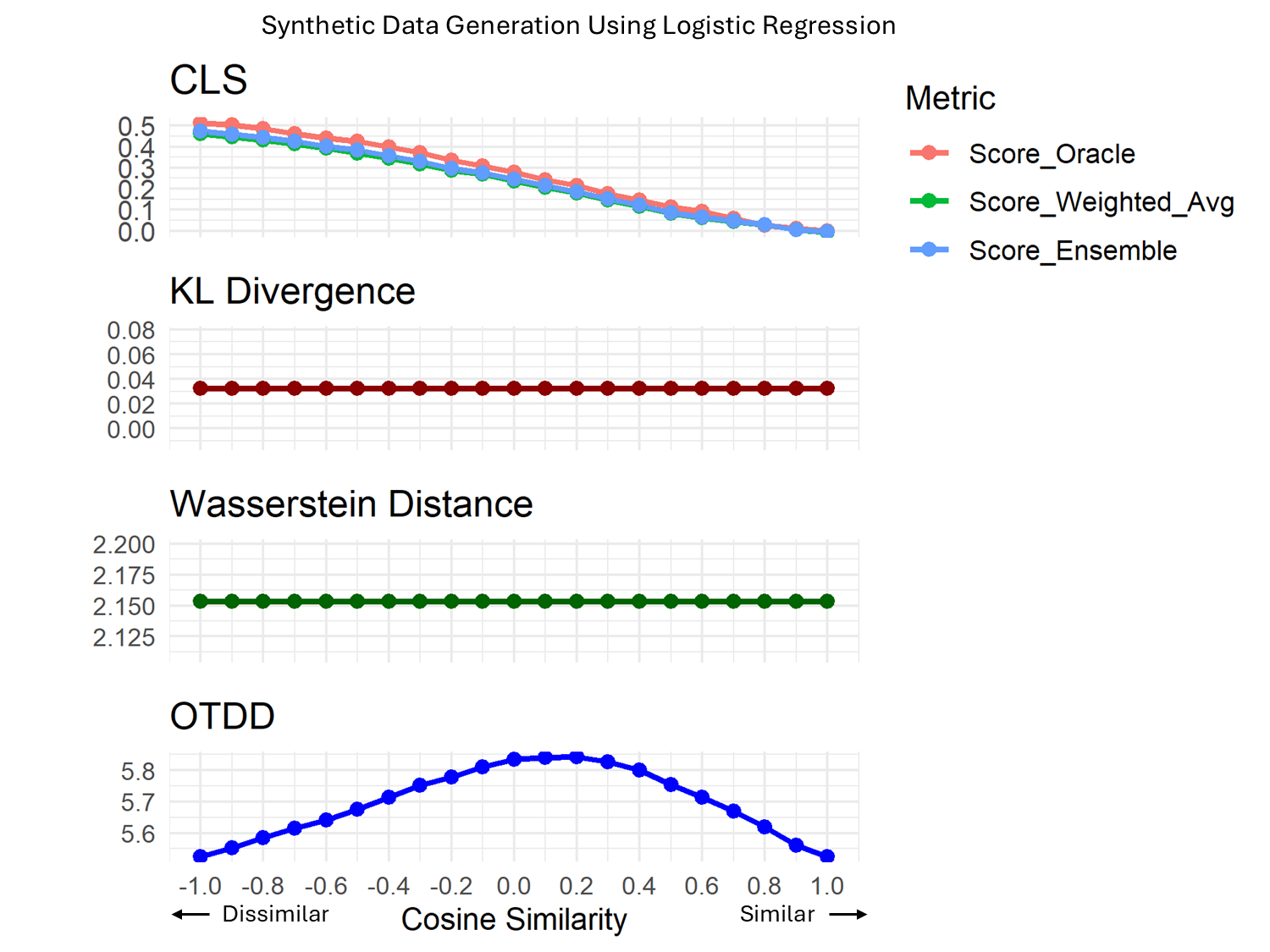}
    \caption{Comparison of CLS vs other similarity metrics
under the Logistic Regression Setting}
    \label{fig:na_logit}
\end{figure}

\begin{figure}[!htbp]
    \centering
    \includegraphics[width=0.7\textwidth]{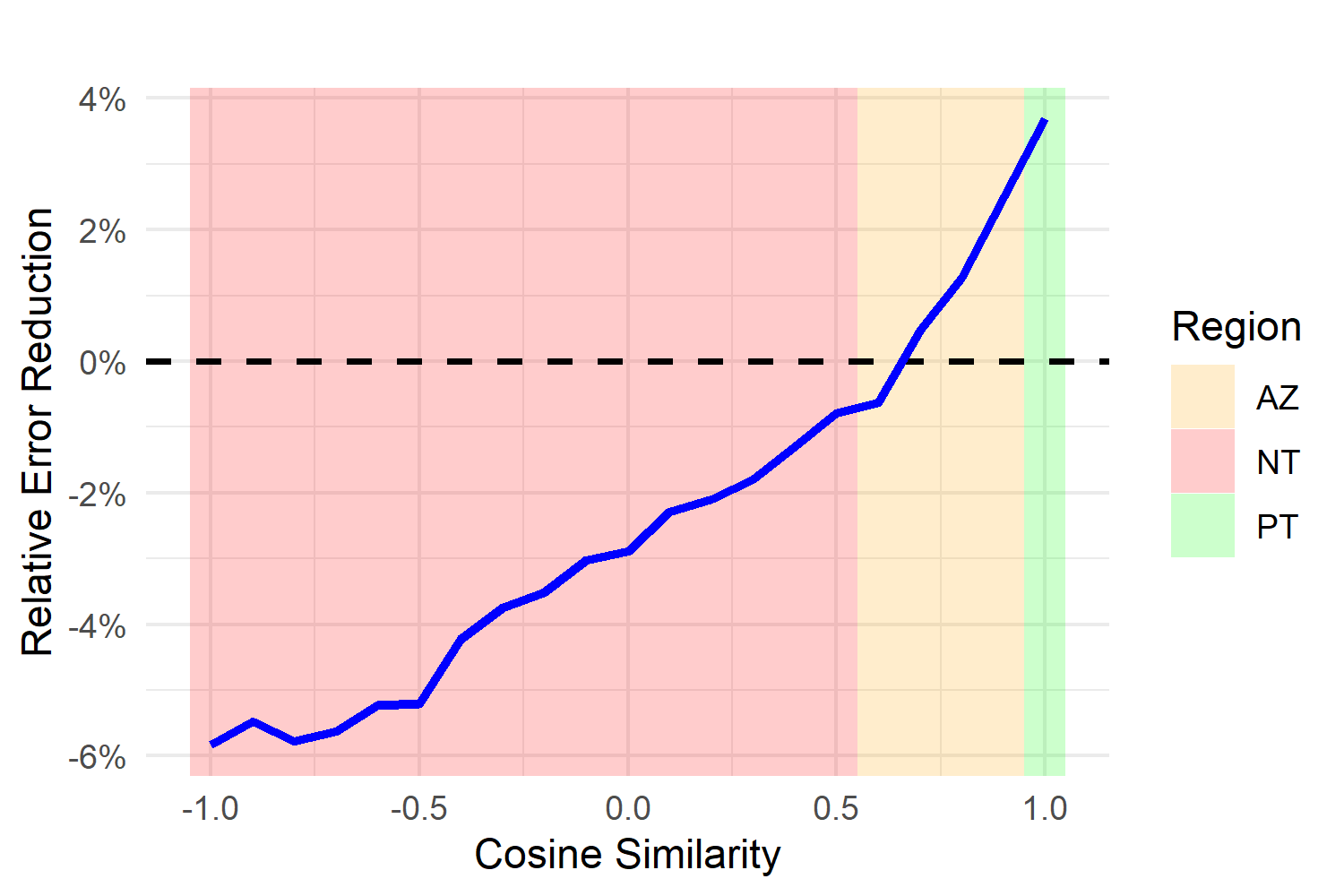}
    \caption{Transferability zones (positive, ambiguous,
and negative) identified under the logistic regression model}
    \label{fig:trans_logit}
\end{figure}

All results for this setting are presented in Figure~\ref{fig:na_logit}, Table~\ref{table1.4.1}, Table~\ref{table1.4.1w}, and Figure~\ref{fig:trans_logit}.
 
\begin{table}[htbp]
\centering
\scalebox{0.9}{ 
\setlength{\tabcolsep}{2pt} 
\renewcommand{\arraystretch}{0.9} 
\begin{tabularx}{\textwidth}{
>{\centering\arraybackslash}X||
>{\centering\arraybackslash}X 
>{\centering\arraybackslash}X 
>{\centering\arraybackslash}X 
>{\centering\arraybackslash}X 
>{\centering\arraybackslash}X 
>{\centering\arraybackslash}X 
>{\centering\arraybackslash}X }
\toprule
\makecell{\textbf{Cosine}\\\textbf{Sim.}} &
\makecell{\textbf{Score}\\\textbf{Oracle}} &
\makecell{\textbf{Score}\\\textbf{Unw.}\\\textbf{Avg}} &
\makecell{\textbf{Score}\\\textbf{Wtd.}\\\textbf{Avg}} &
\makecell{\textbf{Score}\\\textbf{Ens.}} &
\makecell{\textbf{KL}\\\textbf{Div.}} &
\makecell{\textbf{Wass.}\\\textbf{Dist.}} &
\textbf{OTDD} \\
\midrule
-1.0 & 0.5148 & 0.4559 & 0.4624 & 0.4756 & 0.0323 & 2.1536 & 5.5249 \\
-0.9 & 0.5044 & 0.4442 & 0.4486 & 0.4605 & 0.0323 & 2.1536 & 5.5521 \\
-0.8 & 0.4872 & 0.4284 & 0.4330 & 0.4452 & 0.0323 & 2.1536 & 5.5855 \\
-0.7 & 0.4618 & 0.4105 & 0.4155 & 0.4256 & 0.0323 & 2.1536 & 5.6148 \\
-0.6 & 0.4406 & 0.3872 & 0.3923 & 0.4027 & 0.0323 & 2.1536 & 5.6421 \\
-0.5 & 0.4271 & 0.3654 & 0.3698 & 0.3826 & 0.0323 & 2.1536 & 5.6754 \\
-0.4 & 0.3976 & 0.3415 & 0.3462 & 0.3582 & 0.0323 & 2.1536 & 5.7133 \\
-0.3 & 0.3704 & 0.3136 & 0.3179 & 0.3292 & 0.0323 & 2.1536 & 5.7515 \\
-0.2 & 0.3347 & 0.2834 & 0.2879 & 0.2972 & 0.0323 & 2.1536 & 5.7787 \\
-0.1 & 0.3082 & 0.2631 & 0.2681 & 0.2766 & 0.0323 & 2.1536 & 5.8094 \\
 0.0 & 0.2792 & 0.2335 & 0.2379 & 0.2443 & 0.0323 & 2.1536 & 5.8335 \\
 0.1 & 0.2423 & 0.2041 & 0.2068 & 0.2160 & 0.0323 & 2.1536 & 5.8410 \\
 0.2 & 0.2140 & 0.1758 & 0.1794 & 0.1854 & 0.0323 & 2.1536 & 5.8421 \\
 0.3 & 0.1753 & 0.1435 & 0.1457 & 0.1510 & 0.0323 & 2.1536 & 5.8261 \\
 0.4 & 0.1456 & 0.1138 & 0.1154 & 0.1220 & 0.0323 & 2.1536 & 5.7996 \\
 0.5 & 0.1128 & 0.0809 & 0.0815 & 0.0847 & 0.0323 & 2.1536 & 5.7536 \\
 0.6 & 0.0927 & 0.0601 & 0.0614 & 0.0653 & 0.0323 & 2.1536 & 5.7132 \\
 0.7 & 0.0579 & 0.0429 & 0.0436 & 0.0462 & 0.0323 & 2.1536 & 5.6700 \\
 0.8 & 0.0259 & 0.0280 & 0.0283 & 0.0279 & 0.0323 & 2.1536 & 5.6185 \\
 0.9 & 0.0117 & 0.0058 & 0.0059 & 0.0060 & 0.0323 & 2.1536 & 5.5615 \\
 1.0 & -0.0009 & -0.0069 & -0.0068 & -0.0060 & 0.0323 & 2.1536 & 5.5249 \\
\midrule
\multicolumn{1}{c||}{$\left|\rho_{s}\right|$} & 1.0000 & 1.0000 & 1.0000 & 1.0000 & 0.0000 & 0.0000 & 0.1558 \\
\bottomrule
\end{tabularx}
} 
\caption{Comparison of similarity metrics across varying cosine similarity values under the Logistic Regression setting}  
\label{table1.4.1}
\end{table}

\begin{table}[htbp]
\centering
\scalebox{0.9}{ 
\setlength{\tabcolsep}{2pt} 
\renewcommand{\arraystretch}{0.9} 
\begin{tabularx}{\textwidth}{>{\centering\arraybackslash}X||
>{\centering\arraybackslash}X 
>{\centering\arraybackslash}X 
>{\centering\arraybackslash}X 
>{\centering\arraybackslash}X |
>{\centering\arraybackslash}X 
>{\centering\arraybackslash}X 
>{\centering\arraybackslash}X 
>{\centering\arraybackslash}X }
\toprule
\makecell{\textbf{Cosine}\\\textbf{Similar.}} &
\makecell{\textbf{Logistic}\\\textbf{Regr.}} &
\makecell{\textbf{SVM}\\\textbf{Linear}} &
\makecell{\textbf{SVM}\\\textbf{Radial}} &
\makecell{\textbf{Xgb}\\\textbf{Tree}} &
\makecell{\textbf{Score}\\\textbf{Unw.}\\\textbf{Avg}} &
\makecell{\textbf{Score}\\\textbf{Wtd.}\\\textbf{Avg}} &
\makecell{\textbf{Score}\\\textbf{Ens.}} &
\makecell{\textbf{Score}\\\textbf{Oracle}} \\
\midrule
-1.0 & 0.4789 & 0.4771 & 0.4693 & 0.3982 & 0.4559 & 0.4624 & 0.4756 & 0.5148 \\
-0.9 & 0.4647 & 0.4614 & 0.4558 & 0.3949 & 0.4442 & 0.4486 & 0.4605 & 0.5044 \\
-0.8 & 0.4477 & 0.4448 & 0.4410 & 0.3800 & 0.4284 & 0.4330 & 0.4452 & 0.4872 \\
-0.7 & 0.4316 & 0.4265 & 0.4216 & 0.3623 & 0.4105 & 0.4155 & 0.4256 & 0.4618 \\
-0.6 & 0.4072 & 0.4006 & 0.4002 & 0.3406 & 0.3872 & 0.3923 & 0.4027 & 0.4406 \\
-0.5 & 0.3846 & 0.3790 & 0.3780 & 0.3202 & 0.3654 & 0.3698 & 0.3826 & 0.4271 \\
-0.4 & 0.3593 & 0.3560 & 0.3530 & 0.2977 & 0.3415 & 0.3462 & 0.3582 & 0.3976 \\
-0.3 & 0.3330 & 0.3257 & 0.3231 & 0.2725 & 0.3136 & 0.3179 & 0.3292 & 0.3704 \\
-0.2 & 0.3029 & 0.2975 & 0.2913 & 0.2419 & 0.2834 & 0.2879 & 0.2972 & 0.3347 \\
-0.1 & 0.2815 & 0.2780 & 0.2687 & 0.2242 & 0.2631 & 0.2681 & 0.2766 & 0.3082 \\
 0.0 & 0.2536 & 0.2449 & 0.2389 & 0.1965 & 0.2335 & 0.2379 & 0.2443 & 0.2792 \\
 0.1 & 0.2215 & 0.2116 & 0.2055 & 0.1777 & 0.2041 & 0.2068 & 0.2160 & 0.2423 \\
 0.2 & 0.1929 & 0.1873 & 0.1760 & 0.1469 & 0.1758 & 0.1794 & 0.1854 & 0.2140 \\
 0.3 & 0.1550 & 0.1509 & 0.1443 & 0.1240 & 0.1435 & 0.1457 & 0.1510 & 0.1753 \\
 0.4 & 0.1238 & 0.1187 & 0.1093 & 0.1034 & 0.1138 & 0.1154 & 0.1220 & 0.1456 \\
 0.5 & 0.0891 & 0.0817 & 0.0797 & 0.0731 & 0.0809 & 0.0815 & 0.0847 & 0.1128 \\
 0.6 & 0.0672 & 0.0601 & 0.0545 & 0.0587 & 0.0601 & 0.0614 & 0.0653 & 0.0927 \\
 0.7 & 0.0482 & 0.0463 & 0.0375 & 0.0397 & 0.0429 & 0.0436 & 0.0462 & 0.0579 \\
 0.8 & 0.0338 & 0.0295 & 0.0229 & 0.0258 & 0.0280 & 0.0283 & 0.0279 & 0.0259 \\
 0.9 & 0.0105 & 0.0074 & 0.0016 & 0.0037 & 0.0058 & 0.0059 & 0.0060 & 0.0117 \\
 1.0 & -0.0057 & -0.0095 & -0.0061 & -0.0062 & -0.0069 & -0.0068 & -0.0060 & -0.0009 \\
\midrule
\textbf{Diff.} & 0.0256 & 0.0302 & 0.0351 & 0.0680 & 0.0397 & 0.0365 & 0.0291 & - \\
\bottomrule
\end{tabularx}
}
\caption{$\text{Diff}$ between $\widehat{CLS}$ and oracle CLS under the Logistic Regression setting}
\label{table1.4.1w}
\end{table}

\subsubsection{Probit Model Setting}

Consider a binary classification problem where the features \( \mathbf{x}_i \in \mathbb{R}^p \) (for sample index \( i = 1, \ldots, n_t \) for the target dataset and \( i = 1, \ldots, n_s \) for the source dataset) are independently and identically distributed (i.i.d.) as \( \mathbf{x}_i \sim \mathcal{N}_p(0, I_p) \), and the relationship between features and labels follows a linear model
\[
y_i = \mathbb{I}[(\beta^\top \mathbf{x}_i + \xi_i) \geq 0],
\]
where \( \beta \in \mathbb{R}^p \) is the parameter vector, \( \xi_i \sim \mathcal{N}(0,1) \) is a random noise term, and \( \mathbb{I}[\cdot] \) denotes the indicator function. This formulation corresponds to the Probit model.

We consider two datasets: the target dataset, characterized by the parameter vector \( \beta^{(t)} \), and the source dataset, characterized by the parameter vector \( \beta^{(s)} \). We replace \( \beta \) in the above linear Probit model with \( \beta^{(t)} \) and \( \beta^{(s)} \) to generate the target and source datasets, respectively. The coefficient vector \( \beta^{(t)} \) for the target data is drawn from \( \mathcal{N}_p\!\left(\tfrac{1}{4}\mathbf{1}, \tfrac{1}{16}\mathbf{I}_p\right) \), where \( \mathbf{1} \) denotes a \( p \)-dimensional vector of ones and \( \mathbf{I}_p \) denotes the \( p \times p \) identity matrix. The vector \( \beta^{(s)} \) is generated using the method described in Section~\ref{subsec:generate} to control the cosine similarity between \( \beta^{(t)} \) and \( \beta^{(s)} \). We set the feature dimension \( p = 10 \).

The Oracle CLS is computed using the formula derived in the proof of Theorem 1.

\begin{figure}[!htbp]
    \centering
    \includegraphics[width=0.8\textwidth]{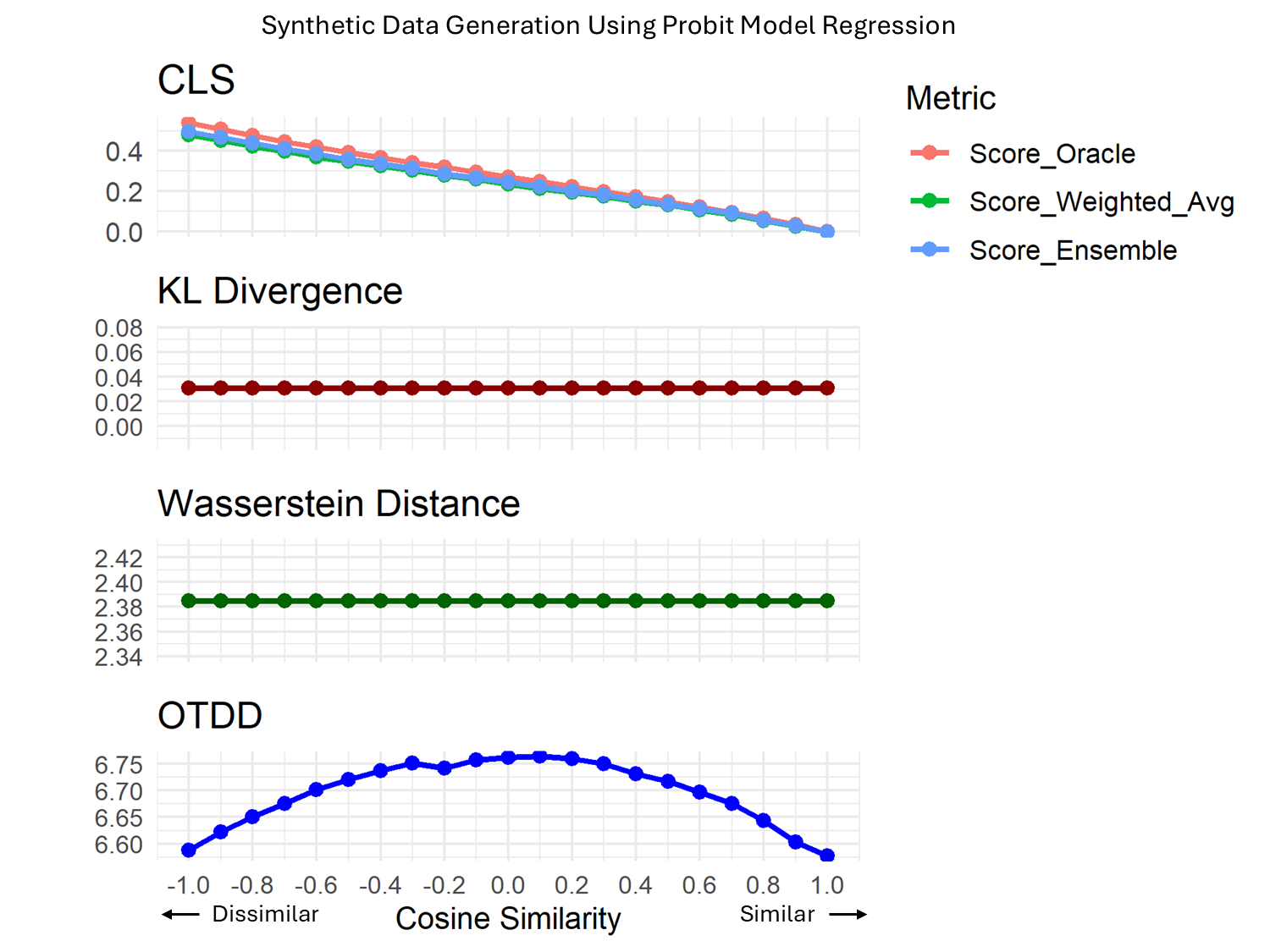}
    \caption{Comparison of CLS vs other similarity metrics
under the Probit Model}
    \label{fig:na_probit}
\end{figure}

\begin{figure}[!htbp]
    \centering
    \includegraphics[width=0.7\textwidth]{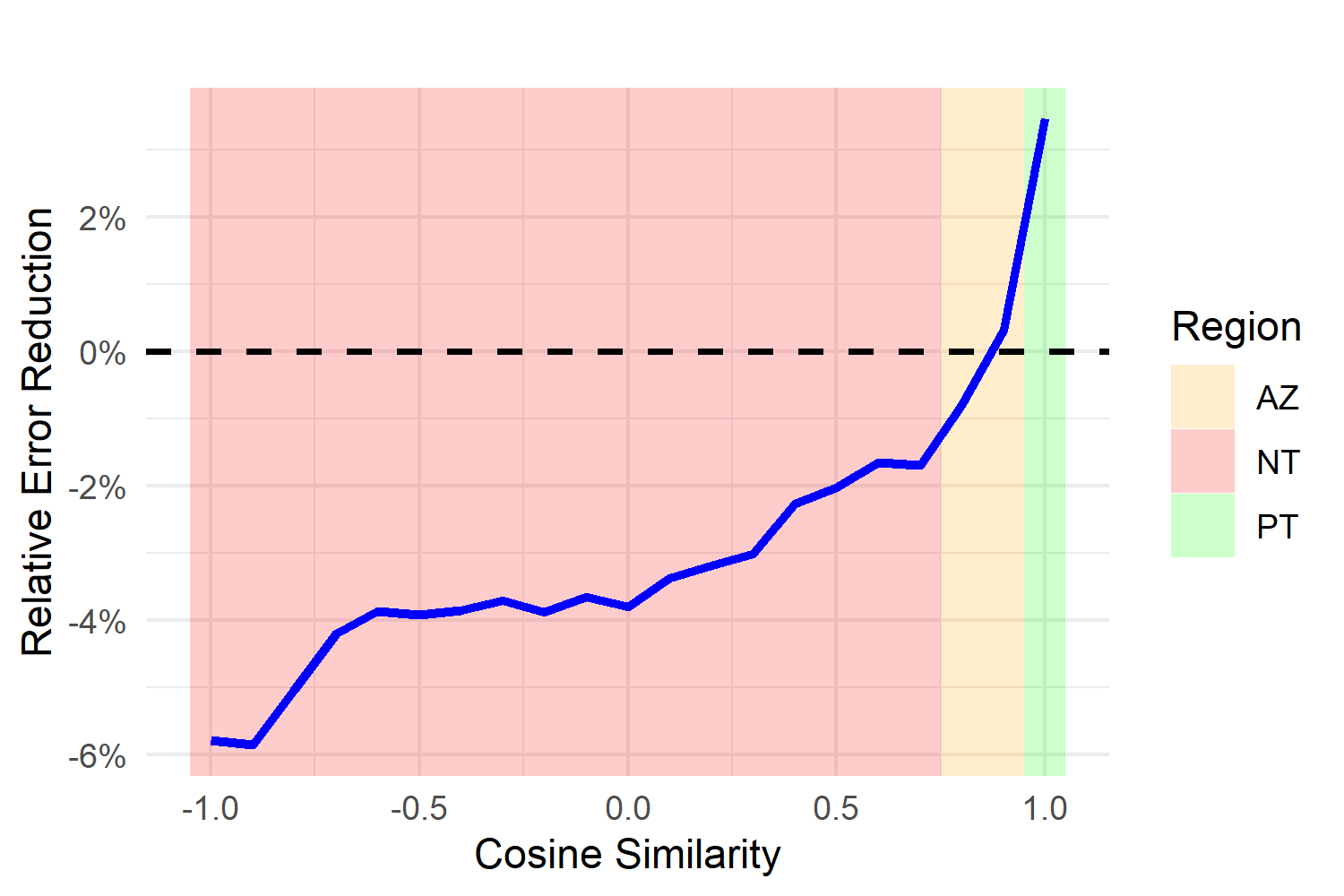}
    \caption{Transferability zones (positive, ambiguous,
and negative) identified under the probit model}
    \label{fig:trans_probit}
\end{figure}

All results for this setting are presented in Figure~\ref{fig:na_probit}, Table~\ref{table1.4.2}, Table~\ref{table1.4.2w}, and Figure~\ref{fig:trans_probit}.

\begin{table}[htbp]
\centering
\scalebox{0.9}{ 
\setlength{\tabcolsep}{2pt} 
\renewcommand{\arraystretch}{0.9} 
\begin{tabularx}{\textwidth}{
>{\centering\arraybackslash}X||
>{\centering\arraybackslash}X 
>{\centering\arraybackslash}X 
>{\centering\arraybackslash}X 
>{\centering\arraybackslash}X 
>{\centering\arraybackslash}X 
>{\centering\arraybackslash}X 
>{\centering\arraybackslash}X }
\toprule
\makecell{\textbf{Cosine}\\\textbf{Sim.}} &
\makecell{\textbf{Score}\\\textbf{Oracle}} &
\makecell{\textbf{Score}\\\textbf{Unw.}\\\textbf{Avg}} &
\makecell{\textbf{Score}\\\textbf{Wtd.}\\\textbf{Avg}} &
\makecell{\textbf{Score}\\\textbf{Ens.}} &
\makecell{\textbf{KL}\\\textbf{Div.}} &
\makecell{\textbf{Wass.}\\\textbf{Dist.}} &
\textbf{OTDD} \\
\midrule
-1.0 & 0.5410 & 0.4632 & 0.4791 & 0.4951 & 0.0307 & 2.385 & 6.5877 \\
-0.9 & 0.5068 & 0.4365 & 0.4511 & 0.4677 & 0.0307 & 2.385 & 6.6223 \\
-0.8 & 0.4757 & 0.4105 & 0.4240 & 0.4380 & 0.0307 & 2.385 & 6.6510 \\
-0.7 & 0.4467 & 0.3858 & 0.3980 & 0.4113 & 0.0307 & 2.385 & 6.6751 \\
-0.6 & 0.4193 & 0.3581 & 0.3688 & 0.3851 & 0.0307 & 2.385 & 6.7016 \\
-0.5 & 0.3930 & 0.3358 & 0.3471 & 0.3585 & 0.0307 & 2.385 & 6.7205 \\
-0.4 & 0.3676 & 0.3157 & 0.3255 & 0.3362 & 0.0307 & 2.385 & 6.7369 \\
-0.3 & 0.3428 & 0.2959 & 0.3043 & 0.3125 & 0.0307 & 2.385 & 6.7511 \\
-0.2 & 0.3185 & 0.2697 & 0.2777 & 0.2853 & 0.0307 & 2.385 & 6.7414 \\
-0.1 & 0.2944 & 0.2526 & 0.2591 & 0.2678 & 0.0307 & 2.385 & 6.7576 \\
 0.0 & 0.2705 & 0.2290 & 0.2349 & 0.2430 & 0.0307 & 2.385 & 6.7623 \\
 0.1 & 0.2466 & 0.2083 & 0.2132 & 0.2218 & 0.0307 & 2.385 & 6.7642 \\
 0.2 & 0.2225 & 0.1900 & 0.1948 & 0.2012 & 0.0307 & 2.385 & 6.7591 \\
 0.3 & 0.1981 & 0.1706 & 0.1743 & 0.1806 & 0.0307 & 2.385 & 6.7506 \\
 0.4 & 0.1734 & 0.1488 & 0.1516 & 0.1552 & 0.0307 & 2.385 & 6.7316 \\
 0.5 & 0.1479 & 0.1280 & 0.1309 & 0.1357 & 0.0307 & 2.385 & 6.7165 \\
 0.6 & 0.1217 & 0.1034 & 0.1053 & 0.1127 & 0.0307 & 2.385 & 6.6966 \\
 0.7 & 0.0943 & 0.0829 & 0.0844 & 0.0898 & 0.0307 & 2.385 & 6.6759 \\
 0.8 & 0.0653 & 0.0532 & 0.0534 & 0.0564 & 0.0307 & 2.385 & 6.6439 \\
 0.9 & 0.0342 & 0.0259 & 0.0256 & 0.0280 & 0.0307 & 2.385 & 6.6035 \\
 1.0 & 0.0000 & -0.0025 & -0.0029 & -0.0029 & 0.0307 & 2.385 & 6.5771 \\
\midrule
\multicolumn{1}{c||}{$\left|\rho_{s}\right|$} & 1.0000 & 1.0000 & 1.0000 & 1.0000 & 0.0000 & 0.0000 & 0.0351 \\
\bottomrule
\end{tabularx}
} 
\caption{Comparison of similarity metrics across varying cosine similarity values under the Probit Model setting}
\label{table1.4.2}
\end{table}

\begin{table}[htbp]
\centering
\scalebox{0.9}{ 
\setlength{\tabcolsep}{2pt} 
\renewcommand{\arraystretch}{0.9} 
\begin{tabularx}{\textwidth}{>{\centering\arraybackslash}X||
>{\centering\arraybackslash}X 
>{\centering\arraybackslash}X 
>{\centering\arraybackslash}X 
>{\centering\arraybackslash}X |
>{\centering\arraybackslash}X 
>{\centering\arraybackslash}X 
>{\centering\arraybackslash}X 
>{\centering\arraybackslash}X }
\toprule
\makecell{\textbf{Cos.}\\\textbf{Sim.}} &
\makecell{\textbf{Log.}\\\textbf{Regr.}} &
\makecell{\textbf{SVM}\\\textbf{Linear}} &
\makecell{\textbf{SVM}\\\textbf{Radial}} &
\makecell{\textbf{Xgb}\\\textbf{Tree}} &
\makecell{\textbf{Score}\\\textbf{Unw.}\\\textbf{Avg}} &
\makecell{\textbf{Score}\\\textbf{Wtd.}\\\textbf{Avg}} &
\makecell{\textbf{Score}\\\textbf{Ens.}} &
\makecell{\textbf{Score}\\\textbf{Oracle}} \\
\midrule
-1.0 & 0.5085 & 0.4998 & 0.4595 & 0.3850 & 0.4632 & 0.4791 & 0.4951 & 0.5410 \\
-0.9 & 0.4783 & 0.4720 & 0.4318 & 0.3639 & 0.4365 & 0.4511 & 0.4677 & 0.5068 \\
-0.8 & 0.4502 & 0.4426 & 0.4082 & 0.3410 & 0.4105 & 0.4240 & 0.4380 & 0.4757 \\
-0.7 & 0.4225 & 0.4153 & 0.3879 & 0.3174 & 0.3858 & 0.3980 & 0.4113 & 0.4467 \\
-0.6 & 0.3928 & 0.3853 & 0.3584 & 0.2958 & 0.3581 & 0.3688 & 0.3851 & 0.4193 \\
-0.5 & 0.3702 & 0.3643 & 0.3368 & 0.2719 & 0.3358 & 0.3471 & 0.3585 & 0.3930 \\
-0.4 & 0.3484 & 0.3404 & 0.3159 & 0.2582 & 0.3157 & 0.3255 & 0.3362 & 0.3676 \\
-0.3 & 0.3251 & 0.3202 & 0.2976 & 0.2408 & 0.2959 & 0.3043 & 0.3125 & 0.3428 \\
-0.2 & 0.2995 & 0.2922 & 0.2688 & 0.2184 & 0.2697 & 0.2777 & 0.2853 & 0.3185 \\
-0.1 & 0.2787 & 0.2719 & 0.2534 & 0.2064 & 0.2526 & 0.2591 & 0.2678 & 0.2944 \\
 0.0 & 0.2524 & 0.2459 & 0.2301 & 0.1876 & 0.2290 & 0.2349 & 0.2430 & 0.2705 \\
 0.1 & 0.2288 & 0.2228 & 0.2092 & 0.1726 & 0.2083 & 0.2132 & 0.2218 & 0.2466 \\
 0.2 & 0.2086 & 0.2031 & 0.1915 & 0.1569 & 0.1900 & 0.1948 & 0.2012 & 0.2225 \\
 0.3 & 0.1838 & 0.1831 & 0.1720 & 0.1436 & 0.1706 & 0.1743 & 0.1806 & 0.1981 \\
 0.4 & 0.1598 & 0.1580 & 0.1517 & 0.1258 & 0.1488 & 0.1516 & 0.1552 & 0.1734 \\
 0.5 & 0.1393 & 0.1351 & 0.1297 & 0.1077 & 0.1280 & 0.1309 & 0.1357 & 0.1479 \\
 0.6 & 0.1126 & 0.1079 & 0.1022 & 0.0908 & 0.1034 & 0.1053 & 0.1127 & 0.1217 \\
 0.7 & 0.0919 & 0.0864 & 0.0814 & 0.0718 & 0.0829 & 0.0844 & 0.0898 & 0.0943 \\
 0.8 & 0.0572 & 0.0535 & 0.0548 & 0.0473 & 0.0532 & 0.0534 & 0.0564 & 0.0653 \\
 0.9 & 0.0288 & 0.0274 & 0.0269 & 0.0205 & 0.0259 & 0.0256 & 0.0280 & 0.0342 \\
 1.0 & -0.0015 & -0.0032 & 0.0020 & -0.0071 & -0.0025 & -0.0029 & -0.0029 & 0.0000 \\
\midrule
\textbf{Diff.} & 0.0164 & 0.0217 & 0.0388 & 0.0792 & 0.0390 & 0.0324 & 0.0239 & - \\
\bottomrule
\end{tabularx}
}
\caption{$\text{Diff}$ between $\widehat{CLS}$ and oracle CLS under the Probit Model setting}
\label{table1.4.2w}
\end{table}

\subsubsection{Linear Discriminant Analysis(LDA) Setting}

We consider a simple Linear Discriminant Analysis (LDA) setting. We generate the target dataset with \( n_t \) i.i.d. samples \( \{(\mathbf{x}_i^{(t)}, y_i^{(t)})\}_{i=1}^{n_t} \), and the source dataset with \( n_s \) i.i.d. samples \( \{(\mathbf{x}_i^{(s)}, y_i^{(s)})\}_{i=1}^{n_s} \), and we set the feature dimension $p=10$. The class labels take values in \( \{0, 1\} \), and for both datasets, we assign exactly half of the samples to each class in order to ensure perfect class balance and reduce label variance during evaluation.

Conditional on the class label \( l \in \{0, 1\} \), the features follow Gaussian distributions:
\[
\mathbf{X}_i^{(t)} \mid Y_i^{(t)} = l \sim \mathcal{N}_p\left( (-1)^{1 - l} \mu^{(t)}, I_p \right), \quad 
\mathbf{X}_i^{(s)} \mid Y_i^{(s)} = l \sim \mathcal{N}_p\left( (-1)^{1 - l} \mu^{(s)}, I_p \right),
\]
where \( \mu^{(t)}, \mu^{(s)} \in \mathbb{R}^p \) are the class means for the target and source data, respectively, and \( I_p \) is the \( p \times p \) identity matrix.

The coefficient vector $\mu^{(t)}$ for target data is set to $\mu^{(t)} = (0.3, 0.3, \ldots, 0.3) \in \mathbb{R}^{p}$. The vector \( \mu^{(s)} \) is generated using the method described in Section~\ref{subsec:generate} to control the cosine similarity between \( \mu^{(t)} \) and \( \mu^{(s)} \).

The Oracle CLS is derived in Theorem 2.

\begin{figure}[!htbp]
    \centering
    \includegraphics[width=0.8\textwidth]{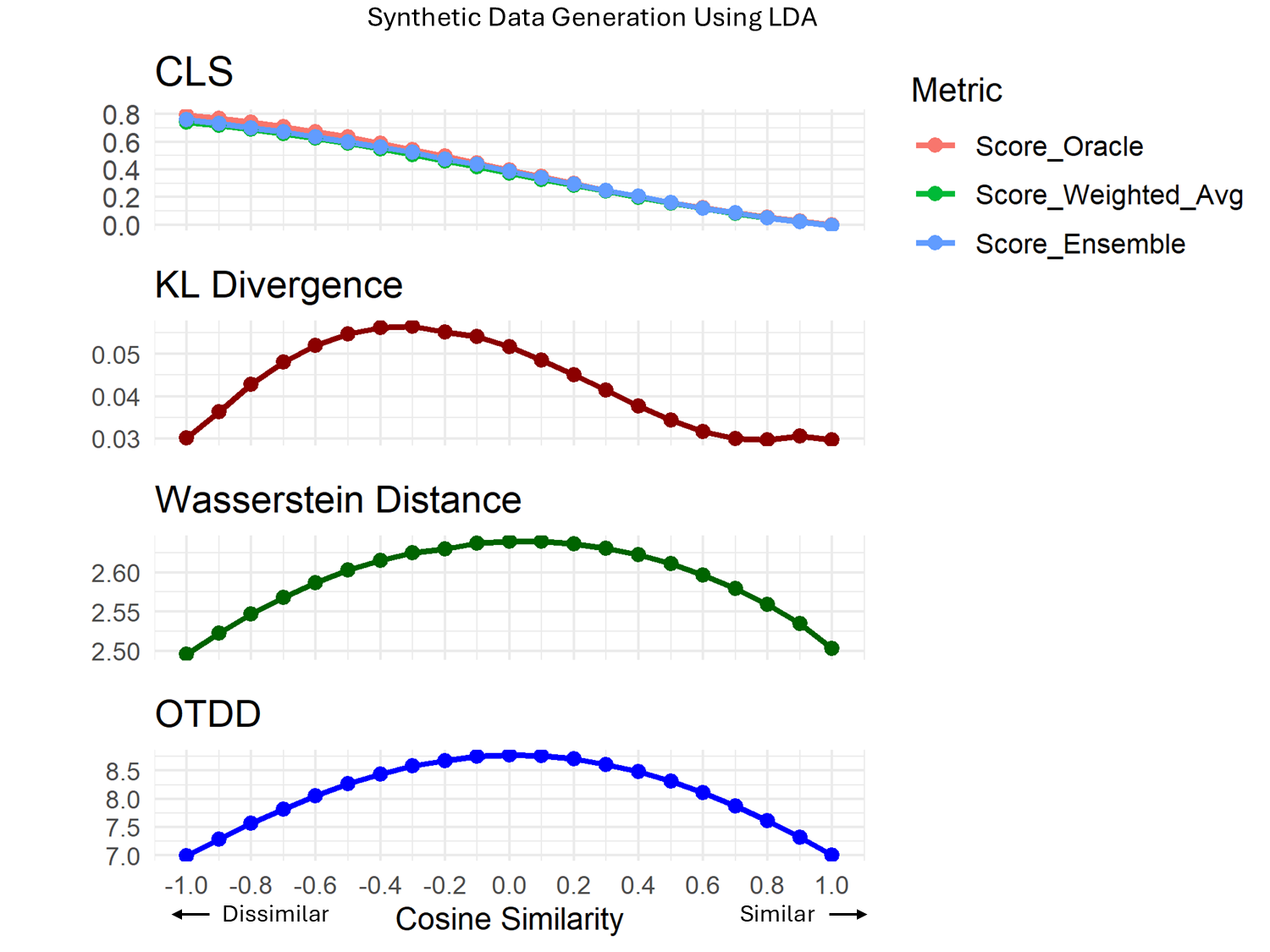}
    \caption{Comparison of CLS vs other similarity metrics
under the LDA setting}
    \label{fig:na_LDA}
\end{figure}

\begin{figure}[!htbp]
    \centering
    \includegraphics[width=0.7\textwidth]{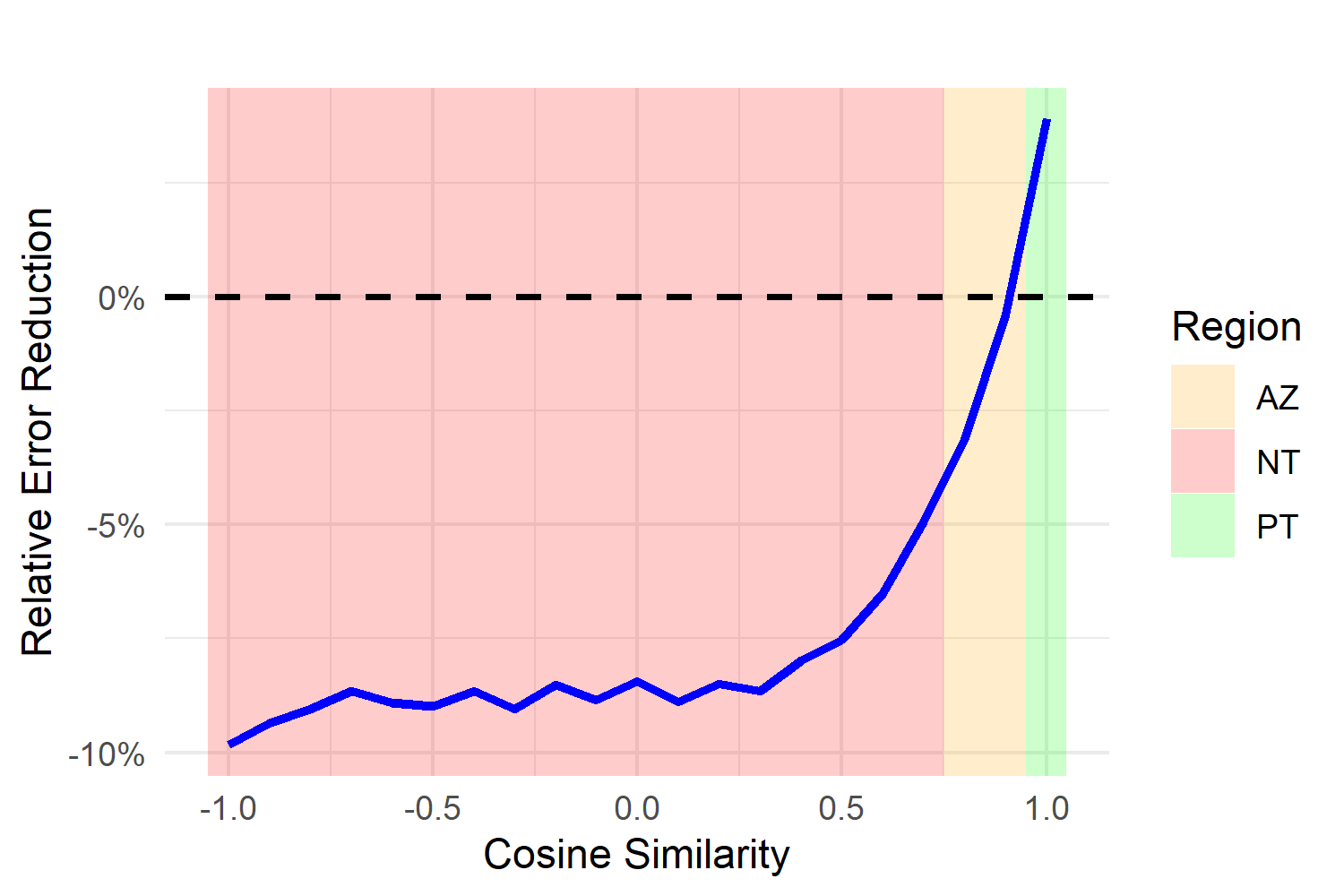}
    \caption{Transferability zones (positive, ambiguous,
and negative) identified under the LDA model}
    \label{fig:trans_LDA}
\end{figure}

All results for this setting are presented in Table~\ref{table1.4.3}, Table~\ref{table1.4.3w}, Figure~\ref{fig:na_LDA}, and Figure~\ref{fig:trans_LDA}.

\begin{table}[htbp]
\centering
\scalebox{0.9}{ 
\setlength{\tabcolsep}{2pt} 
\renewcommand{\arraystretch}{0.9} 
\begin{tabularx}{\textwidth}{
>{\centering\arraybackslash}X||
>{\centering\arraybackslash}X 
>{\centering\arraybackslash}X 
>{\centering\arraybackslash}X 
>{\centering\arraybackslash}X 
>{\centering\arraybackslash}X 
>{\centering\arraybackslash}X 
>{\centering\arraybackslash}X }
\toprule
\makecell{\textbf{Cosine}\\\textbf{Sim.}} &
\makecell{\textbf{Score}\\\textbf{Oracle}} &
\makecell{\textbf{Score}\\\textbf{Unw.}\\\textbf{Avg}} &
\makecell{\textbf{Score}\\\textbf{Wtd.}\\\textbf{Avg}} &
\makecell{\textbf{Score}\\\textbf{Ens.}} &
\makecell{\textbf{KL}\\\textbf{Div.}} &
\makecell{\textbf{Wass.}\\\textbf{Dist.}} &
\textbf{OTDD} \\
\midrule
-1.0 & 0.7941 & 0.7358 & 0.7432 & 0.7608 & 0.0301 & 2.4954 & 6.9912 \\
-0.9 & 0.7696 & 0.7119 & 0.7180 & 0.7342 & 0.0364 & 2.5225 & 7.2830 \\
-0.8 & 0.7413 & 0.6840 & 0.6890 & 0.7016 & 0.0428 & 2.5470 & 7.5616 \\
-0.7 & 0.7091 & 0.6536 & 0.6579 & 0.6708 & 0.0480 & 2.5682 & 7.8205 \\
-0.6 & 0.6731 & 0.6197 & 0.6237 & 0.6361 & 0.0521 & 2.5868 & 8.0542 \\
-0.5 & 0.6335 & 0.5836 & 0.5870 & 0.5975 & 0.0548 & 2.6026 & 8.2636 \\
-0.4 & 0.5906 & 0.5455 & 0.5486 & 0.5610 & 0.0563 & 2.6155 & 8.4406 \\
-0.3 & 0.5449 & 0.5038 & 0.5067 & 0.5222 & 0.0565 & 2.6253 & 8.5818 \\
-0.2 & 0.4969 & 0.4577 & 0.4602 & 0.4749 & 0.0551 & 2.6302 & 8.6780 \\
-0.1 & 0.4474 & 0.4170 & 0.4188 & 0.4364 & 0.0541 & 2.6375 & 8.7489 \\
 0.0 & 0.3970 & 0.3701 & 0.3711 & 0.3880 & 0.0516 & 2.6397 & 8.7747 \\
 0.1 & 0.3467 & 0.3262 & 0.3266 & 0.3403 & 0.0486 & 2.6395 & 8.7606 \\
 0.2 & 0.2972 & 0.2829 & 0.2824 & 0.2936 & 0.0451 & 2.6365 & 8.7053 \\
 0.3 & 0.2492 & 0.2418 & 0.2408 & 0.2478 & 0.0414 & 2.6309 & 8.6108 \\
 0.4 & 0.2035 & 0.1998 & 0.1984 & 0.2049 & 0.0377 & 2.6223 & 8.4785 \\
 0.5 & 0.1606 & 0.1577 & 0.1559 & 0.1600 & 0.0344 & 2.6109 & 8.3089 \\
 0.6 & 0.1210 & 0.1196 & 0.1187 & 0.1192 & 0.0317 & 2.5968 & 8.1058 \\
 0.7 & 0.0850 & 0.0835 & 0.0834 & 0.0836 & 0.0300 & 2.5796 & 7.8720 \\
 0.8 & 0.0528 & 0.0499 & 0.0505 & 0.0506 & 0.0297 & 2.5594 & 7.6094 \\
 0.9 & 0.0245 & 0.0229 & 0.0235 & 0.0232 & 0.0306 & 2.5350 & 7.3221 \\
 1.0 & 0.0000 & -0.0053 & -0.0054 & -0.0054 & 0.0297 & 2.5030 & 7.0043 \\
\midrule
\multicolumn{1}{c||}{$\rho_s$} & -1.0000 & -1.0000 & -1.0000 & -1.0000 & -0.5299 & 0.0779 & 0.0714 \\
\bottomrule
\end{tabularx}
}
\caption{Comparison of similarity metrics across varying cosine similarity values under the LDA setting}
\label{table1.4.3}
\end{table}

\begin{table}[htbp]
\centering
\scalebox{0.9}{ 
\setlength{\tabcolsep}{2pt} 
\renewcommand{\arraystretch}{0.9} 
\begin{tabularx}{\textwidth}{>{\centering\arraybackslash}X||
>{\centering\arraybackslash}X 
>{\centering\arraybackslash}X 
>{\centering\arraybackslash}X 
>{\centering\arraybackslash}X |
>{\centering\arraybackslash}X 
>{\centering\arraybackslash}X 
>{\centering\arraybackslash}X 
>{\centering\arraybackslash}X }
\toprule
\makecell{\textbf{Cos.}\\\textbf{Sim.}} &
\makecell{\textbf{Log.}\\\textbf{Regr.}} &
\makecell{\textbf{SVM}\\\textbf{Linear}} &
\makecell{\textbf{SVM}\\\textbf{Radial}} &
\makecell{\textbf{Xgb}\\\textbf{Tree}} &
\makecell{\textbf{Score}\\\textbf{Unw.}\\\textbf{Avg}} &
\makecell{\textbf{Score}\\\textbf{Wtd.}\\\textbf{Avg}} &
\makecell{\textbf{Score}\\\textbf{Ens.}} &
\makecell{\textbf{Score}\\\textbf{Oracle}} \\
\midrule
-1.0 & 0.7586 & 0.7577 & 0.7514 & 0.6754 & 0.7358 & 0.7432 & 0.7608 & 0.7941 \\
-0.9 & 0.7346 & 0.7356 & 0.7293 & 0.6482 & 0.7119 & 0.7180 & 0.7342 & 0.7696 \\
-0.8 & 0.7059 & 0.7062 & 0.7024 & 0.6214 & 0.6840 & 0.6890 & 0.7016 & 0.7413 \\
-0.7 & 0.6736 & 0.6774 & 0.6695 & 0.5940 & 0.6536 & 0.6579 & 0.6708 & 0.7091 \\
-0.6 & 0.6384 & 0.6387 & 0.6350 & 0.5668 & 0.6197 & 0.6237 & 0.6361 & 0.6731 \\
-0.5 & 0.6000 & 0.6006 & 0.5945 & 0.5392 & 0.5836 & 0.5870 & 0.5975 & 0.6335 \\
-0.4 & 0.5576 & 0.5642 & 0.5499 & 0.5104 & 0.5455 & 0.5486 & 0.5610 & 0.5906 \\
-0.3 & 0.5166 & 0.5198 & 0.4992 & 0.4798 & 0.5038 & 0.5067 & 0.5222 & 0.5449 \\
-0.2 & 0.4697 & 0.4737 & 0.4471 & 0.4404 & 0.4577 & 0.4602 & 0.4749 & 0.4969 \\
-0.1 & 0.4251 & 0.4273 & 0.4003 & 0.4155 & 0.4170 & 0.4188 & 0.4364 & 0.4474 \\
 0.0 & 0.3754 & 0.3764 & 0.3486 & 0.3799 & 0.3701 & 0.3711 & 0.3880 & 0.3970 \\
 0.1 & 0.3296 & 0.3298 & 0.3005 & 0.3451 & 0.3262 & 0.3266 & 0.3403 & 0.3467 \\
 0.2 & 0.2848 & 0.2828 & 0.2566 & 0.3073 & 0.2829 & 0.2824 & 0.2936 & 0.2972 \\
 0.3 & 0.2415 & 0.2394 & 0.2183 & 0.2679 & 0.2418 & 0.2408 & 0.2478 & 0.2492 \\
 0.4 & 0.1984 & 0.1956 & 0.1789 & 0.2265 & 0.1998 & 0.1984 & 0.2049 & 0.2035 \\
 0.5 & 0.1554 & 0.1534 & 0.1432 & 0.1786 & 0.1577 & 0.1559 & 0.1600 & 0.1606 \\
 0.6 & 0.1193 & 0.1176 & 0.1110 & 0.1306 & 0.1196 & 0.1187 & 0.1192 & 0.1210 \\
 0.7 & 0.0833 & 0.0846 & 0.0793 & 0.0870 & 0.0835 & 0.0834 & 0.0836 & 0.0850 \\
 0.8 & 0.0502 & 0.0513 & 0.0502 & 0.0477 & 0.0499 & 0.0505 & 0.0506 & 0.0528 \\
 0.9 & 0.0226 & 0.0249 & 0.0234 & 0.0207 & 0.0229 & 0.0235 & 0.0232 & 0.0245 \\
 1.0 & -0.0058 & -0.0065 & -0.0066 & -0.0024 & -0.0053 & -0.0054 & -0.0054 & 0.0000 \\
\midrule
\textbf{Diff.} & 0.0192 & 0.0185 & 0.0312 & 0.0486 & 0.0274 & 0.0257 & 0.0162 & - \\
\bottomrule
\end{tabularx}
}
\caption{$\text{Diff}$ between $\widehat{CLS}$ and oracle CLS under the LDA setting}
\label{table1.4.3w}
\end{table}

\newpage
\subsubsection{Mixture Gaussian Setting}

We consider a binary classification setting under a mixture gaussian model with feature dimension \(p = 10\). The label space is \(\{0, 1\}\), and we generate both target and source datasets.

For the target dataset, we generate \(n_t\) i.i.d. samples \(\{(\mathbf{X}_i^{(t)}, Y_i^{(t)})\}_{i=1}^{n_t}\), where the class-conditional distributions are:
\[
\mathbf{X}_i^{(t)} \mid Y_i^{(t)} = 1 \sim \mathcal{N}_p(\mu^{(t)}, I_p), \quad 
\mathbf{X}_i^{(t)} \mid Y_i^{(t)} = 0 \sim \mathcal{N}_p(-\mu^{(t)}, I_p),
\]
with the class mean defined as \(\mu^{(t)} = (0.3, 0.3, \ldots, 0.3) \in \mathbb{R}^{p}\), and \(I_p\) being the \(10 \times 10\) identity matrix.

For the source dataset, we generate \(n_s\) i.i.d. samples \(\{(\mathbf{X}_i^{(s)}, Y_i^{(s)})\}_{i=1}^{n_s}\), where the class-conditional distribution is modeled as a probabilistic mixture of the target and source Gaussian components, controlled by a shift parameter \(\alpha \in [0, 1]\):
\begin{align*}
\mathbf{X}_i^{(s)} \mid Y_i^{(s)} = 1 &\sim (1 - \alpha) \, \mathcal{N}_p(\mu^{(t)}, I_p) + \alpha \, \mathcal{N}_p(\mu^{(s)}, I_p), \\
\mathbf{X}_i^{(s)} \mid Y_i^{(s)} = 0 &\sim (1 - \alpha) \, \mathcal{N}_p(-\mu^{(t)}, I_p) + \alpha \, \mathcal{N}_p(-\mu^{(s)}, I_p),
\end{align*}
where \(\mu^{(s)} \in \mathbb{R}^p\) is the source class-1 mean, each coordinate of \(\mu^{(s)}\) is sampled independently from \(\mathcal{N}(-0.3, 0.5^2)\).

This construction introduces a controllable distribution shift parameter \(\alpha\) between the target and source datasets. Here, the shift parameter $\alpha$ serves as a controllable factor determining the degree of domain divergence. When \(\alpha = 0\), the source distribution matches the target exactly; as \(\alpha\) increases, the source data diverge due to increasing influence from the shifted Gaussian components.

The Oracle CLS is calculated using \(\mu^{(t)},\mu^{(s)},\alpha\) based on the above data generation mechanism.

\begin{figure}[!htbp]
    \centering
    \includegraphics[width=0.8\textwidth]{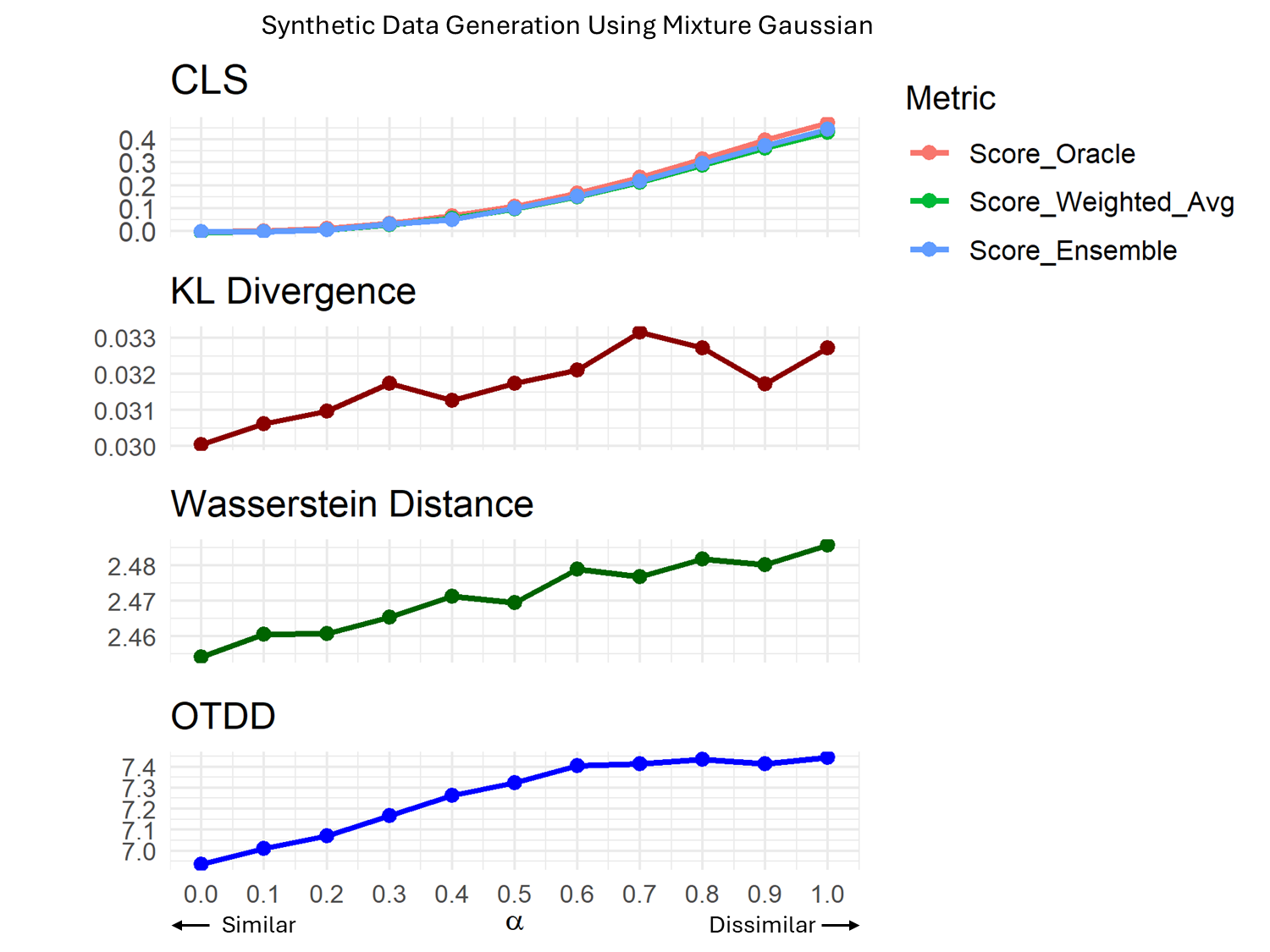}
    \caption{Comparison of CLS vs other similarity metrics
under the Mixture Gaussian Model}
    \label{fig:na_Mix_Gau}
\end{figure}

\begin{figure}[!htbp]
    \centering
    \includegraphics[width=0.7\textwidth]{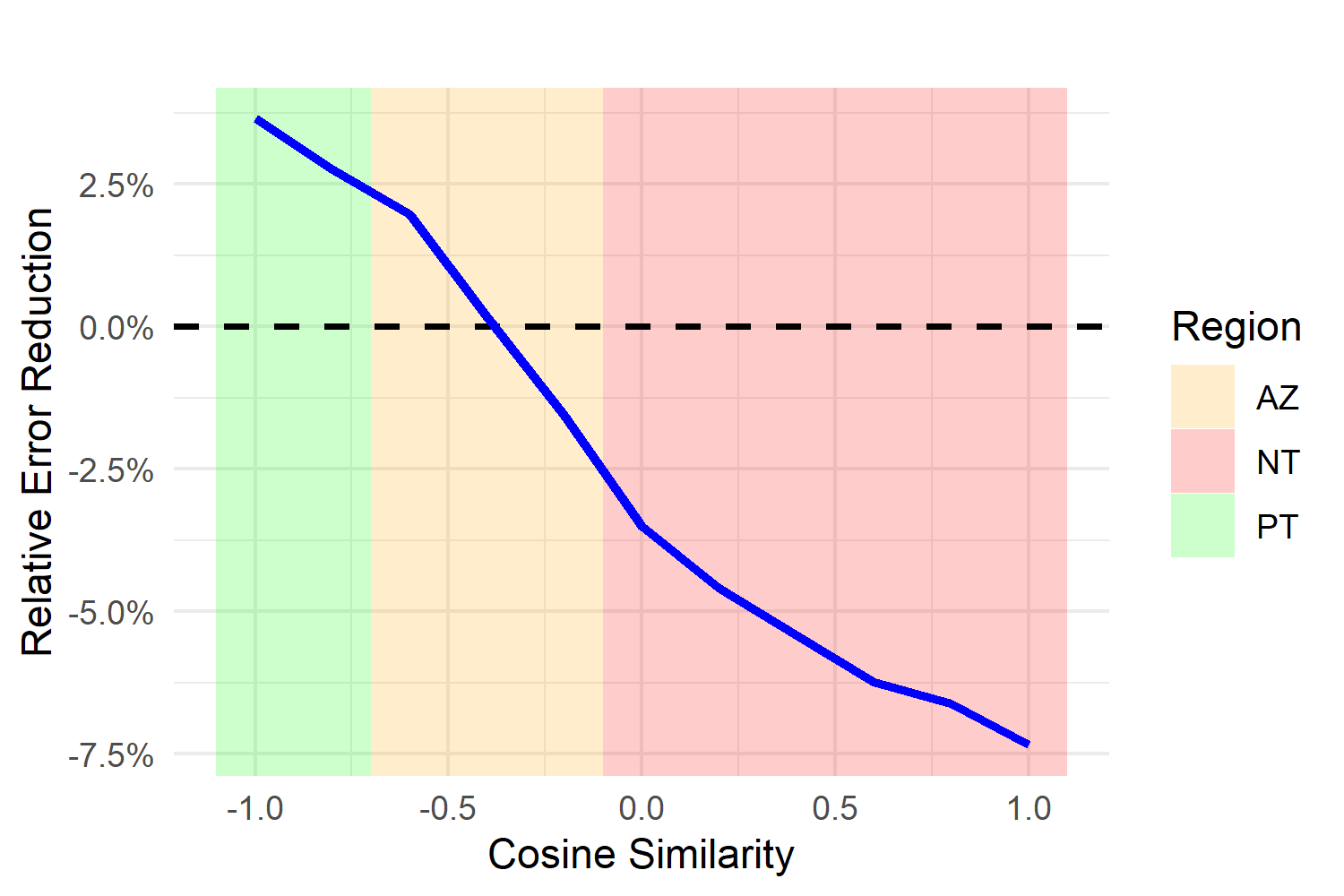}
    \caption{Transferability zones (positive, ambiguous,
and negative) identified under the Mixture Gaussian model}
    \label{fig:trans_Mix_Gau}
\end{figure}

All results for this setting are presented in Table~\ref{table1.4.4}, Table~\ref{table1.4.4w}, Figure~\ref{fig:na_Mix_Gau}, and Figure~\ref{fig:trans_Mix_Gau}.

\begin{table}[htbp]
\centering
\scalebox{0.9}{ 
\setlength{\tabcolsep}{2pt} 
\renewcommand{\arraystretch}{0.9} 
\begin{tabularx}{\textwidth}{
>{\centering\arraybackslash}X||
>{\centering\arraybackslash}X 
>{\centering\arraybackslash}X 
>{\centering\arraybackslash}X 
>{\centering\arraybackslash}X 
>{\centering\arraybackslash}X 
>{\centering\arraybackslash}X 
>{\centering\arraybackslash}X }
\toprule
\makecell{\textbf{Alpha}} &
\makecell{\textbf{Score}\\\textbf{Oracle}} &
\makecell{\textbf{Score}\\\textbf{Unw.}\\\textbf{Avg}} &
\makecell{\textbf{Score}\\\textbf{Wtd.}\\\textbf{Avg}} &
\makecell{\textbf{Score}\\\textbf{Ens.}} &
\makecell{\textbf{KL}\\\textbf{Div.}} &
\makecell{\textbf{Wass.}\\\textbf{Dist.}} &
\textbf{OTDD} \\
\midrule
0.0 & -0.0006 & -0.0042 & -0.0037 & -0.0023 & 0.0300 & 2.4540 & 6.9350 \\
0.1 &  0.0025 & -0.0013 & -0.0012 & -0.0013 & 0.0306 & 2.4606 & 7.0114 \\
0.2 &  0.0128 &  0.0067 &  0.0069 &  0.0061 & 0.0310 & 2.4607 & 7.0709 \\
0.3 &  0.0338 &  0.0285 &  0.0296 &  0.0314 & 0.0317 & 2.4653 & 7.1673 \\
0.4 &  0.0665 &  0.0546 &  0.0556 &  0.0512 & 0.0313 & 2.4712 & 7.2627 \\
0.5 &  0.1077 &  0.0968 &  0.0987 &  0.1007 & 0.0317 & 2.4694 & 7.3231 \\
0.6 &  0.1665 &  0.1473 &  0.1492 &  0.1516 & 0.0321 & 2.4789 & 7.4037 \\
0.7 &  0.2345 &  0.2098 &  0.2132 &  0.2174 & 0.0332 & 2.4767 & 7.4135 \\
0.8 &  0.3149 &  0.2797 &  0.2859 &  0.2954 & 0.0327 & 2.4818 & 7.4338 \\
0.9 &  0.3978 &  0.3558 &  0.3615 &  0.3710 & 0.0317 & 2.4803 & 7.4148 \\
1.0 &  0.4724 &  0.4235 &  0.4295 &  0.4430 & 0.0327 & 2.4858 & 7.4442 \\
\midrule
\multicolumn{1}{c||}{$\rho_s$} & 1.0000 & 1.0000 & 1.0000 & 1.0000 & 0.7909 & 0.9727 & 0.9900 \\
\bottomrule
\end{tabularx}
} 
\caption{Comparison of similarity metrics across varying $\alpha$ values under the Mixture Gaussian setting}
\label{table1.4.4}
\end{table}

\begin{table}[htbp]
\centering
\scalebox{0.9}{ 
\setlength{\tabcolsep}{2pt} 
\renewcommand{\arraystretch}{0.9} 
\begin{tabularx}{\textwidth}{>{\centering\arraybackslash}X||
>{\centering\arraybackslash}X 
>{\centering\arraybackslash}X 
>{\centering\arraybackslash}X 
>{\centering\arraybackslash}X |
>{\centering\arraybackslash}X 
>{\centering\arraybackslash}X 
>{\centering\arraybackslash}X 
>{\centering\arraybackslash}X }
\toprule
\makecell{\textbf{Alpha}} &
\makecell{\textbf{Log.}\\\textbf{Regr.}} &
\makecell{\textbf{SVM}\\\textbf{Linear}} &
\makecell{\textbf{SVM}\\\textbf{Radial}} &
\makecell{\textbf{Xgb}\\\textbf{Tree}} &
\makecell{\textbf{Score}\\\textbf{Unw.}\\\textbf{Avg}} &
\makecell{\textbf{Score}\\\textbf{Wtd.}\\\textbf{Avg}} &
\makecell{\textbf{Score}\\\textbf{Ens.}} &
\makecell{\textbf{Score}\\\textbf{Oracle}} \\
\midrule
0.0 & -0.0062 & -0.0041 &  0.0002 & -0.0067 & -0.0042 & -0.0037 & -0.0023 & -0.0006 \\
0.1 & -0.0007 & -0.0004 & -0.0016 & -0.0026 & -0.0013 & -0.0012 & -0.0013 &  0.0025 \\
0.2 &  0.0079 &  0.0072 &  0.0064 &  0.0051 &  0.0067 &  0.0069 &  0.0061 &  0.0128 \\
0.3 &  0.0291 &  0.0310 &  0.0277 &  0.0262 &  0.0285 &  0.0296 &  0.0314 &  0.0338 \\
0.4 &  0.0587 &  0.0560 &  0.0563 &  0.0474 &  0.0546 &  0.0556 &  0.0512 &  0.0665 \\
0.5 &  0.1028 &  0.1000 &  0.0949 &  0.0895 &  0.0968 &  0.0987 &  0.1007 &  0.1077 \\
0.6 &  0.1510 &  0.1525 &  0.1562 &  0.1295 &  0.1473 &  0.1492 &  0.1516 &  0.1665 \\
0.7 &  0.2227 &  0.2204 &  0.2166 &  0.1794 &  0.2098 &  0.2132 &  0.2174 &  0.2345 \\
0.8 &  0.2958 &  0.2956 &  0.2899 &  0.2377 &  0.2797 &  0.2859 &  0.2954 &  0.3149 \\
0.9 &  0.3735 &  0.3729 &  0.3655 &  0.3111 &  0.3558 &  0.3615 &  0.3710 &  0.3978 \\
1.0 &  0.4444 &  0.4406 &  0.4314 &  0.3777 &  0.4235 &  0.4295 &  0.4430 &  0.4724 \\
\midrule
\textbf{Diff.} & 0.0118 & 0.0125 & 0.0152 & 0.0377 & 0.0192 & 0.0167 & 0.0131 & - \\
\bottomrule
\end{tabularx}
}
\caption{$\text{Diff}$ between $\widehat{CLS}$ and oracle CLS under the Mixture Gaussian setting}
\label{table1.4.4w}
\end{table}

\subsubsection{Quadratic Discriminant Analysis (QDA) Setting}
\label{subsubsec:qda}

We consider a Quadratic Discriminant Analysis (QDA) setting for synthetic data generation in the ``Comparison of CLS Estimation'' subsection. 
We generate the target dataset with $n_t$ i.i.d.\ samples $\{(\mathbf{x}_i^{(t)}, y_i^{(t)})\}_{i=1}^{n_t}$ and the source dataset with $n_s$ i.i.d.\ samples $\{(\mathbf{x}_i^{(s)}, y_i^{(s)})\}_{i=1}^{n_s}$, with feature dimension $p=10$. 
The class labels take values in $\{0,1\}$, and for both datasets we assign exactly half of the samples to each class to ensure perfect class balance.

Conditional on the class label $l \in \{0,1\}$, the features follow Gaussian distributions with class-specific covariance matrices:
\[
\mathbf{X}_i^{(t)} \mid Y_i^{(t)} = l 
\sim 
\mathcal{N}_p\!\left( (-1)^{1-l}\mu^{(t)}, \Sigma_l \right),
\quad
\mathbf{X}_i^{(s)} \mid Y_i^{(s)} = l 
\sim 
\mathcal{N}_p\!\left( (-1)^{1-l}\mu^{(s)}, \Sigma_l \right),
\]
where $\mu^{(t)}, \mu^{(s)} \in \mathbb{R}^p$ denote the class mean vectors for the target and source datasets, respectively.

The class-conditional covariance matrices are defined as
\[
\Sigma_0 = \left(0.8^{|r-c|}\right)_{p\times p},
\qquad
\Sigma_1 = \left(0.2^{|r-c|}\right)_{p\times p},
\]
where $r,c = 1,\dots,p$ index the feature dimensions.

The mean vectors $\mu^{(t)}$ and $\mu^{(s)}$ are generated independently from the Uniform$(0,1)$ distribution. We set $n_t = n_s = 200$, conduct 50 independent replicates, and report the averaged results.

The results for the comparison of CLS estimation methods under this QDA setting are reported in Table~\ref{tab:cls_comparison_qda}.

\begin{table}[htbp]
\centering
\small
\setlength{\tabcolsep}{6pt}
\renewcommand{\arraystretch}{1.1}
\caption{Comparison of CLS estimates and their absolute deviation from Oracle CLS.}
\label{tab:cls_comparison_qda}
\begin{tabular}{lcccccccc}
\toprule
 & \textbf{Log.Reg.} & \textbf{SVM-L} & \textbf{SVM-R} & \textbf{Xgb} 
 & \textbf{Unw.Avg} & \textbf{Wtd.Avg} & \textbf{Ens.} & \textbf{Oracle} \\
\midrule
$\widehat{CLS}$
& 0.2191 & 0.2130 & 0.1221 & 0.1420 
& 0.1740 & 0.1716 & 0.1692 & 0.1437 \\

\midrule

\textbf{Diff.}
& 0.0754 & 0.0693 & 0.0216 & \textbf{0.0018}
& 0.0303 & 0.0279 & 0.0255 & -- \\
\bottomrule
\end{tabular}
\end{table}

\subsubsection{Polynomial Logistic Regression (PLR) Setting}
\label{subsubsec:plr}

We consider a nonlinear data-generating mechanism based on a Polynomial Logistic Regression (PLR) model in the ``Comparison of CLS Estimation'' subsection. 
This setting is designed to introduce complex nonlinear decision boundaries that cannot be captured by linear classifiers.

We generate the target dataset with $n_t$ i.i.d.\ samples 
$\{(\mathbf{x}_i^{(t)}, y_i^{(t)})\}_{i=1}^{n_t}$ 
and the source dataset with $n_s$ i.i.d.\ samples 
$\{(\mathbf{x}_i^{(s)}, y_i^{(s)})\}_{i=1}^{n_s}$, 
with feature dimension $p=10$. 

For both datasets, the feature vectors are drawn independently from a standard multivariate Gaussian distribution:
\[
\mathbf{X}^{(t)} \sim \mathcal{N}_p(\mathbf{0}, I_p),
\qquad
\mathbf{X}^{(s)} \sim \mathcal{N}_p(\mathbf{0}, I_p).
\]

Conditional on the feature vector $\mathbf{X}$, the binary label is generated according to a logistic model with a nonlinear polynomial predictor:
\[
\mathbb{P}(Y=1 \mid \mathbf{X}) 
= 
\frac{1}{1 + \exp\!\left(-f(\mathbf{X})\right)},
\]
where the polynomial function $f(\mathbf{X})$ is defined as
\begin{align*}
f(\mathbf{X}) 
&=
\beta_1 X_1 
+ \beta_2 X_2^2
+ \beta_3 X_3 X_4
+ \beta_4 X_4^3
+ \beta_5 X_5  \\
&\quad
+ \beta_6 X_6 X_7
- \beta_7 X_7^2
+ \beta_8 X_8^5
+ \beta_9 X_9
+ \beta_{10} X_{10}^3 .
\end{align*}

The coefficient vector $\beta = (\beta_1,\dots,\beta_{10})^\top$ 
is specified separately for the target and source datasets, 
thereby inducing distributional differences between the two tasks. 
Binary outcomes are then sampled as
\[
Y \sim \mathrm{Bernoulli}\!\left(
\frac{1}{1 + \exp(-f(\mathbf{X}))}
\right).
\]

This PLR setting generates highly nonlinear class boundaries involving 
quadratic, cubic, interaction, and higher-order terms. 
Unlike the QDA setting, where class separation is determined by Gaussian class-conditional densities with quadratic log-likelihood ratios, 
the PLR model produces decision boundaries governed by nonlinear transformations of the features.

The coefficient vectors $\beta^{(t)}$ and $\beta^{(s)}$ are independently generated from 
$\mathrm{Uniform}(0,1)^{10}$ to induce distributional differences between the target and source tasks. We set $n_t = n_s = 1000$, conduct 50 independent replicates, and report the averaged results.

The results for CLS estimation under this PLR setting are reported in Table~\ref{tab:cls_plr}. 

\begin{table}[htbp]
\centering
\small
\setlength{\tabcolsep}{6pt}
\renewcommand{\arraystretch}{1.1}
\caption{CLS estimates under the PLR setting. 
The coefficient vectors $\beta^{(t)}$ and $\beta^{(s)}$ are independently generated from $\mathrm{Uniform}(0,1)^{10}$.}
\label{tab:cls_plr}
\begin{tabular}{lccccccccc}
\toprule
 & \textbf{Log.Reg.} & \textbf{SVM-L} & \textbf{SVM-R} & \textbf{Xgb} 
 & \textbf{Unw.Avg} & \textbf{Wtd.Avg} & \textbf{Ens.} & \textbf{Oracle} \\
\midrule
$\widehat{\mathrm{CLS}}$
& 0.0768 & 0.0758 & 0.0864 & 0.0913
& 0.0826 & 0.0842 & 0.0984 & 0.1152 \\

\midrule

\textbf{Diff.}
& 0.0383 & 0.0393 & 0.0288 & 0.0238
& 0.0326 & 0.0309 & 0.0167 & -- \\

\bottomrule
\end{tabular}
\end{table}

\subsubsection{Radial Decision Boundary Setting}
\label{subsubsec:radial}

We further consider a radial data-generating mechanism that induces a nonlinear decision boundary determined solely by the Euclidean norm of the feature vector in the ``Comparison of CLS Estimation'' subsection.

We generate the target dataset with $n_t$ i.i.d.\ samples 
$\{(\mathbf{x}_i^{(t)}, y_i^{(t)})\}_{i=1}^{n_t}$ 
and the source dataset with $n_s$ i.i.d.\ samples 
$\{(\mathbf{x}_i^{(s)}, y_i^{(s)})\}_{i=1}^{n_s}$, 
with feature dimension $p=10$ and $n_t=n_s=1000$. 
For both datasets, the feature vectors are independently generated from a standard multivariate Gaussian distribution:
\[
\mathbf{X}^{(t)} \sim \mathcal{N}_p(\mathbf{0}, I_p),
\qquad
\mathbf{X}^{(s)} \sim \mathcal{N}_p(\mathbf{0}, I_p).
\]

Let 
\[
r(\mathbf{X})=\|\mathbf{X}\|_2 = \sqrt{\sum_{j=1}^{p} X_j^2}
\]
denote the Euclidean radius. 
The binary label is deterministically assigned via a spherical threshold rule:
\[
Y = \mathbb{I}\bigl(r(\mathbf{X}) < r_0\bigr).
\]

To induce distributional discrepancy between the two tasks, 
we set different radial thresholds:
\[
r_0^{(t)} = 3.1,
\qquad
r_0^{(s)} = 3.3.
\]

We set $n_t = n_s = 1000$, conduct 50 independent replicates, and report the averaged results.

The results for CLS estimation under this radial setting are reported in Table~\ref{tab:cls_radial}.

\begin{table}[htbp]
\centering
\small
\setlength{\tabcolsep}{6pt}
\renewcommand{\arraystretch}{1.1}
\caption{CLS estimates under the radial setting and $p=10$.}
\label{tab:cls_radial}
\begin{tabular}{lccccccccc}
\toprule
 & \textbf{Log.Reg.} & \textbf{SVM-L} & \textbf{SVM-R} & \textbf{Xgb} 
 & \textbf{Unw.Avg} & \textbf{Wtd.Avg} & \textbf{Ens.} & \textbf{Oracle} \\
\midrule
$\widehat{\mathrm{CLS}}$
& 0.0275 & 0.0208 & 0.0555 & 0.0241
& 0.0320 & 0.0555 & 0.0555 & 0.1091 \\

\midrule
\textbf{Diff.}
& 0.0816 & 0.0884 & 0.0537 & 0.0850
& 0.0772 & 0.0537 & 0.0537 & -- \\
\bottomrule
\end{tabular}
\end{table}

\subsection{Synthetic Data Generation for Regression}
\label{subsec:r}

For regression tasks, we consider two settings: 
(i) linear regression, and 
(ii) nonlinear regression, where features are generated using sinusoidal, polynomial, interaction, and exponential transformations with added Gaussian noise. 

For CLS estimation, we employ four regression models: linear regression, linear SVM, radial basis function (RBF) SVM, and XGBoost.

For the dataset similarity metric OTDD (Optimal Transport Dataset Distance), it is specifically designed for supervised classification tasks, and thus cannot be directly computed in regression settings due to the continuous nature of the labels.

Among the four transfer learning methods evaluated in Section~5.3 (Evaluation of Transferable Zones), both TrAdaBoost and Dynamic TrAdaBoost were originally developed for binary classification tasks and are therefore not applied in the regression experiments. To enable transfer learning in regression settings, we additionally incorporate the TrAdaBoost.R2 algorithm~\citep{pardoe2010boosting}, which extends TrAdaBoost to handle regression problems.

\subsubsection{Linear Regression Setting}

Consider a binary classification problem where the features \( \mathbf{x}_i \in \mathbb{R}^p \) (for sample index \( i = 1, \ldots, n_t \) for the target dataset and \( i = 1, \ldots, n_s \) for the source dataset) are independently and identically distributed (i.i.d.) as \( \mathbf{x}_i \sim \mathcal{N}_p(0, I_p) \), and the relationship between features and labels follows a linear model
\[
y_i = \mathbb{I}[(\beta^\top \mathbf{x}_i + \xi_i) \geq 0],
\]
where \( \beta \in \mathbb{R}^p \) is the parameter vector, \( \xi_i \sim \mathcal{N}(0,1) \) is a random noise term, and \( \mathbb{I}[\cdot] \) denotes the indicator function.

We consider two datasets: the target dataset, characterized by the parameter vector \( \beta^{(t)} \), and the source dataset, characterized by the parameter vector \( \beta^{(s)} \). We replace \( \beta \) in the above linear Probit model with \( \beta^{(t)} \) and \( \beta^{(s)} \) to generate the target and source datasets, respectively. The coefficient vector \( \beta^{(t)} \) for the target data is drawn from \( \mathcal{N}_p\!\left(\tfrac{1}{4}\mathbf{1}, \tfrac{1}{16}\mathbf{I}_p\right) \), where \( \mathbf{1} \) denotes a \( p \)-dimensional vector of ones and \( \mathbf{I}_p \) denotes the \( p \times p \) identity matrix. The vector \( \beta^{(s)} \) is generated using the method described in Section~\ref{subsec:generate} to control the cosine similarity between \( \beta^{(t)} \) and \( \beta^{(s)} \). We set the feature dimension \( p = 10 \).

We consider two datasets: the target dataset, characterized by the parameter vector \( \beta^{(t)} \), and the source dataset, characterized by the parameter vector \( \beta^{(s)} \). We replace \(\beta\) mentioned above in the linear regression model with \( \beta^{(t)},\beta^{(s)}\) to generate the target and source dataset. The coefficient vector $\beta^{(t)}$ for the target data is drawn from $\mathcal{N}_p\!\left(\tfrac{1}{4}\mathbf{1}, \tfrac{1}{16}\mathbf{I}_p\right)$, 
where $\mathbf{1}$ denotes a $p$-dimensional vector of ones, 
and $\mathbf{I}_p$ denotes the $p \times p$ identity matrix. The vector \( \beta^{(s)} \) is generated using the method described in Section~\ref{subsec:generate} to control the cosine similarity between \( \beta^{(t)} \) and \( \beta^{(s)} \). We set the feature dimension $p=10$.

The Oracle CLS is computed using \(\beta^{(t)},
 \beta^{(s)}\) based on the above data generation mechanism.

\begin{figure}[!htbp]
    \centering
    \includegraphics[width=0.8\textwidth]{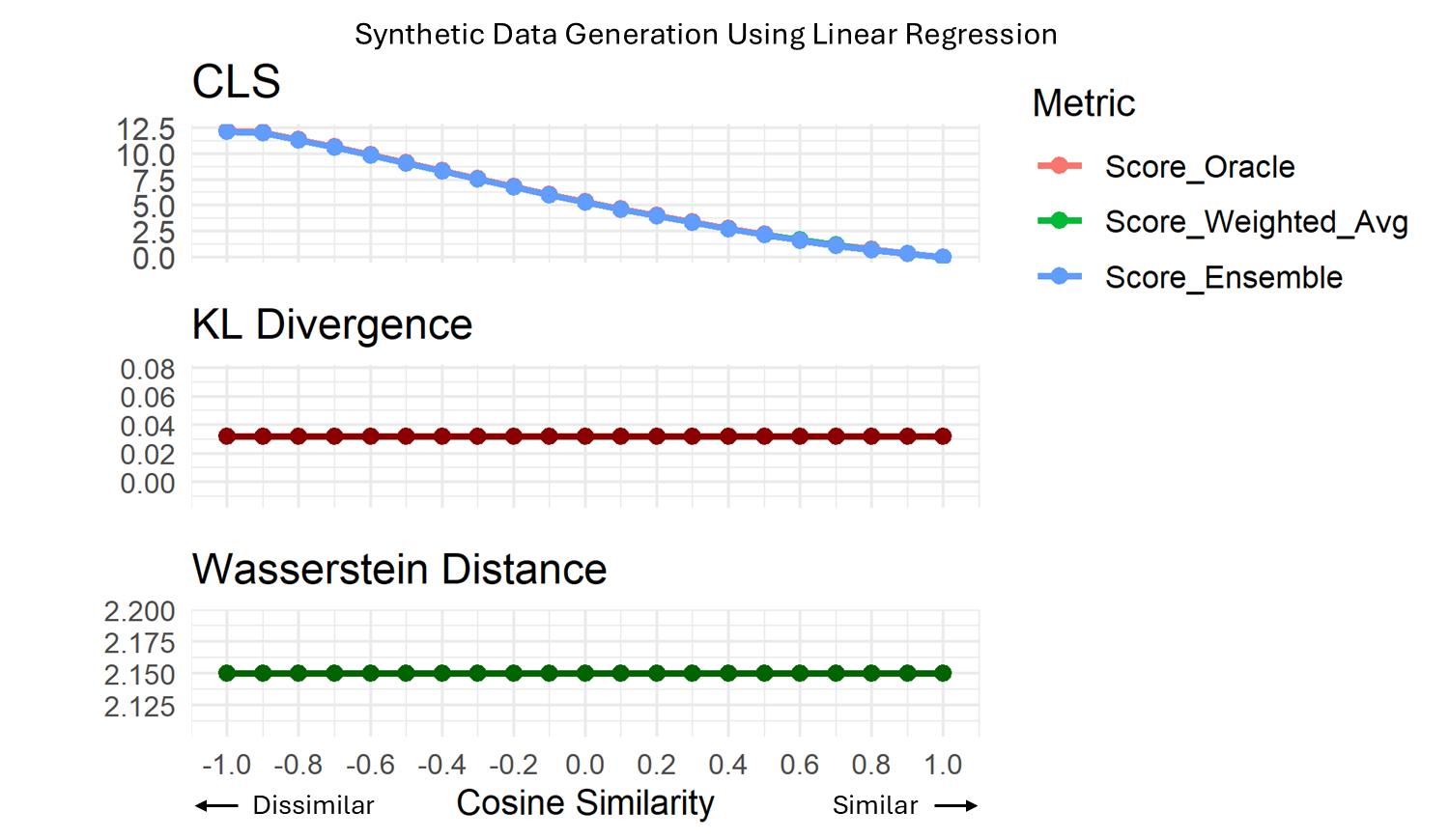}
    \caption{Comparison of CLS vs other similarity metrics
under the Linear Regression Setting}
    \label{fig:na_linr}
\end{figure}

\begin{figure}[!htbp]
    \centering
    \includegraphics[width=0.7\textwidth]{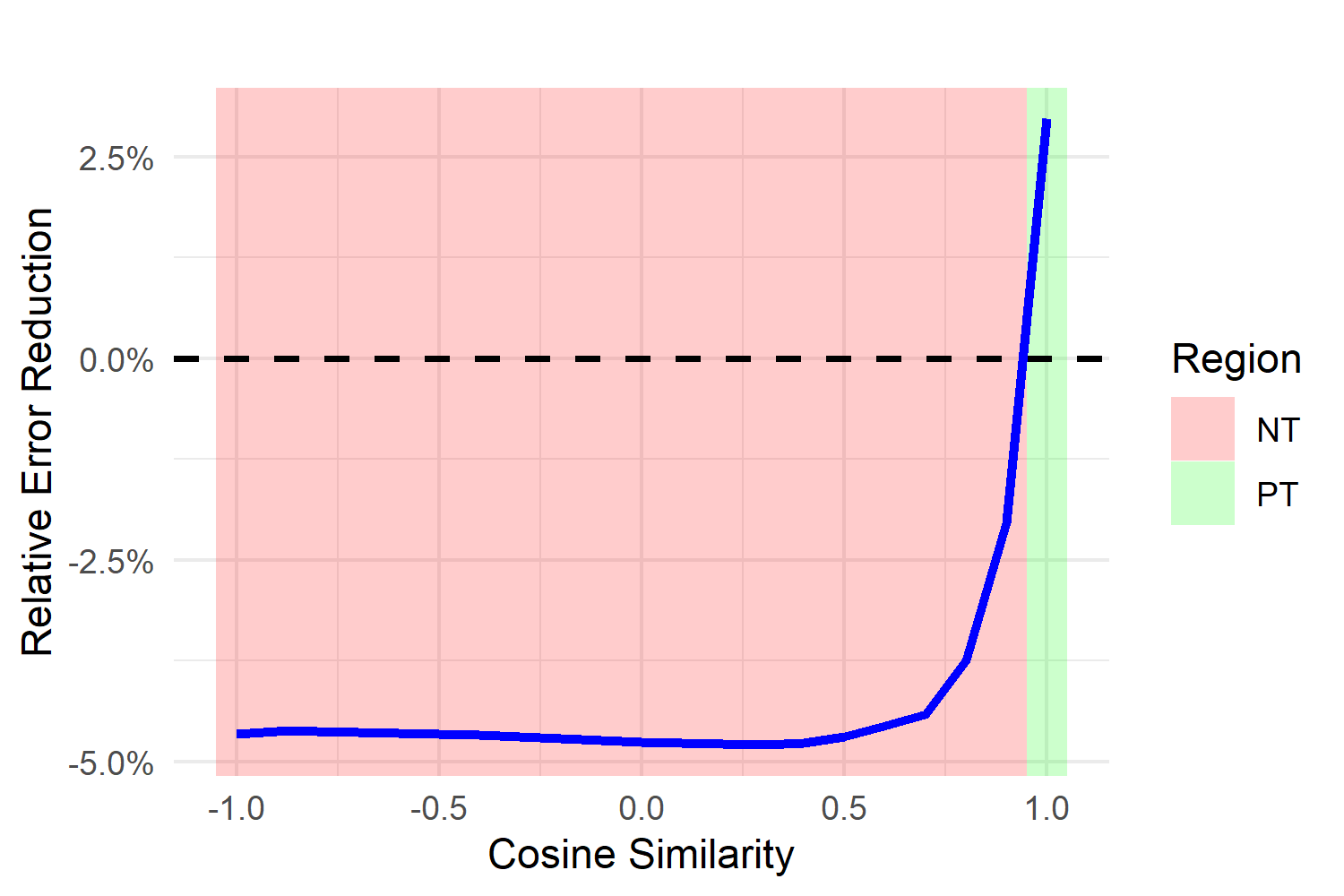}
    \caption{Transferability zones (positive, ambiguous,
and negative) identified under the linear regression model}
    \label{fig:trans_linr}
\end{figure}

All results for this setting are presented in Table~\ref{table1.4.6}, Table~\ref{table1.4.6w}, Figure~\ref{fig:na_linr}, and Figure~\ref{fig:trans_linr}.

\begin{table}[htbp]
\centering
\scalebox{0.9}{ 
\setlength{\tabcolsep}{2pt} 
\renewcommand{\arraystretch}{0.9} 
\begin{tabularx}{\textwidth}{
>{\centering\arraybackslash}X||
>{\centering\arraybackslash}X 
>{\centering\arraybackslash}X 
>{\centering\arraybackslash}X 
>{\centering\arraybackslash}X 
>{\centering\arraybackslash}X 
>{\centering\arraybackslash}X }
\toprule
\makecell{\textbf{Cosine}\\\textbf{Sim.}} &
\makecell{\textbf{Score}\\\textbf{Oracle}} &
\makecell{\textbf{Score}\\\textbf{Unw.}\\\textbf{Avg}} &
\makecell{\textbf{Score}\\\textbf{Wtd.}\\\textbf{Avg}} &
\makecell{\textbf{Score}\\\textbf{Ens.}} &
\makecell{\textbf{KL}\\\textbf{Div.}} &
\makecell{\textbf{Wass.}\\\textbf{Dist.}} \\
\midrule
-1.0 & 12.2696 & 10.9741 & 12.1673 & 12.1651 & 0.0319 & 2.1502 \\
-0.9 & 12.1056 & 10.8870 & 12.0155 & 12.0108 & 0.0319 & 2.1502 \\
-0.8 & 11.4376 & 10.2965 & 11.3561 & 11.3512 & 0.0319 & 2.1502 \\
-0.7 & 10.6938 & 9.6244  & 10.6194 & 10.6145 & 0.0319 & 2.1502 \\
-0.6 & 9.9206  & 8.9311  & 9.8513  & 9.8460  & 0.0319 & 2.1502 \\
-0.5 & 9.1379  & 8.2136  & 9.0719  & 9.0660  & 0.0319 & 2.1502 \\
-0.4 & 8.3566  & 7.4999  & 8.2929  & 8.2868  & 0.0319 & 2.1502 \\
-0.3 & 7.5843  & 6.7967  & 7.5226  & 7.5154  & 0.0319 & 2.1502 \\
-0.2 & 6.8261  & 6.1090  & 6.7642  & 6.7566  & 0.0319 & 2.1502 \\
-0.1 & 6.0863  & 5.4448  & 6.0253  & 6.0180  & 0.0319 & 2.1502 \\
 0.0 & 5.3682  & 4.7952  & 5.3072  & 5.2994  & 0.0319 & 2.1502 \\
 0.1 & 4.6749  & 4.1756  & 4.6158  & 4.6085  & 0.0319 & 2.1502 \\
 0.2 & 4.0089  & 3.5817  & 3.9507  & 3.9432  & 0.0319 & 2.1502 \\
 0.3 & 3.3729  & 3.0132  & 3.3156  & 3.3091  & 0.0319 & 2.1502 \\
 0.4 & 2.7692  & 2.4722  & 2.7133  & 2.7069  & 0.0319 & 2.1502 \\
 0.5 & 2.2004  & 1.9611  & 2.1455  & 2.1402  & 0.0319 & 2.1502 \\
 0.6 & 1.6691  & 1.4832  & 1.6164  & 1.6113  & 0.0319 & 2.1502 \\
 0.7 & 1.1784  & 1.0339  & 1.1281  & 1.1237  & 0.0319 & 2.1502 \\
 0.8 & 0.7321  & 0.6280  & 0.6860  & 0.6827  & 0.0319 & 2.1502 \\
 0.9 & 0.3357  & 0.2655  & 0.2985  & 0.2964  & 0.0319 & 2.1502 \\
 1.0 & 0.0000  & -0.0317 & -0.0105 & -0.0112 & 0.0319 & 2.1502 \\
\midrule
\multicolumn{1}{c||}{|$\rho_s$|} & 1.0000 & 1.0000 & 1.0000 & 1.0000 & 0.000 & 0.000 \\
\bottomrule
\end{tabularx}
}
\caption{Comparison of similarity metrics across varying cosine similarity values under the Linear Regression setting}
\label{table1.4.6}
\end{table}

\begin{table}[htbp]
\centering
\scalebox{0.9}{ 
\setlength{\tabcolsep}{2pt} 
\renewcommand{\arraystretch}{0.9} 
\begin{tabularx}{\textwidth}{>{\centering\arraybackslash}X||
>{\centering\arraybackslash}X 
>{\centering\arraybackslash}X 
>{\centering\arraybackslash}X 
>{\centering\arraybackslash}X |
>{\centering\arraybackslash}X 
>{\centering\arraybackslash}X 
>{\centering\arraybackslash}X 
>{\centering\arraybackslash}X }
\toprule
\makecell{\textbf{Cos.}\\\textbf{Sim.}} &
\makecell{\textbf{Lin.}\\\textbf{Regr.}} &
\makecell{\textbf{SVM}\\\textbf{Linear}} &
\makecell{\textbf{SVM}\\\textbf{Radial}} &
\makecell{\textbf{Xgb}\\\textbf{Tree}} &
\makecell{\textbf{Score}\\\textbf{Unw.}\\\textbf{Avg}} &
\makecell{\textbf{Score}\\\textbf{Wtd.}\\\textbf{Avg}} &
\makecell{\textbf{Score}\\\textbf{Ens.}} &
\makecell{\textbf{Score}\\\textbf{Oracle}} \\
\midrule
-1.0 & 12.1771 & 12.1327 & 9.8074 & 9.7792 & 10.9741 & 12.1673 & 12.1651 & 12.2696 \\
-0.9 & 12.0253 & 11.9902 & 9.6966 & 9.8359 & 10.8870 & 12.0155 & 12.0108 & 12.1056 \\
-0.8 & 11.3644 & 11.3379 & 9.1607 & 9.3233 & 10.2965 & 11.3561 & 11.3512 & 11.4376 \\
-0.7 & 10.6259 & 10.6052 & 8.5598 & 8.7068 & 9.6244 & 10.6194 & 10.6145 & 10.6938 \\
-0.6 & 9.8569  & 9.8407  & 7.9329 & 8.0937 & 8.9311 & 9.8513 & 9.8460 & 9.9206 \\
-0.5 & 9.0773  & 9.0635  & 7.2972 & 7.4165 & 8.2136 & 9.0719 & 9.0660 & 9.1379 \\
-0.4 & 8.2983  & 8.2863  & 6.6613 & 6.7537 & 7.4999 & 8.2929 & 8.2868 & 8.3566 \\
-0.3 & 7.5276  & 7.5192  & 6.0340 & 6.1059 & 6.7967 & 7.5226 & 7.5154 & 7.5843 \\
-0.2 & 6.7706  & 6.7641  & 5.4174 & 5.4839 & 6.1090 & 6.7642 & 6.7566 & 6.8261 \\
-0.1 & 6.0314  & 6.0278  & 4.8160 & 4.9040 & 5.4448 & 6.0253 & 6.0180 & 6.0863 \\
 0.0 & 5.3137  & 5.3098  & 4.2340 & 4.3235 & 4.7952 & 5.3072 & 5.2994 & 5.3682 \\
 0.1 & 4.6205  & 4.6188  & 3.6738 & 3.7894 & 4.1756 & 4.6158 & 4.6085 & 4.6749 \\
 0.2 & 3.9546  & 3.9540  & 3.1368 & 3.2812 & 3.5817 & 3.9507 & 3.9432 & 4.0089 \\
 0.3 & 3.3187  & 3.3187  & 2.6250 & 2.7906 & 3.0132 & 3.3156 & 3.3091 & 3.3729 \\
 0.4 & 2.7153  & 2.7159  & 2.1411 & 2.3164 & 2.4722 & 2.7133 & 2.7069 & 2.7692 \\
 0.5 & 2.1472  & 2.1498  & 1.6877 & 1.8597 & 1.9611 & 2.1455 & 2.1402 & 2.2004 \\
 0.6 & 1.6174  & 1.6208  & 1.2639 & 1.4305 & 1.4832 & 1.6164 & 1.6113 & 1.6691 \\
 0.7 & 1.1292  & 1.1323  & 0.8734 & 1.0006 & 1.0339 & 1.1281 & 1.1237 & 1.1784 \\
 0.8 & 0.6873  & 0.6892  & 0.5207 & 0.6148 & 0.6280 & 0.6860 & 0.6827 & 0.7321 \\
 0.9 & 0.2988  & 0.2997  & 0.2114 & 0.2520 & 0.2655 & 0.2985 & 0.2964 & 0.3357 \\
 1.0 & -0.0106 & -0.0154 & -0.0325 & -0.0683 & -0.0317 & -0.0105 & -0.0112 & 0.0000 \\
\midrule
\textbf{Diff.} & 0.0562 & 0.0651 & 1.1909 & 1.0826 & 0.5987 & 0.0607 & 0.0661 & - \\
\bottomrule
\end{tabularx}
}
\caption{$\text{Diff}$ between $\widehat{CLS}$ and oracle CLS under the Linear Regression setting}
\label{table1.4.6w}
\end{table}

\subsubsection{Non-linear Regression Setting}
Consider a regression problem where the features \( \mathbf{x}_i \in \mathbb{R}^p \) (for sample index \( i = 1, \ldots, n_t \) for the target dataset and \( i = 1, \ldots, n_s \) for the source dataset) are independently and identically distributed (i.i.d.) as \( \mathbf{x}_i \sim \mathcal{N}_p(0, I_p) \). We set the feature dimension to \( p = 10 \). Among these ten features, only the first five are used to generate the response variable \( y_i \) through the following nonlinear function:
\[
y_i = \beta_1 \sin(x_{i1}) + \beta_2 x_{i2}^2 + \beta_3 x_{i3}x_{i4} + \beta_4 \exp(x_{i5}) + \epsilon_i,
\]
where \( \beta = (\beta_1, \beta_2, \beta_3, \beta_4) \) is the parameter vector, \( \epsilon_i \sim \mathcal{N}(0,1) \) denotes Gaussian noise, and \( x_{ij} \) represents the \( j \)-th feature of the \( i \)-th sample.

We consider two datasets: the target dataset, characterized by the parameter vector \( \beta^{(t)} \), and the source dataset, characterized by the parameter vector \( \beta^{(s)} \). We replace \( \beta \) in the above nonlinear regression model with \( \beta^{(t)} \) and \( \beta^{(s)} \) to generate the target and source datasets, respectively. The coefficient vector \( \beta^{(t)} \) for the target data is drawn from \( \mathcal{N}_4\!\left(\tfrac{1}{4}\mathbf{1}_4, \tfrac{1}{16}\mathbf{I}_4\right) \), where \( \mathbf{1}_4 \) denotes a four-dimensional vector of ones and \( \mathbf{I}_4 \) denotes the \( 4 \times 4 \) identity matrix. The vector \( \beta^{(s)} \) is generated using the method described in Section~\ref{subsec:generate} to control the cosine similarity between \( \beta^{(t)} \) and \( \beta^{(s)} \).

The Oracle CLS is computed using \(\beta^{(t)},
 \beta^{(s)}\) based on the above data generation mechanism.

\begin{figure}[!htbp]
    \centering
    \includegraphics[width=0.8\textwidth]{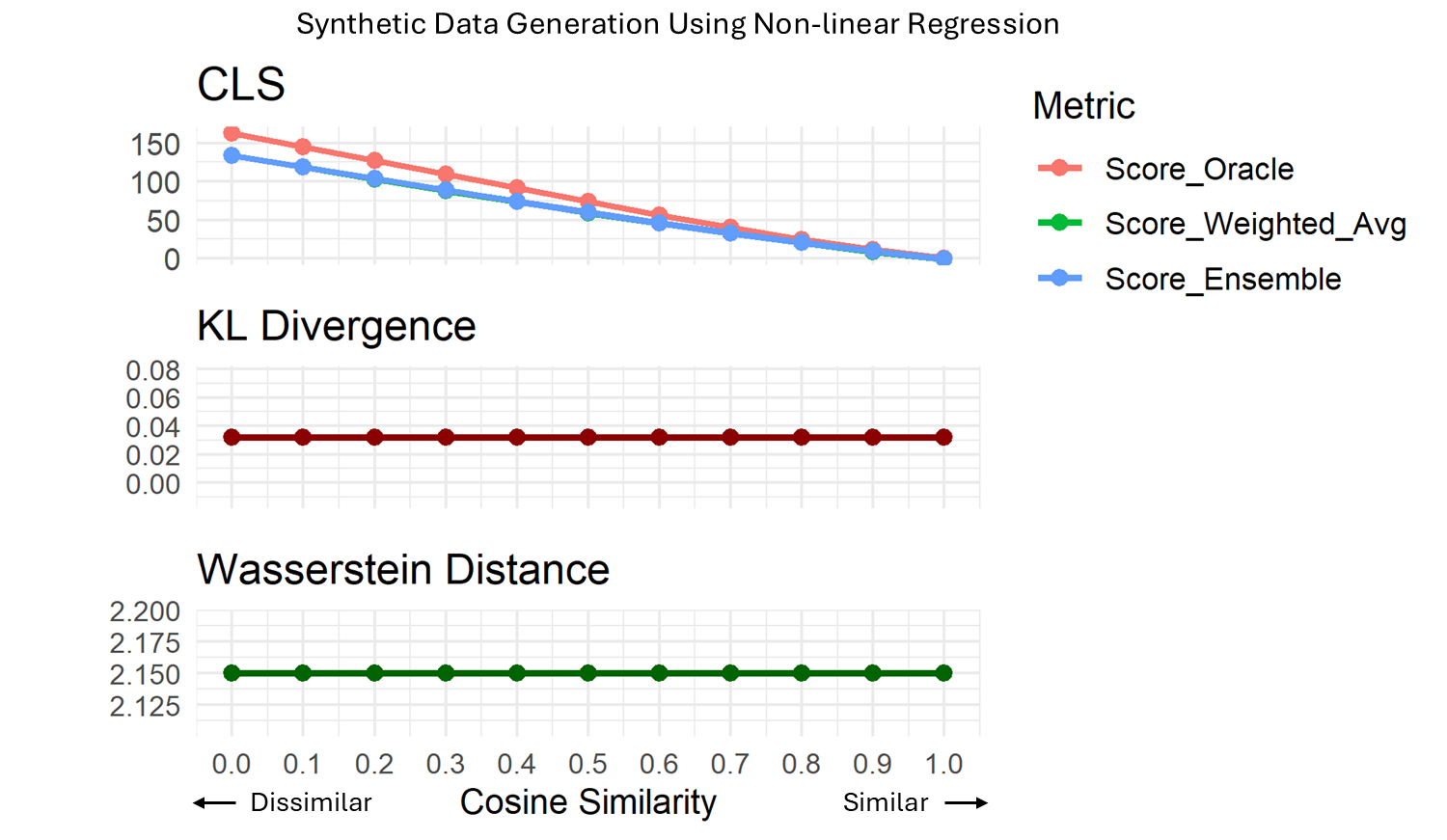}
    \caption{Comparison of CLS vs other similarity metrics
under the Non-linear Regression Setting}
    \label{fig:na_nlinr}
\end{figure}

\begin{figure}[!htbp]
    \centering
    \includegraphics[width=0.7\textwidth]{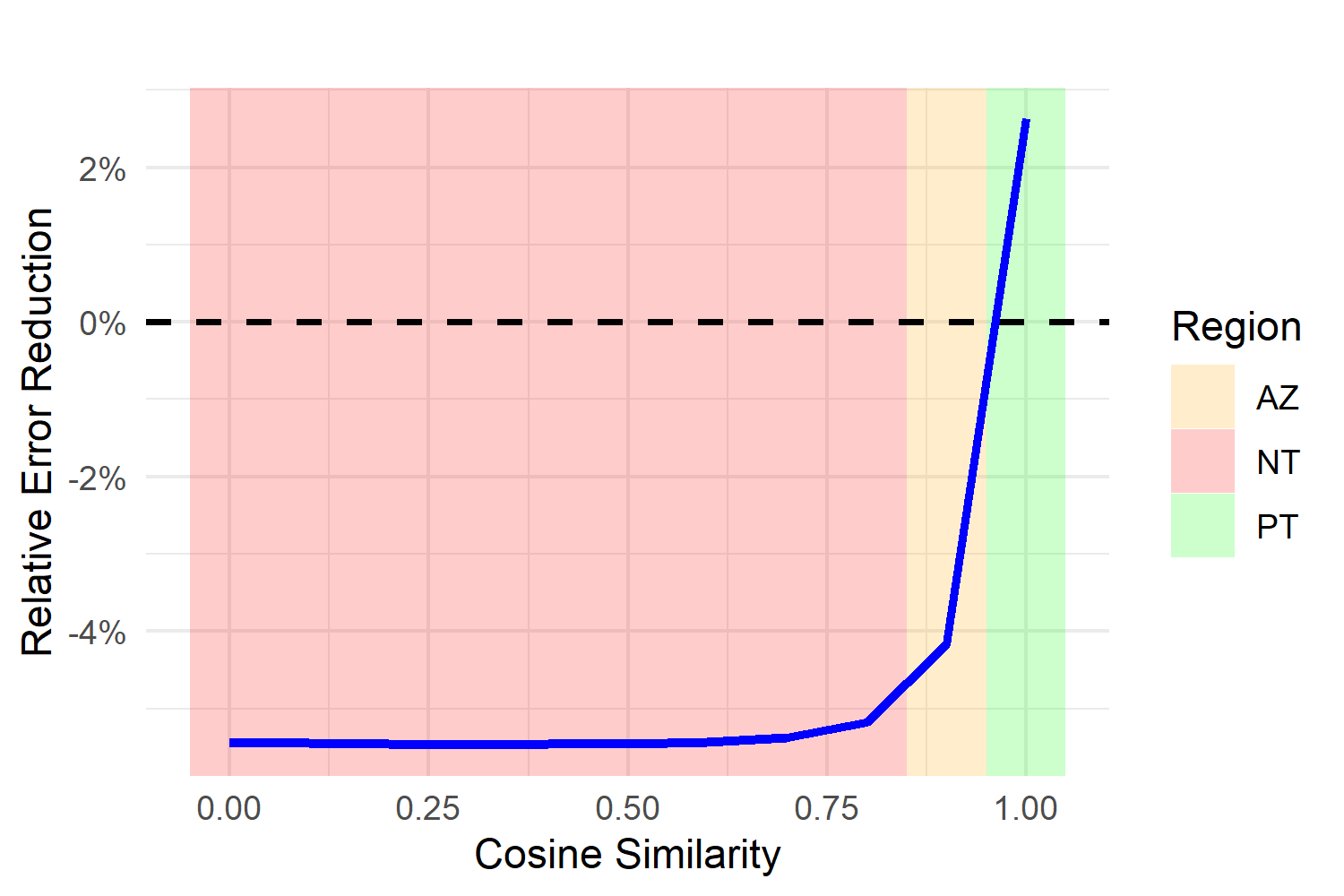}
    \caption{Transferability zones (positive, ambiguous,
and negative) identified under the non-linear regression model}
    \label{fig:trans_nlinr}
\end{figure}

All results for this setting are presented in Table~\ref{table1.4.7}, Table~\ref{table1.4.7w}, Figure~\ref{fig:na_nlinr}, and Figure~\ref{fig:trans_nlinr}.

\begin{table}[htbp]
\centering
\scalebox{0.9}{ 
\setlength{\tabcolsep}{2pt} 
\renewcommand{\arraystretch}{0.9} 
\begin{tabularx}{\textwidth}{
>{\centering\arraybackslash}X||
>{\centering\arraybackslash}X 
>{\centering\arraybackslash}X 
>{\centering\arraybackslash}X 
>{\centering\arraybackslash}X 
>{\centering\arraybackslash}X 
>{\centering\arraybackslash}X }
\toprule
\makecell{\textbf{Cosine}\\\textbf{Sim.}} &
\makecell{\textbf{Score}\\\textbf{Oracle}} &
\makecell{\textbf{Score}\\\textbf{Unw.}\\\textbf{Avg}} &
\makecell{\textbf{Score}\\\textbf{Wtd.}\\\textbf{Avg}} &
\makecell{\textbf{Score}\\\textbf{Ens.}} &
\makecell{\textbf{KL}\\\textbf{Div.}} &
\makecell{\textbf{Wass.}\\\textbf{Dist.}} \\
\midrule
0.0 & 163.1004 & 114.4225 & 133.8301 & 134.1597 & 0.0319 & 2.1502 \\
0.1 & 145.4410 & 101.9064 & 118.4945 & 118.8173 & 0.0319 & 2.1502 \\
0.2 & 127.4692 & 89.1052  & 102.9292 & 103.5232 & 0.0319 & 2.1502 \\
0.3 & 109.3899 & 76.4648  & 88.2859  & 88.9679  & 0.0319 & 2.1502 \\
0.4 & 91.4039  & 63.7664  & 73.6296  & 74.1245  & 0.0319 & 2.1502 \\
0.5 & 73.7163  & 51.3485  & 59.2396  & 59.7947  & 0.0319 & 2.1502 \\
0.6 & 56.5480  & 39.2574  & 45.6945  & 45.8577  & 0.0319 & 2.1502 \\
0.7 & 40.1526  & 27.7033  & 33.0285  & 33.2374  & 0.0319 & 2.1502 \\
0.8 & 24.8494  & 16.9727  & 20.3167  & 20.8286  & 0.0319 & 2.1502 \\
0.9 & 11.1095  & 7.2266   & 8.7648   & 9.1721   & 0.0319 & 2.1502 \\
1.0 & 0.0000   & -0.6144  & -0.5602  & -0.4073  & 0.0319 & 2.1502 \\
\midrule
\multicolumn{1}{c||}{|$\rho_s$|} & 1.0000 & 1.0000 & 1.0000 & 1.0000 & 0.0000 & 0.0000 \\
\bottomrule
\end{tabularx}
}
\caption{Comparison of similarity metrics across varying cosine similarity values under the Non-linear Regression setting}
\label{table1.4.7}
\end{table}

\begin{table}[htbp]
\centering
\scalebox{0.9}{ 
\setlength{\tabcolsep}{2pt} 
\renewcommand{\arraystretch}{0.9} 
\begin{tabularx}{\textwidth}{>{\centering\arraybackslash}X||
>{\centering\arraybackslash}X 
>{\centering\arraybackslash}X 
>{\centering\arraybackslash}X 
>{\centering\arraybackslash}X |
>{\centering\arraybackslash}X 
>{\centering\arraybackslash}X 
>{\centering\arraybackslash}X 
>{\centering\arraybackslash}X }
\toprule
\makecell{\textbf{Cos.}\\\textbf{Sim.}} &
\makecell{\textbf{Lin.}\\\textbf{Regr.}} &
\makecell{\textbf{SVM}\\\textbf{Linear}} &
\makecell{\textbf{SVM}\\\textbf{Radial}} &
\makecell{\textbf{Xgb}\\\textbf{Tree}} &
\makecell{\textbf{Score}\\\textbf{Unw.}\\\textbf{Avg}} &
\makecell{\textbf{Score}\\\textbf{Wtd.}\\\textbf{Avg}} &
\makecell{\textbf{Score}\\\textbf{Ens.}} &
\makecell{\textbf{Score}\\\textbf{Oracle}} \\
\midrule
0.0 & 110.3842 & 90.6661 & 120.3661 & 136.2734 & 114.4225 & 133.8301 & 134.1597 & 163.1004 \\
0.1 & 98.2945  & 81.0120 & 107.4225 & 120.8966 & 101.9064 & 118.4945 & 118.8173 & 145.4410 \\
0.2 & 86.0105  & 71.1300 & 94.2608  & 105.0196 & 89.1052  & 102.9292 & 103.5232 & 127.4692 \\
0.3 & 73.6706  & 61.1366 & 80.9996  & 90.0525  & 76.4648  & 88.2859  & 88.9679  & 109.3899 \\
0.4 & 61.4109  & 51.1891 & 67.7074  & 74.7582  & 63.7664  & 73.6296  & 74.1245  & 91.4039  \\
0.5 & 49.3708  & 41.2126 & 54.5649  & 60.2458  & 51.3485  & 59.2396  & 59.7947  & 73.7163  \\
0.6 & 37.7011  & 31.3307 & 41.6150  & 46.3829  & 39.2574  & 45.6945  & 45.8577  & 56.5480  \\
0.7 & 26.5761  & 21.4889 & 29.0567  & 33.6913  & 27.7033  & 33.0285  & 33.2374  & 40.1526  \\
0.8 & 16.2178  & 12.8479 & 17.6235  & 21.2016  & 16.9727  & 20.3167  & 20.8286  & 24.8494  \\
0.9 & 6.9607   & 5.4516  & 7.5386   & 8.9554   & 7.2266   & 8.7648   & 9.1721   & 11.1095  \\
1.0 & -0.3019  & -0.2947 & -0.6618  & -1.1991  & -0.6144  & -0.5602  & -0.4073  & 0.0000   \\
\midrule
\textbf{Diff.} & 25.1714 & 34.1827 & 20.2443 & 13.3547 & 23.2383 & 14.5025 & 14.1004 & - \\
\bottomrule
\end{tabularx}
}
\caption{$\text{Diff}$ between $\widehat{CLS}$ and oracle CLS under the Non-linear Regression setting}
\label{table1.4.7w}
\end{table}

\subsection{Synthetic Data Generation for Multi-class Classification}
\label{subsec:mc}

For multi-class classification, we design a four-class task in which samples are projected onto two orthogonal directions and assigned to quadrants based on the signs of the projections, with Gaussian noise added for perturbation.

For CLS estimation, we employ four models for multi-class classification: multinomial logistic regression, linear SVM, radial basis function (RBF) SVM, and XGBoost.

Among the four transfer learning methods evaluated in Section~5.3 (Evaluation of Transferable Zones), both TrAdaBoost and Dynamic TrAdaBoost are originally developed for binary classification, and therefore are not applied in the multi-class classification experiments.

\subsubsection{4-class Classification Setting}

We consider a four-class classification problem in which the feature vectors 
\( \mathbf{x}_i \in \mathbb{R}^p \) (for sample index \( i = 1, \ldots, n_t \) for the target dataset and \( i = 1, \ldots, n_s \) for the source dataset) are independently and identically distributed (i.i.d.) as \( \mathbf{x}_i \sim \mathcal{N}_p(\mathbf{0}, I_p) \), with the feature dimension set to \( p = 10 \). We generate the data using two parameter vectors \( \beta_1, \beta_2 \in \mathbb{R}^p \), which are constructed to be orthogonal, i.e., \( \beta_1 \perp \beta_2 \).

Given the first coefficient vector $\beta_1$, the second coefficient vector $\beta_2$ is constructed such that 
$\beta_1^{\top}\beta_2 = 0$ and 
$\|\beta_2\| = \|\beta_1\|$, 
using the Gram--Schmidt orthogonalization procedure.

Given $\beta_1$ and $\beta_2$, the class label for each sample $\mathbf{x}_i$ 
is determined by the signs of its projections onto the two parameter vectors:
\[
\zeta_1 = \mathbf{x}_i^{\top}\beta_1 + \epsilon_1, \quad
\zeta_2 = \mathbf{x}_i^{\top}\beta_2 + \epsilon_2,
\]
where $\epsilon_1, \epsilon_2 \sim \mathcal{N}(0, \sigma^2)$ are independent Gaussian noise terms 
that introduce soft decision boundaries and we set the noise level \( \sigma = 0.3 \).
The resulting class label $y_i$ is assigned according to the quadrant in the projection space:
\[
y_i =
\begin{cases}
1, & \text{if } \zeta_1 < 0 \text{ and } \zeta_2 < 0,\\
2, & \text{if } \zeta_1 < 0 \text{ and } \zeta_2 \ge 0,\\
3, & \text{if } \zeta_1 \ge 0 \text{ and } \zeta_2 < 0,\\
4, & \text{if } \zeta_1 \ge 0 \text{ and } \zeta_2 \ge 0.
\end{cases}
\]
This construction yields a synthetic dataset with four linearly separable classes 
in the $(\zeta_1, \zeta_2)$ plane, 
while the Gaussian noise $(\epsilon_1, \epsilon_2)$ controls the degree of boundary smoothness.

We consider two datasets: the target dataset characterized by the pair \((\beta_1^{(t)}, \beta_2^{(t)})\) 
and the source dataset characterized by the pair \((\beta_1^{(s)}, \beta_2^{(s)})\). We replace \((\beta_1,\beta_2)\) mentioned above with \( (\beta_1^{(t)}, \beta_2^{(t)}),(\beta_1^{(s)}, \beta_2^{(s)})\) to generate the target and source dataset separately. 

The coefficient vector $\beta_1^{(t)}$ for the target data is drawn from $\mathcal{N}_p\!\left(\tfrac{1}{4}\mathbf{1}, \tfrac{1}{16}\mathbf{I}_p\right)$, where $\mathbf{1}$ denotes a $p$-dimensional vector of ones,  and $\mathbf{I}_p$ denotes the $p \times p$ identity matrix. The coefficient vector \(\beta_2^{(t)}\) is constructed such that \(\beta_1^{(t)\top}\beta_2^{(t)}=0\) and \( \|\beta_2^{(t)}\| = \|\beta_1^{(t)}\| \) using the Gram--Schmidt orthogonalization procedure.

We control the similarity between the datasets by adjusting the cosine similarity between 
\(\beta_1^{(t)}\) and \(\beta_1^{(s)}\) using the procedure described in Section~\ref{subsec:generate}, and \(\beta_2^{(s)}\) is constructed such that \(\beta_1^{(s)\top}\beta_2^{(s)}=0\) and \( \|\beta_2^{(s)}\| = \|\beta_1^{(s)}\| \) using the Gram--Schmidt orthogonalization procedure.

The Oracle CLS is then computed using \((\beta_1^{(t)}, \beta_2^{(t)})\) and \((\beta_1^{(s)}, \beta_2^{(s)})\) based on the above data generation mechanism.

\begin{figure}[!htbp]
    \centering
    \includegraphics[width=0.8\textwidth]{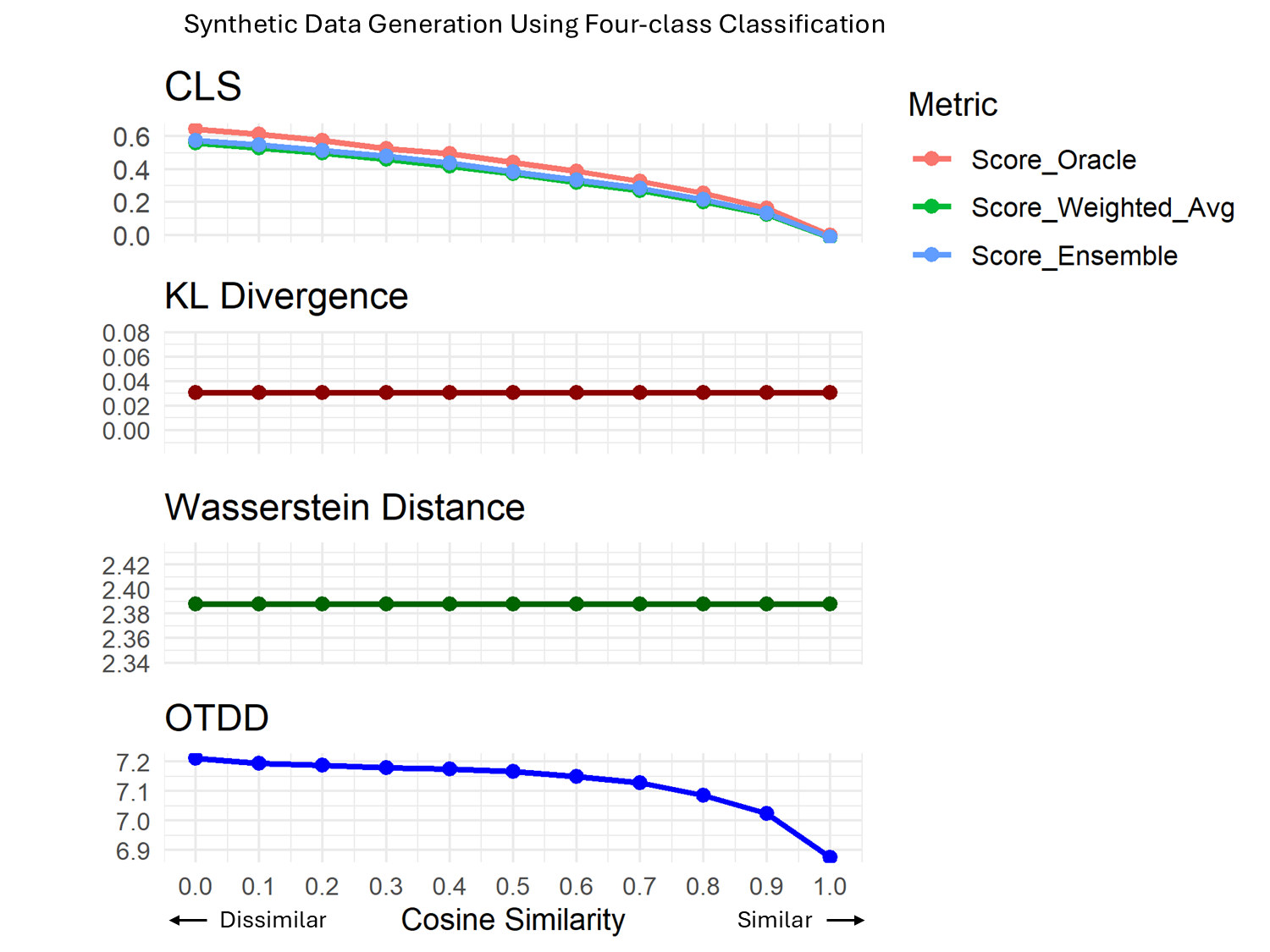}
    \caption{Comparison of CLS vs other similarity metrics
under the 4-Class Classification Setting}
    \label{fig:na_multi}
\end{figure}

\begin{figure}[!htbp]
    \centering
    \includegraphics[width=0.7\textwidth]{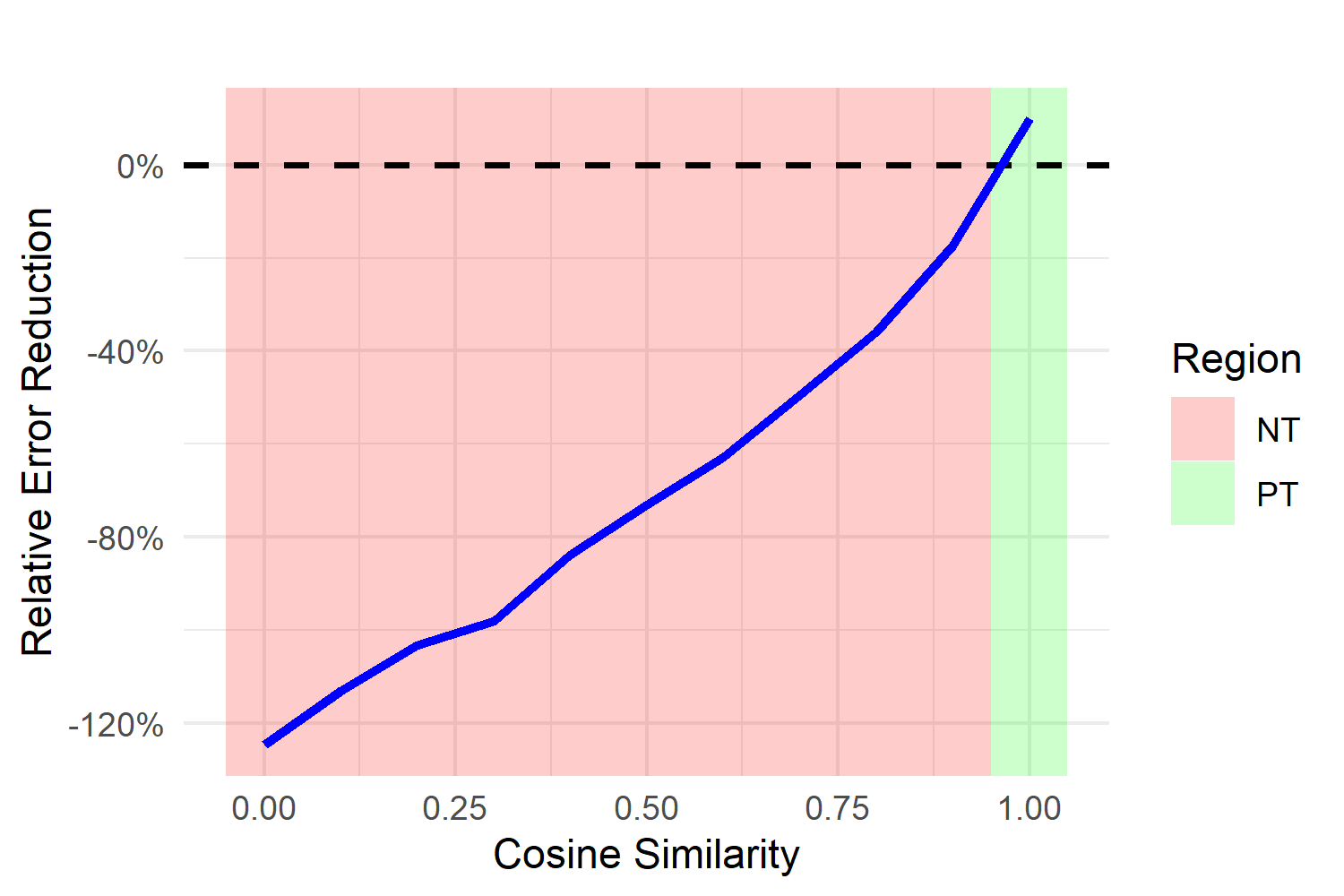}
    \caption{Transferability zones (positive, ambiguous,
and negative) identified under the 4-class classification model}
    \label{fig:trans_multi}
\end{figure}

All results for this setting are presented in Table~\ref{table1.4.8}, Table~\ref{table1.4.8w}, Figure~\ref{fig:na_multi}, and Figure~\ref{fig:trans_multi}.

\begin{table}[htbp]
\centering
\scalebox{0.9}{ 
\setlength{\tabcolsep}{2pt} 
\renewcommand{\arraystretch}{0.9} 
\begin{tabularx}{\textwidth}{
>{\centering\arraybackslash}X||
>{\centering\arraybackslash}X 
>{\centering\arraybackslash}X 
>{\centering\arraybackslash}X 
>{\centering\arraybackslash}X 
>{\centering\arraybackslash}X 
>{\centering\arraybackslash}X 
>{\centering\arraybackslash}X }
\toprule
\makecell{\textbf{Cosine}\\\textbf{Sim.}} &
\makecell{\textbf{Score}\\\textbf{Oracle}} &
\makecell{\textbf{Score}\\\textbf{Unw.}\\\textbf{Avg}} &
\makecell{\textbf{Score}\\\textbf{Wtd.}\\\textbf{Avg}} &
\makecell{\textbf{Score}\\\textbf{Ens.}} &
\makecell{\textbf{KL}\\\textbf{Div.}} &
\makecell{\textbf{Wass.}\\\textbf{Dist.}} &
\textbf{OTDD} \\
\midrule
0.0 & 0.6453 & 0.5161 & 0.5599 & 0.5754 & 0.0306 & 2.388 & 7.2108 \\
0.1 & 0.6112 & 0.4909 & 0.5302 & 0.5490 & 0.0306 & 2.388 & 7.1949 \\
0.2 & 0.5764 & 0.4574 & 0.4967 & 0.5147 & 0.0306 & 2.388 & 7.1876 \\
0.3 & 0.5243 & 0.4240 & 0.4594 & 0.4777 & 0.0306 & 2.388 & 7.1794 \\
0.4 & 0.4934 & 0.3864 & 0.4184 & 0.4367 & 0.0306 & 2.388 & 7.1745 \\
0.5 & 0.4427 & 0.3416 & 0.3708 & 0.3844 & 0.0306 & 2.388 & 7.1661 \\
0.6 & 0.3862 & 0.2933 & 0.3205 & 0.3345 & 0.0306 & 2.388 & 7.1492 \\
0.7 & 0.3264 & 0.2433 & 0.2685 & 0.2838 & 0.0306 & 2.388 & 7.1274 \\
0.8 & 0.2555 & 0.1818 & 0.2021 & 0.2142 & 0.0306 & 2.388 & 7.0864 \\
0.9 & 0.1625 & 0.1078 & 0.1232 & 0.1308 & 0.0306 & 2.388 & 7.0238 \\
1.0 & 0.0000 & -0.0144 & -0.0161 & -0.0120 & 0.0306 & 2.388 & 6.8747 \\
\midrule
\multicolumn{1}{c||}{$\left|\rho_{s}\right|$} 
& 1.0000 & 1.0000 & 1.0000 & 1.0000 & 0.0000 & 0.0000 & 1.0000 \\
\bottomrule
\end{tabularx}
}
\caption{Comparison of similarity metrics across varying cosine similarity values under the 4-class Classification setting}
\label{table1.4.8}
\end{table}

\begin{table}[htbp]
\centering
\scalebox{0.9}{ 
\setlength{\tabcolsep}{2pt} 
\renewcommand{\arraystretch}{0.9} 
\begin{tabularx}{\textwidth}{>{\centering\arraybackslash}X||
>{\centering\arraybackslash}X 
>{\centering\arraybackslash}X 
>{\centering\arraybackslash}X 
>{\centering\arraybackslash}X |
>{\centering\arraybackslash}X 
>{\centering\arraybackslash}X 
>{\centering\arraybackslash}X 
>{\centering\arraybackslash}X }
\toprule
\makecell{\textbf{Cos.}\\\textbf{Sim.}} &
\makecell{\textbf{MLR}} &
\makecell{\textbf{SVM}\\\textbf{Linear}} &
\makecell{\textbf{SVM}\\\textbf{Radial}} &
\makecell{\textbf{Xgb}\\\textbf{Tree}} &
\makecell{\textbf{Score}\\\textbf{Unw.}\\\textbf{Avg}} &
\makecell{\textbf{Score}\\\textbf{Wtd.}\\\textbf{Avg}} &
\makecell{\textbf{Score}\\\textbf{Ens.}} &
\makecell{\textbf{Score}\\\textbf{Oracle}} \\
\midrule
0.0 & 0.5781 & 0.5656 & 0.5017 & 0.4190 & 0.5161 & 0.5599 & 0.5754 & 0.6453 \\
0.1 & 0.5511 & 0.5367 & 0.4751 & 0.4006 & 0.4909 & 0.5302 & 0.5490 & 0.6112 \\
0.2 & 0.5162 & 0.5027 & 0.4399 & 0.3709 & 0.4574 & 0.4967 & 0.5147 & 0.5764 \\
0.3 & 0.4792 & 0.4675 & 0.4100 & 0.3393 & 0.4240 & 0.4594 & 0.4777 & 0.5243 \\
0.4 & 0.4340 & 0.4264 & 0.3725 & 0.3127 & 0.3864 & 0.4184 & 0.4367 & 0.4934 \\
0.5 & 0.3849 & 0.3791 & 0.3320 & 0.2705 & 0.3416 & 0.3708 & 0.3844 & 0.4427 \\
0.6 & 0.3327 & 0.3270 & 0.2849 & 0.2288 & 0.2933 & 0.3205 & 0.3345 & 0.3862 \\
0.7 & 0.2827 & 0.2725 & 0.2351 & 0.1831 & 0.2433 & 0.2685 & 0.2838 & 0.3264 \\
0.8 & 0.2109 & 0.2108 & 0.1755 & 0.1297 & 0.1818 & 0.2021 & 0.2142 & 0.2555 \\
0.9 & 0.1302 & 0.1254 & 0.1041 & 0.0717 & 0.1078 & 0.1232 & 0.1308 & 0.1625 \\
1.0 & -0.0128 & -0.0106 & -0.0153 & -0.0190 & -0.0144 & -0.0161 & -0.0120 & 0.0000 \\
\midrule
\textbf{Diff.} 
& 0.0488 
& 0.0564 
& 0.1008 
& 0.1560 
& 0.0905 
& 0.0628 
& 0.0486 
& - \\
\bottomrule
\end{tabularx}
}
\caption{$\text{Diff}$ between $\widehat{CLS}$ and oracle CLS under the 4-Class Classification Setting}
\label{table1.4.8w}
\end{table}

\section{More Details of Applications to Real-world Data}

All experiments were performed on the High Performance Computing (HPC) cluster, equipped with one NVIDIA Tesla P100 GPU (16GB), 4 CPU cores, and 32GB of system memory. Each experiment was repeated 10 times, and the reported results represent the average across all runs.

Our code and data will be released publicly upon acceptance.
Our code will be available at \href{https://github.com/ShudongSun/CLS}{https://github.com/ShudongSun/CLS}.

\subsection{eICU Mortality Prediction}

More details about this experiment are described here.

We study ICU mortality prediction using the eICU Collaborative Research Database \citep{pollard2018eicu}, a multi-center dataset containing de-identified health records from U.S. hospitals between 2014 and 2015. To predict the post-ICU survival status of patients, we constructed a feature set that integrates multiple aspects of patient care. The features include demographics and admission information (age, gender, ethnicity, height, weight, admission type, and ICU type), hospital-level descriptors (hospital identifier, bed capacity category, region, and teaching status), acute physiological variables from the APACHE APS system (vital signs, laboratory values, Glasgow Coma Scale components, and indicators of organ support such as intubation, ventilation, dialysis, and urine output), as well as chronic health conditions and comorbidities (diabetes, cirrhosis, metastatic cancer, immunosuppression, elective surgery status, and others). After removing records with missing data, we calculated the post-ICU mortality rate for each hospital. To balance the outcome distribution, we undersampled hospital data so that the number of survivors and non-survivors was equal.

Our target task is to predict post-ICU survival status in teaching hospitals. We designated the hospital with the highest mortality rate (9.47\%, with $n_{\text{post-ICU alive}} = 43$) as the target dataset. Among the remaining hospitals, we selected the top 10 source hospitals with the largest number of post-ICU survivors.

Additional experimental details and results are described in the main text.

\subsection{Canine Image Classification}

More details about this experiment are described here.

Our target dataset is the Roboflow Dogs vs Wolves dataset, which contains 194 dog images and 412 wolf images \citep{zhong2023learning, zhong2024intrinsic, zhong2025splitz, zhong2024learning, zhang2024filtered}. The relatively small size and class imbalance make this dataset difficult to classify accurately. As source datasets, we considered: (i) Kaggle Dogs vs Wolves (1,000 dogs and 1,000 wolves), which is more balanced and substantially larger; (ii) Kaggle Cats vs Dogs (a random sample of 1,000 cats and 1,000 dogs), which shares one class with the target task but introduces a distinct source species; and (iii) Kaggle Horses vs Camels (200 horses and 200 camels), which has a similar binary structure but with entirely different categories. These source datasets provide different levels of semantic and distributional similarity relative to the target task.

We employed the Encoder--Head CLS introduced in Algorithm~3, using ResNet-18 as the backbone encoder. And we initialized the model with ImageNet-pretrained weights to ensure stable convergence. As before, a shared encoder was trained jointly on the target and source datasets, followed by domain-specific heads for cross-domain evaluation.

For empirical validation, we used the 5-fold cross-validation to estimate the test errors.

\end{document}